\documentclass[10pt,journal,compsoc]{IEEEtran}

\usepackage[T1]{fontenc}
\usepackage{amsmath,amsfonts}
\usepackage{algorithmic}
\usepackage{algorithm}
\usepackage{array}
\usepackage{textcomp}
\usepackage{stfloats}
\usepackage{url}
\usepackage{verbatim}
\usepackage{graphicx}
\usepackage{cite}
\usepackage{threeparttable}
\usepackage{xcolor}
\usepackage{multirow}
\usepackage{gensymb}
\usepackage{booktabs}
\usepackage{amssymb}
\usepackage{enumitem}
\usepackage{hyperref}
\usepackage[hyphenbreaks]{breakurl}
\usepackage{mathtools}
\usepackage{xspace}
\usepackage{diagbox}
\usepackage{graphbox}
\usepackage{colortbl}
\usepackage{subfigure}

\usepackage[capitalize]{cleveref}
\crefname{section}{Sec.}{Secs.}
\Crefname{section}{Section}{Sections}
\Crefname{table}{Table}{Tables}
\crefname{table}{Tab.}{Tabs.}

\hypersetup{hidelinks} 

\makeatletter
\newcommand*{\rom}[1]{\expandafter\@slowromancap\romannumeral #1@}
\makeatother

\newcommand{\ie}{{\emph{i.e.}}\xspace}
\newcommand{\eg}{{\emph{e.g.}}\xspace}

\newcommand{\etal}{{\emph{et al.}}}

\graphicspath{ {./figure/} }

\hypersetup{
    colorlinks=true,
    linkcolor=black,
    citecolor=green,
    urlcolor=magenta
}

\ifCLASSINFOpdf
\else
\fi

\begin{document}

\title{ Towards Realistic Low-Light Image Enhancement via ISP--Driven Data Modeling}

\author{Zhihua~Wang, Yu~Long, Qinghua~Lin,  Kai~Zhang$^\dagger$, Yazhu Zhang, Yuming~Fang, Li~Liu, Xiaochun~Cao
        
\IEEEcompsocitemizethanks{
\IEEEcompsocthanksitem Z. Wang is with the Department of Computer Science, City University of Hong Kong, Kowloon, Hong Kong (e-mail: zhihua.wang@my.cityu.edu.hk).

\IEEEcompsocthanksitem  Y. Long and Q. Lin are with the Department of Engineering, Shenzhen MSU-BIT University, Shenzhen, China (e-mail: longyu2001@bit.edu.cn, 2112405034@mail2.gdut.edu.cn).

\IEEEcompsocthanksitem K. Zhang is with the School of Intelligence Science and Technology, Nanjing University, Suzhou, China (e-mail: kaizhang@nju.edu.cn). 

\IEEEcompsocthanksitem Y. Fang is with the
School of Information Technology, Jiangxi University of Finance and Economics, Nanchang, China (e-mail: fa0001ng@e.ntu.edu.sg).

\IEEEcompsocthanksitem L. Liu is with the College of Electronic Science and Technology,
National University of Defense Technology, 
Changsha, China (e-mail: liuli\_nudt@nudt.edu.cn). 

\IEEEcompsocthanksitem Y. Zhang and X. Cao are with the School of Cyber Science and Technology, Shenzhen Campus of Sun Yat-sen University, Shenzhen, China (e-mail:zhangyzh236@mail.sysu.edu.cn, caoxiaochun@mail.sysu.edu.cn).

% \IEEEcompsocthanksitem C. Huang and X. Cao are with the School of
% Cyber Science and Technology, Shenzhen Campus
% of Sun Yat-sen University, Shenzhen 518000, China (e-mail: huangch253@mail.sysu.edu.cn, caoxiaochun@mail.sysu.edu.cn).

}% <-this % stops an unwanted space
\thanks{$^\dagger$ indicates the corresponding author.}
}

\markboth{}%
{Shell \MakeLowercase{\textit{et al.}}: Bare Demo of IEEEtran.cls for Computer Society Journals}

\IEEEtitleabstractindextext{%
\begin{abstract}

Deep neural networks (DNNs) have recently become the leading method for low-light image enhancement (LLIE). However, despite significant progress, their outputs may still exhibit issues such as amplified noise, incorrect white balance, or unnatural enhancements when deployed in real-world applications. A key challenge is the lack of diverse, large-scale training data that captures the complexities of low-light conditions and imaging pipelines.
In this paper, we propose a novel image signal processing (ISP)–driven data synthesis pipeline that addresses these challenges by generating unlimited paired training data. Specifically, our pipeline begins with easily collected high-quality normal-light images, which are first unprocessed into the RAW format using a reverse ISP. We then synthesize low-light degradations directly in the RAW domain. The resulting data is subsequently processed through a series of ISP stages—including white balance adjustment, color space conversion, tone mapping, and gamma correction—with controlled variations introduced at each stage. This broadens the degradation space and enhances the diversity of the training data, enabling the generated data to capture a wide range of degradations and the complexities inherent in the ISP pipeline. To demonstrate the effectiveness of our synthetic pipeline, we conduct extensive experiments using a vanilla U-Net model consisting solely of convolutional layers, group normalization, GeLU activation, and convolutional block attention modules (CBAMs). Extensive testing across multiple datasets reveals that the vanilla U-Net model trained with our data synthesis pipeline delivers high-fidelity, visually appealing enhancement results, surpassing state-of-the-art (SOTA) methods both quantitatively and qualitatively. Additionally, we apply our pipeline to existing LLIE methods, demonstrating its significant impact on improving their practicality and generalizability in real-world applications. Our codes are publicly available at \url{https://github.com/SMBU-MM/LLIE}.

\end{abstract}

}

\maketitle

\IEEEdisplaynontitleabstractindextext

% \fi
\IEEEpeerreviewmaketitle

\IEEEraisesectionheading{\section{Introduction}
\label{sec:introduction}}

\begin{figure}[h!]
    \centering
    \begin{minipage}[]{0.5\textwidth}
    \hskip -1em
        \centering
             \subfigure[Input]{\includegraphics[width=0.3\linewidth]{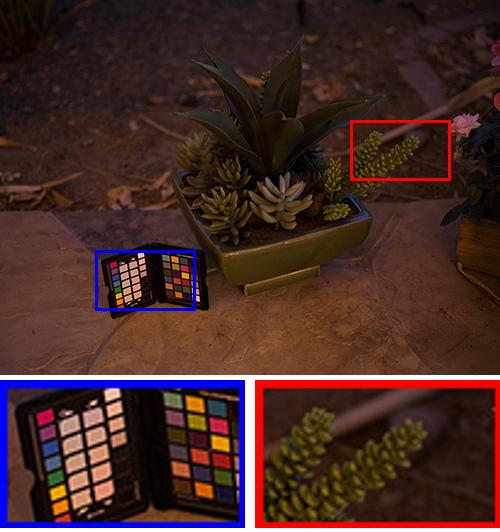}}\hskip.4em
            \subfigure[SNR-Net]{\includegraphics[width=0.3\linewidth]{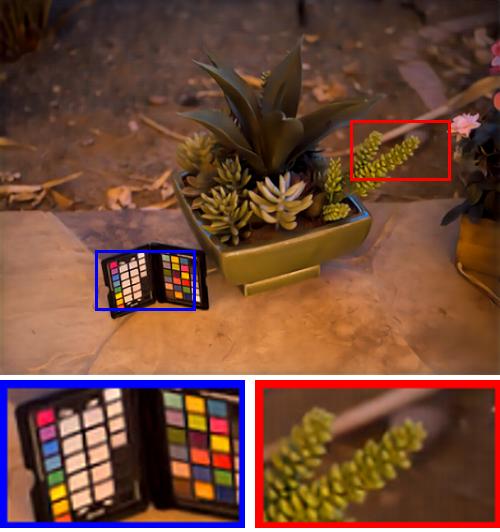}}\hskip.4em
            \subfigure[\textbf{SNR-Net}]{\includegraphics[width=0.3\linewidth]{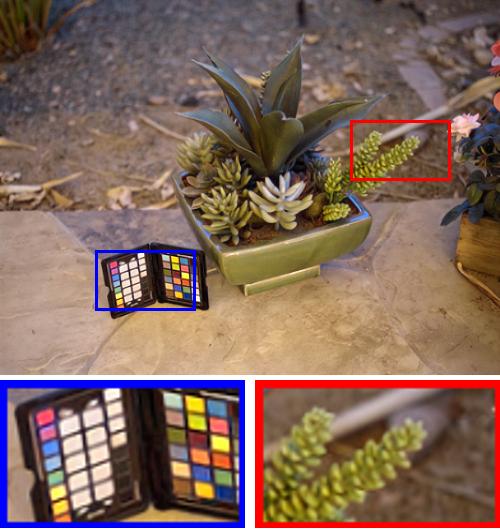}}
    \end{minipage}

    \begin{minipage}[]{0.5\textwidth}
    \hskip -1em
        \centering
             \subfigure[Input]{\includegraphics[width=0.3\linewidth]{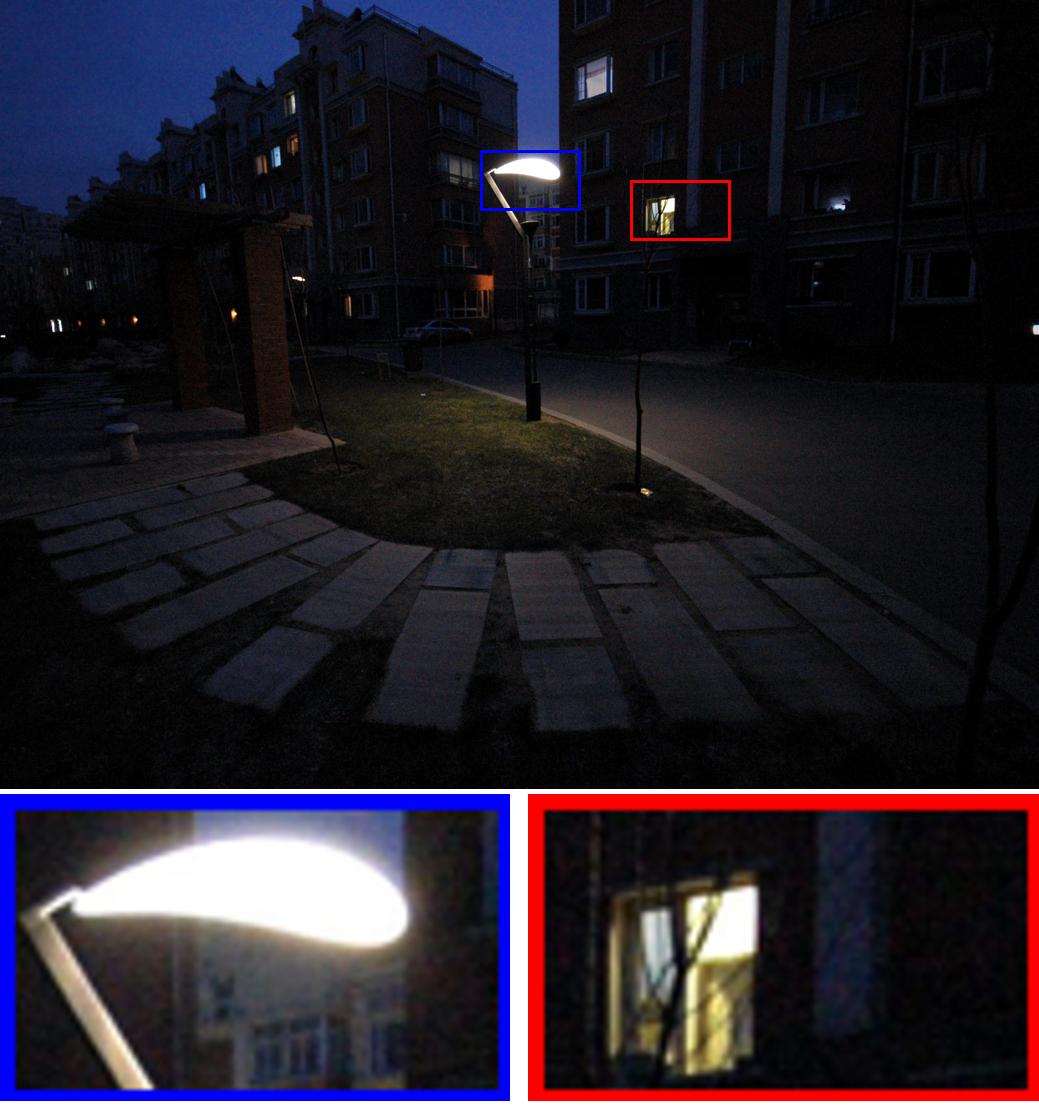}} \hskip.1em
            \subfigure[Retinexformer]{\includegraphics[width=0.3\linewidth]{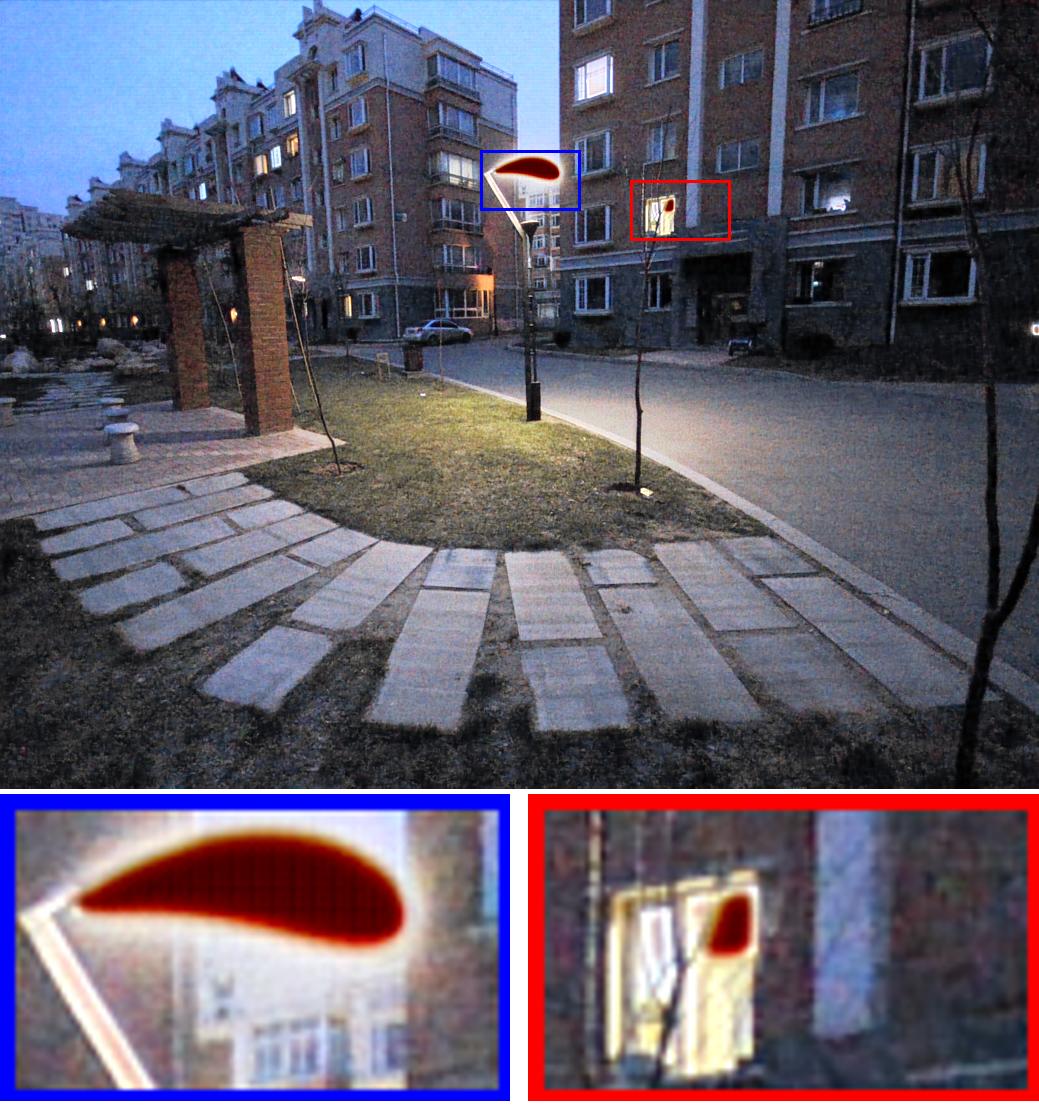}}\hskip.4em
            \subfigure[\textbf{Retinexformer}]{\includegraphics[width=0.3\linewidth]{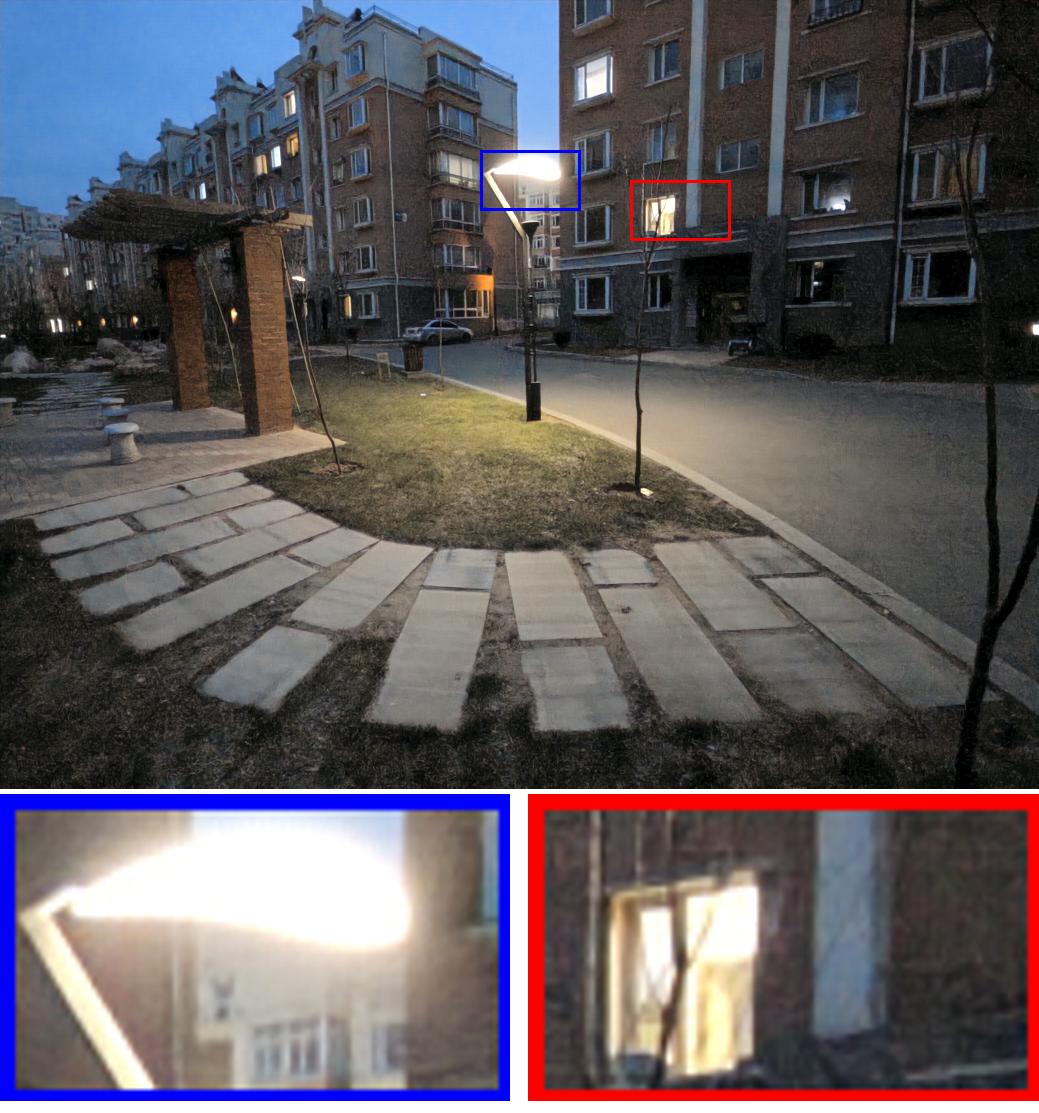}}
    \end{minipage}
    
    \caption{Visual comparisons of recent SOTA LLIE methods, \eg, ~SNR-Net \cite{xu2022snrnet} and Retinexformer \cite{cai2023retinexformer}, trained on small-scale LOL-v2 \cite{yang2019coarse} \textit{versus} our large-scale synthetic data (indicated by \textbf{bold}). Methods only trained on LOL-v2 exhibit issues such as incorrect white balance (top) and abnormal enhancement (bottom). In contrast, \textbf{SNR-Net} and \textbf{Retinexformer} can produce more visually pleasant results.}
    \vspace{-10pt}
    \label{fig:problem}
\end{figure}
\IEEEPARstart{D}{igital} Images captured in insufficiently illuminated environments often suffer from degraded image quality, such as poor visibility, low contrast, and unexpected noise~\cite{ma2023bilevel}. These issues negatively impact both human visual experience and high-level vision tasks~\cite{liu2022learning}. These problems can be addressed with hardware-based methods. For example, extending the exposure time can capture more photons; however, this can lead to blurring if the scene contains movement \cite{cheng2016learning}. Additionally, using a flash for light compensation frequently results in unwanted highlights and uneven lighting, with objects closer to the camera being brightened more than those farther away \cite{petschnigg2004digital}. In contrast to upgrading cameras, which may require additional hardware, Low-Light Image Enhancement (LLIE) techniques are designed to autonomously enhance the visibility of images captured in low-light conditions. LLIE is an active research field related to various system-level applications, such as visual surveillance \cite{yang2019coarse}, autonomous driving \cite{li2021deep}, and computational photography \cite{yuan2007image}.

LLIE is not solely a problem of light adjustment; it also involves addressing issues such as noise burst and color deviation, which are often concealed in darkness due to the limited capabilities of photographic devices \cite{guo2023low}. In the early years, traditional LLIE primarily relied on straightforward illumination amplification such as Gamma correction  \cite{guo2017lime}, Histogram Equalization (HE) \cite{kim1997contrast}, and Retinex-based methods  \cite{land1971lightness}. Although these techniques rely on well-established hand-crafted features, their effectiveness diminishes in such challenging scenarios, leading to unsatisfactory outcomes.  Recent years have witnessed a significant surge in efforts to address the challenges of LLIE through deep learning-based methods~\cite{zhang2019kindpp}. By leveraging the advanced representational capabilities of deep neural networks (DNNs) and the availability of extensive datasets, these contemporary techniques have established new performance benchmarks for LLIE, achieving unprecedented levels of quality and accuracy in some cases compared to traditional methods~\cite{cai2023retinexformer, bai2024retinexmamba}.

Despite notable advances, many DNN-based LLIE methods still struggle to generalize to real-world scenarios \cite{ma2022toward}. As illustrated in Fig.~\ref{fig:problem}, state-of-the-art (SOTA) models trained on small-scale datasets like LOL often exhibit amplified noise, color distortions, and unnatural enhancements. We attribute these issues to insufficiently diverse training data that fail to capture complex low-light conditions and the intricacies of imaging pipelines. This gap underscores the critical need for more effective training data to improve the practicability and generalizability of DNN-based LLIE methods in real-world applications. To mitigate data scarcity, some LLIE methods attempt to combine real-captured and synthetic data \cite{chen2018retinex, yang2021sparse}, yet traditional low-light synthesis pipelines are limited to narrow, sRGB-domain transformations (e.g., gamma correction, Gaussian noise) or manual software adjustments \cite{chen2018retinex, yang2021sparse}. Consequently, they overlook the RAW-to-sRGB degradations introduced by image signal processing (ISP), and their reliance on manual adjustments restricts both scalability and diversity in dataset generation.

In this paper, we introduce a practical, systematic pipeline for synthesizing realistic low-light images by accurately modeling the physics of digital sensors and the complex ISP workflow. Our process begins by collecting high-quality normal-light images, which we unprocess into RAW data using a reverse ISP \cite{brooks2019unprocessing}. We then decrease exposure intensity in the RAW domain to simulate low-light conditions. These synthesized images are subsequently processed through multiple ISP stages—white balance, color space conversion, tone mapping, and gamma correction—each introducing variations that expand the degradation space. This design captures a broad spectrum of real-world degradations while preserving the intricate details of the ISP pipeline, allowing us to generate unlimited paired training data that closely mirrors real-world low-light challenges and  thereby improves the practicability and generalizability of LLIE models. To validate the effectiveness of our low-light synthetic pipeline, we train a simple vanilla U-Net architecture comprising convolutional layers, group normalization, GeLU activation, and convolutional block attention modules (CBAMs). 
Experimental results show that even with this simple design, the LLIE model trained on synthetic data from our pipeline achieves SOTA performance.  Furthermore, given the significant advances in architecture design for LLIE tasks, we also apply our synthetic pipeline to train existing SOTA LLIE models, demonstrating that it not only enhances performance across standard benchmarks but also increases practicality and robustness in real-world applications (see Fig.~\ref{fig:problem}).

Overall, our contributions can be summarized as follows: 
\begin{itemize}
    \item We introduce a general low-light synthetic pipeline that simulates the formation of natural low-light images by modeling key components of the image processing workflow from the RAW domain to the sRGB domain. This enables the generation of unlimited paired data that captures a broad spectrum of low-light degradations, enhancing the realism and diversity of training datasets.

    \item We leverage the unlimited synthetic low-light data generated by our proposed pipeline to train both a simple vanilla U-Net model and existing SOTA LLIE methods, with the goal of improving their practicality and generalizability. In our pipeline, we employ a two-stage training strategy: first, we train the models using the generated data, and then we fine-tune them to better align with the target image styles, addressing biases introduced by pre-trained datasets.

    \item We conduct extensive experiments across multiple datasets to demonstrate the effectiveness of incorporating our low-light synthetic pipeline for training LLIE methods. The results show that our synthetic data significantly improves LLIE methods in both quantitative and qualitative evaluations, even for a simple vanilla U-Net model. Furthermore, we show that our methods bring notable improvements in performance, practicality, and generalizability to existing LLIE methods.

\end{itemize}
The remainder of this paper is organized as follows: Section \ref{sec:related_works} reviews the related literature. Section \ref{sec:proposed_dataset} outlines the proposed low-light image synthesis pipeline. Section \ref{sec:experiment_and_results} details the experimental setups and analyzes the results. Finally, Section \ref{sec:conclusion} concludes this study.

\section{Related Works}
\label{sec:related_works}
In this section, we first introduce existing LLIE datasets, followed by a review of LLIE methods developed over the past decades. We then discuss data synthesis techniques for low-level vision tasks and highlight the significance of in-camera ISP. Finally, we present a comprehensive survey of potential low-light applications.

\begin{table}[t]
    \caption{Summary of commonly used LLIE datasets. In this table, Train indicates whether the dataset is suitable for end-to-end supervised training; Paired signifies the presence of paired images; Task denotes suitability for high-level vision tasks; video suggests that the dataset contains video data. The abbreviations Y, N, and B represent Yes, No, and Both, respectively.}
    \label{tab:dataset_summary}
    \centering
    \setlength\tabcolsep{2pt}
    \begin{threeparttable}
        \begin{tabular}{l|ccccccc}
            \toprule 
            Dataset & Number & Format & Real/Syn & Train & Paired & Task & Video\\
            \midrule 
            LOL-v1 \cite{chen2018retinex} & 1,500 & RGB & Both & Y & Y & N & N \\
            LOL-v2 \cite{yang2021sparse} & 1,789 & RGB  & Both & Y & Y & N & N \\
            SDSD \cite{wang2021sdsd} & 37,500 & RGB  & Real & Y & Y & N & Y \\
            LSRW \cite{hai2023r2rnet} & 5,650 & RGB  & Real & Y & Y & N & N \\
            SID \cite{chen2018learning} & 5,094 & RGB  & Real & Y & Y & N & N \\
            SICE \cite{cai2018learning} & 4,413 & RGB & Real & Y & Y & N & N \\
            VE-LOL-L \cite{liu2021benchmarking} & 2,500 & RGB & Both & Y & Y & N & N \\
            FiveK \cite{bychkovsky2011fivek} & 5,000 & RAW & Real & Y & Y & N & N \\
            SMOID \cite{jiang2019learning} & 179 & RAW & Real & Y & Y & N & Y\\
            \hline
            NPE \cite{wang2013naturalness}  & 8 & RGB  & Real & N & N & N & N \\
            MEF \cite{ma2015perceptual} & 17 & RGB  & Real & N & N & N & N \\
            DICM \cite{lee2013contrast} & 64 & RGB  & Real & N & N & N & N \\
            LIME \cite{guo2017lime} & 10 & RGB  & Real & N & N & N & N \\
            VV\tnote{1} & 24 &  RGB & Real & N & N & N & N \\
            \hline
            ACDC \cite{sakaridis2021acdc} & 4,006 & RGB   & Real & Y & N & Y & N \\
            DarkFace \cite{darkface} & 6,000 & RGB  & Real & Y & N & Y & N \\
            ExDark \cite{loh2019getting} & 7,363 & RGB  & Real & Y & N & Y & N  \\
            VE-LOL-H \cite{liu2021benchmarking} & 10,940 & RGB  & Real & Y & N & Y & N  \\
            \bottomrule
        \end{tabular}
        \begin{tablenotes}
            \item[1] \url{https://sites.google.com/site/vonikakis/datasets}.
        \end{tablenotes}
    \end{threeparttable}
    
\end{table}

\subsection{Benchmark Datasets}
As illustrated in Table \ref{tab:dataset_summary}, existing benchmark data for LLIE can be categorized into paired datasets, primarily used for supervised learning of LLIE methods, and unpaired datasets, which are intended either for testing purposes or for high-level vision tasks with corresponding vision labels.

\noindent\textbf{Paired Datasets}: Supervised LLIE methods rely on paired datasets, but existing ones have limitations. For example, LOL-v1 \cite{chen2018retinex} includes real and synthetic image pairs, with low-light conditions simulated using Adobe Lightroom, which doesn't capture real-world complexity. Datasets like LOL-v2 \cite{yang2021sparse}, VE-LOL \cite{liu2021benchmarking}, SID \cite{chen2018learning}, and SICE \cite{cai2018learning} also face challenges such as small size, reliance on synthetic data, and labor-intensive collection of real images, leading to alignment issues and limited diversity.

\noindent\textbf{Unpaired Datasets}: Many datasets offer under-exposed images mainly for testing purposes, such as NPE \cite{wang2013naturalness}, LIME \cite{guo2017lime}, MEF \cite{ma2015perceptual}, and DICM \cite{lee2013contrast}, but they lack paired normal-light images. Some unpaired datasets focus on high-level vision tasks, like ExDark \cite{loh2019getting} for object detection, ACDC \cite{sakaridis2021acdc} for segmentation, and DarkFace \cite{yang2020advancing} for face detection. However, the absence of corresponding normal-light images or high-level annotations hinders end-to-end optimization of both perceptual enhancement and vision task performance in low-light conditions.

\subsection{Low-Light Image Enhancement Methods}
In the early years, traditional LLIE methods  primarily relied on techniques such as HE and Retinex as ad-hoc solutions. Since then, DNN-based methods have started to dominate this field.

\noindent\textbf{Traditional Methods}: HE-based methods spread out the frequent intensity values of an image to enhance its global contrast. The basic HE approach \cite{pizer1987adaptive} focuses on global adjustment, which can result in poor local illumination and amplified degradation, such as noise, blur, and artifacts. To mitigate these issues, Pizer \etal \cite{pizer1990contrast} performed HE on partitioned regions with local contrast constraints to suppress noise. Ibrahim \etal \cite{ibrahim2007brightness} introduced mean brightness preservation in HE to prevent visual deterioration. Retinex-based methods are grounded in the Retinex theory of color vision \cite{land1971lightness, mertens2007exposure}, which suggests that an image can be decomposed into a reflectance map and an illumination map. Lee \etal \cite{lee2013adaptive} were pioneers in incorporating Retinex theory into image enhancement. Fu \etal \cite{fu2016weighted} designed a weighted variational model, instead of a logarithmic transform, for improved prior modeling and edge preservation. These Retinex-based models heavily rely on handcrafted priors to achieve satisfactory image enhancement. Other approaches, such as dehazing-based and statistical methods, have also been employed for LLIE \cite{zheng2022low}. Dehazing-based methods treat inverted low-light images as a dehazing problem to enhance them \cite{li2015low}. Statistical methods employ statistical models, physical properties, and expert domain knowledge for LLIE \cite{ying2017new}. Traditional LLIE methods, despite incorporating well-designed handcrafted priors and intricate optimization steps, generally exhibit inferior performance compared to the deep learning methods and may suffer from poor inference latency.

\noindent\textbf{Deep Learning-Based Methods}: Most DNN-based LLIE methods rely on supervised learning techniques, which typically use paired image datasets, either synthesized or real-captured~\cite{zheng2022low}. For instance, LLNet \cite{lore2017llnet} employs a stacked sparse denoising autoencoder for multi-scale image enhancement. KinD \cite{zhang2019kindpp} and its extension KinD++ \cite{zhang2021beyond} combine Retinex theory with data-driven methods to handle light adjustment and degradation. Lv \etal \cite{lv2021attention} proposed a pipeline for simulating high-fidelity low-light images, which supports the creation of large-scale paired datasets for low-light enhancement.
However, collecting real-world paired data or developing a synthesis pipeline for realistic low-light scenes is costly and time-consuming. To address this, unsupervised \cite{jiang2019enlightengan}, semi-supervised \cite{yang2021band}, and zero-shot learning \cite{zhang2024zero} techniques reduce dependency on paired data. Unsupervised methods often introduce data bias and struggle with generalization, especially with domain gaps. Zero-shot learning approaches use carefully designed priors and data-free loss functions to mitigate this.
Regarding network architecture, U-shaped networks \cite{zhang2019kindpp} are prevalent in LLIE due to their capacity to preserve high-resolution details and low-resolution semantic features. Recently, transformer-based \cite{cai2023retinexformer} and Mamba-based \cite{bai2024retinexmamba, zhang2024llemamba} approaches have gained attention. However, limited scene and lighting variations, coupled with a lack of high-level annotations, hinder these models' ability to generalize to other low-light vision tasks.

\subsection{Image Signal Processing}

The ISP is responsible for converting RAW sensor data into human-readable sRGB images, performing the critical RAW-to-sRGB mapping~\cite{brooks2019unprocessing, xing2021invertible}. ISPs can be implemented in both hardware and software forms~\cite{mosleh2020hardware}. Hardware ISPs typically consist of proprietary black-box components with limited user-adjustable parameters, confined to a set of registers with specific operational ranges~\cite{sun2024rl}. Traditionally, imaging experts manually adjust these hyperparameters on a small dataset using a combination of visual inspection and image quality metrics, a process that does not necessarily optimize performance for higher-level analytic vision tasks~\cite{wu2019visionisp}. Recently, numerous DNN-based methods have been introduced, outperforming traditional techniques~\cite{xing2021invertible}. High-end smartphones, equipped with GPU and TPU hardware, often employ software ISPs, which are applied to specific steps of the ISP process, such as demosaicing~\cite{tan2017color}, white balance~\cite{afifi2020deep}, or even to replace the entire ISP pipeline~\cite{xing2021invertible}. In this paper, we propose a deep neural network designed to mimic the entire ISP process for low-light image reconstruction. Our novel synthetic pipeline for low-light image degradation accurately replicates the ISP process from the RAW domain to the sRGB domain.

\subsection{Data Synthesis for Low-Level Vision}
Synthetic data generation has emerged as a vital strategy for addressing data scarcity in the training of deep neural networks for low-level enhancement or restoration tasks, such as LLIE \cite{yang2021sparse, lv2021attention}, image denoising \cite{brooks2019unprocessing}, deraining \cite{choi2022synthesized}, and deblurring \cite{rim2022realistic}. By filling gaps left by limited real-world datasets, synthetic datasets help improve both training and performance. However, the practical effectiveness and generalizability of models trained on synthetic data hinge on how closely the synthetic distributions align with real capture conditions \cite{lv2021attention}. Various methods have been proposed to generate low-light images for training. RetinexNet \cite{chen2018retinex} combines synthetic and real-captured pairs, adjusting low-light images in Adobe Lightroom to match the Y-channel distribution of actual low-light images. MBLLEN \cite{lv2018mbllen} applies gamma correction and Poisson noise, while GladNet \cite{wang2018gladnet} uses exposure, vibrance, and contrast adjustments on raw images. AGLLNet \cite{lv2021attention} goes further by applying linear and gamma transformations along with mixed Gaussian-Poisson noise to emulate real low-light conditions. However, these methods frequently rely on extensive manual adjustments, limiting dataset size and realism. Additionally, generating synthetic data solely in the sRGB domain overlooks the complex degradations introduced by ISP, making it more difficult to accurately replicate real-world low-light scenarios.

\begin{figure*}[t]
    \includegraphics[width=1.00\textwidth]{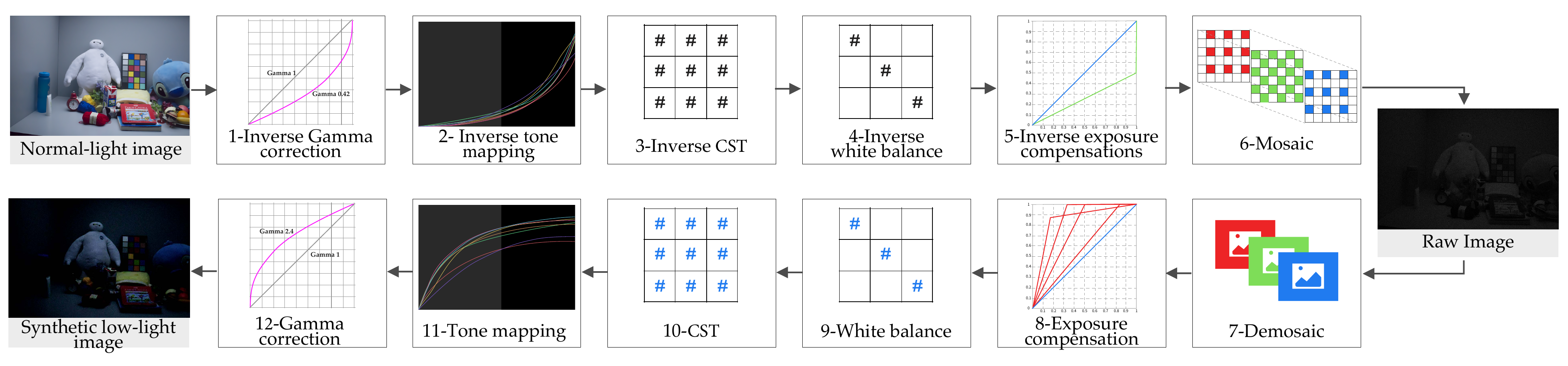}
    \caption{Major components of in-camera image processing pipeline.}
    \label{fig:isp_pipeline}
\end{figure*}

\section{Proposed Low-Light Image Synthesis}
\label{sec:proposed_dataset}
In this section, we present a sophisticated camera simulation pipeline designed to synthesize low-light images from easily collected high-quality normal-light images \cite{brooks2019unprocessing, zheng2020optical, zhang2023practical}. The pipeline begins with the unprocessing\footnote{“Unprocessing” refers to the reverse ISP procedure that converts sRGB images back to their corresponding RAW format.} of sRGB images to obtain their RAW equivalents. Following this, we synthesize low-light degradations in the RAW domain and subsequently process the data through several stages of ISP, which mainly include white balance adjustment, color space conversion, tone mapping, and gamma correction.

\subsection{Preliminaries}

Modern digital imaging systems utilize a ISP to convert RAW sensor data into visually appealing images with reduced noise. The processed image is subsequently saved as an RGB file, such as an sRGB image in JPEG or PNG format, represented by 
\begin{align}
\mathbf{y}^{sRGB} = \operatorname{ISP}(\mathbf{y}^{RAW}, d^{meta}, \beta),
\end{align}
where $\mathbf{y}^{RAW}$ denotes a RAW input, $\mathbf{y}^{sRGB}$ is the corresponding sRGB output, $\operatorname{ISP}$ denotes the camera ISP system, $d^{meta}$ represents the shooting parameters recorded by the camera, and $\beta$ denotes the ISP configurations. Because online images often undergo post-processing or editing, resulting in meta-data being unavailable, our pipeline performs low-light synthesis without relying on it. Typically, the ISP pipeline includes several adjustment stages—such as demosaicing, white balance, tone mapping, and gamma correction—as illustrated in Fig. \ref{fig:isp_pipeline}. In our low-light synthesis pipeline, we introduce variations at each of these stages to broaden the degradation space and enhance the diversity of the low-light data.

\subsection{Brightness Adjustment}

The brightness level of image is primarily determined by two factors: International Organization for Standardization (ISO) gain and exposure time. ISO gain refers to the sensitivity of the camera's sensor to light. Higher ISO values increase sensitivity in low-light conditions but introduce more noise. Exposure time dictates the duration for which the camera collects light from the scene; longer exposures allow more photons to hit the sensor, increasing brightness. In this paper, we linearly adjust the exposure levels to effectively synthesize low-light images \cite{hu2018exposure}. This process simulates the natural accumulation of light, thereby creating realistic low-light conditions in synthesized images. In our pipeline, the low-light RAW image $\mathbf{y}_{lol}$ is obtained by dividing the RAW image $\mathbf{y}^{RAW}$ by an exposure scaling parameter $2^{e}$ across the three color channels, which is defined as:
\begin{align}
    \mathbf{y}_{lol} = \frac{\mathbf{y}^{RAW}}{2^{e}}, \quad e \sim \mathcal{U}(0, 4) - 0.5,
\end{align}
where $\mathcal{U}(\cdot, \cdot)$ is the uniform distribution. To ensure synthetic images accurately reflect the characteristics of real dark photography, we analyze the illumination distribution of low-light images. We use real low-light images from publicly available datasets LOL-v1 \cite{chen2018retinex} and LOL-v2 \cite{chen2018retinex}, convert these images to the YC$_b$C$_r$ color space, and calculate the histogram of the Y channel to determine $e$.

\subsection{Noise Degradation Synthesis}
We then synthesizing the low-light noise inherent in RAW images. Practically, the noise in RAW domain is primarily attributed to two key factors: the randomness in photon arrival, termed as ``shot" noise, and the imperfections in the readout electronics, known as ``read'' noise \cite{brooks2019unprocessing}. Shot noise follows a Poisson distribution, with its mean representing the actual light intensity, expressed in photoelectrons. Conversely, read noise is generally modeled by a Gaussian distribution with a mean of zero and a constant variance. 

As suggested by Brooks \etal~\cite{brooks2019unprocessing}, the Poisson-Gaussian noise model can be approximated using a single heteroscedastic Gaussian, defined as
\begin{align}
\mathbf{y}_n & \sim  \mathcal{N}\left(\mu_s = \mathbf{y}_{lol}, \sigma_s^2 = \lambda_r + \lambda_s \mathbf{y}_{lol}\right),
\end{align}
where denotes a uniform distribution and $\mathcal{N}(\mu, \sigma)$
denotes a Gaussian distribution with mean $\mu$ and standard deviation $\sigma$; $\lambda_r$ and $\lambda_s$ represent the analog and digital gains for camera sensor, respectively. In our pipeline, the noise parameter sampling procedure is defined as:
\begin{align}
& \log \left(\lambda_s\right) \sim \mathcal{U}\left(\log(\lambda_s^{(min)}), \log(\lambda_s^{(max)})\right), \nonumber\\
& \log \left(\lambda_r\right) \mid \log \left(\lambda_s\right)\sim\mathcal{N}\left(a_r\log \left(\lambda_s\right)+b_r, \hat{\sigma}_r\right).
\end{align}
where $\lambda_s^{(min)}$ and $\lambda_s^{(max)}$ are the estimated overall system gains at the minimum and maximum ISO of a camera
respectively; $a_r$ and $b_r$ indicate the fitted line’s slope and intercept respectively; $\hat{\sigma}^2_r$ is an unbiased estimator of standard
deviation of the linear regression under a given Gaussian error assumption \cite{zhang2023towards}.

In previous low-light synthesis pipelines \cite{chen2018retinex, yang2021sparse, lv2021attention}, noise is commonly simulated in the sRGB domain. This approach often overlooks the variations in noise distribution that arise from the sensor data processing through various ISP stages. Our low-light noise synthesis pipeline addresses this gap by accounting for the comprehensive re-distributions caused by the varying configurations of ISP parameters. This ensures a more accurate representation of real-world noise characteristics in the synthesized low-light images.

\subsection{Demosaicing} 
Camera sensors capture only the intensity of incoming photons, lacking the ability to differentiate between colors. To encode color information, these sensors are overlaid with color filter arrays, typically arranged in patterns such as the R-G-G-B Bayer pattern, where the red and blue channels retain 25\% of the pixels, and the green channel retains 50\%. The Demosaicing block processes Bayer format images, converting them to RGB by interpolating missing pixel values. In our implementation, we adopt high-quality linear interpolation~\cite{malvar2004demasaicing} for demosaicing, which enhances the bilinear interpolation accuracy by leveraging both adjacent pixels within the same channel and data from neighboring channels. This process produces the demosaiced image, denoted as $\mathbf{y}_{d}$.

\subsection{Color Correction}
Images captured by digital cameras often exhibit color inaccuracies resulting from the spectral sensitivity of the sensor and differing lighting conditions. To mitigate these discrepancies, color correction methods such as white balance and color matrix transformations are applied. 

\subsubsection{White Balance}  
Digital cameras often encounter challenges in accurately rendering colors under varying low-light conditions, where the ambient light's color temperature can introduce unwanted blue, orange, or green color casts in the final images. To address this, cameras apply white balance adjustments designed to produce images with colors that appear natural and visually pleasing under typical lighting conditions. During this adjustment process, the red, green, and blue color channels are multiplied by corresponding white-balance diagonal matrices $W_c$ (where $c$ represents R, G, or B). These matrices are derived from the RAW image data according to the sensor's response to illumination, \ie, calibrating the color of a ``true" white patch under the given illumination. However, due to the difficulties in precise light metering in low-light environments, these weights $W_c$ are often biased and require further calibration. To improve the white balance capabilities of LLIE methods, we introduce variability in the white-balance
diagonal matrix $W_c$ to alternate the image color style, generating a diverse set of images with intentional white balance inaccuracies. In mathematics sense, this
stage is formalized as 
\begin{align}
    \mathbf{y}_{wb} &= \mathbf{y}_{d} \circ W_c, \nonumber\\
    W_c &= [[[w_r; w_g; w_b]]]_{1\times1\times3},
\end{align}
where $\circ$ means pixel-wise product. In our pipeline, we empirically set $w_r$ and $w_g$ as $ 1.2 + 2\mathcal{U}(0,1)$ and $ w_b = 1$. To prevent the dimming of saturated pixels, we apply a highlight-preserving transformation that smoothly adjusts the gains near white \cite{brooks2019unprocessing, afifi2020deep}. 

\subsubsection{Color Space Transform} 
The Color Space Transform (CST) utilizes a $3 \times 3$ color correction matrix (CCM) $C_{l}$ to map the white-balanced raw RGB values to the linear CIE XYZ color space, taking into account the interdependence between color channels to ensure accurate color reproduction. Camera profiles typically include the $C_{l}$ matrix to manage how colors are interpreted by specific camera sensors. In our pipeline, we employ 11 predefined camera profiles from popular brands such as Canon, Huawei, Nikon, and Olympus to control the rendering of color and tonality. Specifically, for a white-balanced RAW image $\mathbf{y}_{wb}$, the color space transformed image $\mathbf{y}_{cst}$ in the CIE XYZ color space is obtained as follows:
\begin{align}
    \mathbf{y}_{cst} = \mathbf{y}_{wb} \otimes C_{l}.
\end{align}
where $\otimes$ denotes matrix multiplication. This procedure, which converts $\mathbf{y}_{wb}$ into $\mathbf{y}_{cst}$, is referred to as the \textit{linear process}. In practice, camera profiles provide two CCM matrices ($C_{l1}$ and $C_{l2}$) calibrated for two different illuminations (Correlated Color Temperature (CCT) 2500K and 6500K). Depending on the white-balance temperature, the corresponding CCM matrix is used. In our pipeline, the CCM matrix $C_{l}$ is a weighted blending of the two factory pre-calibrated matrices $C_{l1}$ and $T_{l2}$, defined as:
\begin{align}
    C_{l} &= gC_{l1} + (1-g)C_{l2},  \qquad g \sim \mathcal{U}(0,1).
\end{align}
In this way, we generate images with varying color temperatures, expanding the color degradation space and enhancing the model's ability to perform color correction.

\subsection{Color Manipulation}
To enhance human perception of images, nonlinear procedures such as tone mapping and Gamma correction are employed \cite{huang2022towards}. 

\subsubsection{Tone Mapping Curves}
In the ISP pipeline, tone mapping applies a nonlinear transfer function to transform input pixel intensities from linear space to output intensities in the logarithmic domain. This process rearranges the pixel intensity distribution, adjusting local contrast, brightness, and color while preserving an aesthetically pleasing appearance. In practice, this operation is often efficiently implemented in hardware using a 1D lookup table (1D LUT). In our pipeline, we utilize global tone mapping, where each pixel value is mapped independently of its location. Specifically, given a predefined tone curve $T = \{\mathbf{t}_{c}\}_{c=1}^3$, consisting of three 1D LUTs for sRGB images, where each LUT has $L$ entries, $\mathbf{t}_c \in \mathbb{R}^L$, each entry maps an input pixel intensity to an enhanced output intensity. We apply these predefined tone curves to the input image $\mathbf{y}_{cst}$ using bilinear interpolation, ensuring a smooth, artifact-free tone-mapped result, defined as:
\begin{align}
    \mathbf{y}_t = \operatorname{Interp}(\mathbf{y}_{cst}, T),
\end{align}
where $\operatorname{Interp}$ computes the value of a target pixel by using the values of its neighboring pixels. To represent a tone curve as a vector, consider $n$ uniformly spaced input tonal values in the domain $[0,1]$, denoted as $\mathbf{b} = [0, 1/(n-1), 2/(n-1), \ldots, 1]^{\top}$. Evaluating the tone curve $T$ at these input values results in the $n$-vector $\mathbf{t} = [T(0), T(1/(n-1)), T(2/(n-1)), \ldots, T(1)]^{\top}$. Here, the superscript ${}^{\top}$ represents the transpose operation. The $i$-th component of $\mathbf{t}$, denoted $t_i$, corresponds to the tone curve value at the $i$-th input tonal value. By employing a suitable interpolation scheme, the vector pair $\mathbf{b}$ and $\mathbf{t}$ fully defines the tone curve. Unlike previous methods that rely on fixed tone curves \cite{brooks2019unprocessing}, our synthetic pipeline utilizes 200 distinct tone curves to adjust input pixel intensities, enabling a diverse range of color appearances.

\begin{figure}[t]
 \hskip -1em
\centering
\subfigure[\scriptsize LOL-v1]{
\begin{minipage}[b]{0.23\textwidth}
\includegraphics[width=\linewidth]{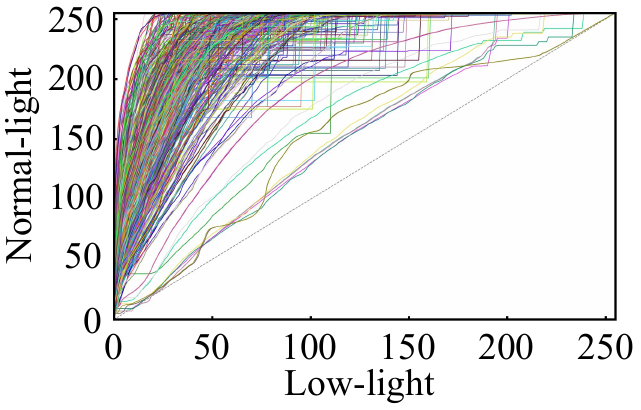}\vspace{0.3pt}
\end{minipage}}\hspace{-4pt}
\subfigure[\scriptsize LOL-v2]{
\begin{minipage}[b]{0.23\textwidth}
\includegraphics[width=\linewidth]{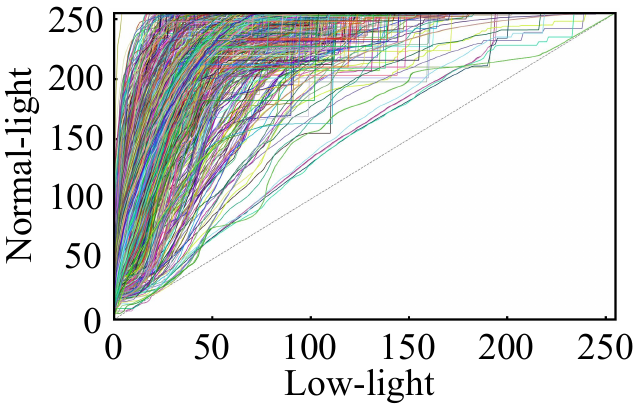}\vspace{0.3pt}
\end{minipage}}

\vspace{-0.3cm}
\hspace*{-1em}\subfigure[\scriptsize AGLLNet's]{
\begin{minipage}[b]{0.23\textwidth}
 \includegraphics[width=\linewidth]{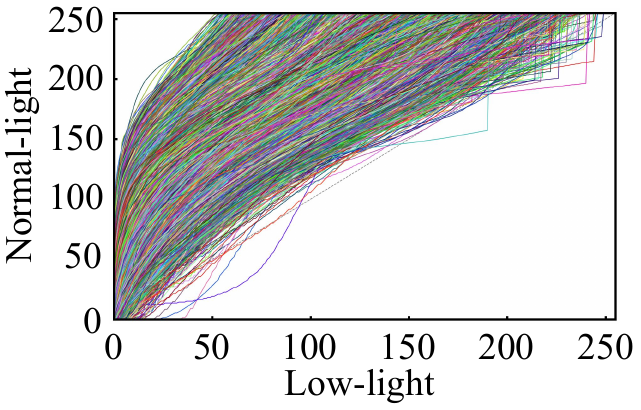}\vspace{0.3pt}
\end{minipage}}\hspace{-4pt}
\subfigure[\scriptsize Ours]{
\begin{minipage}[b]{0.23\textwidth}
  \includegraphics[width=\linewidth]{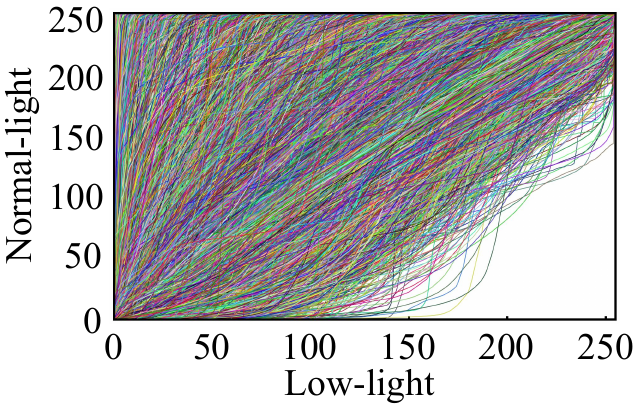}\vspace{0.3pt}
\end{minipage}}

\caption{Comparison of the low-light coverage across different datasets using the exposure adjustment curves, which map the luminance histogram of the low-light images to that of their corresponding normal-light ground truth. A steeper curve suggests a higher degree of underexposure.}
\label{fig:curve}
\end{figure}

\subsubsection{Gamma Correction}
Images that are not Gamma corrected tend to exhibit harsh contrast, with significant loss of detail in both shadow and highlight areas. This issue occurs because rendering calculations are performed in a linear color space, which differs from how our eyes naturally perceive colors. To ensure that these linear images are visually appropriate for human perception, a correction is necessary \cite{zhang2023practical}. This process, known as gamma correction, adjusts linear intensities to nonlinear ones that align with our visual perception, defined as follows:
\begin{align}
   \mathbf{y}^{sRGB}=\left\{
\begin{aligned}
&12.92 \mathbf{y}_{t}, &  \mathbf{y}_{t} \leq 0.0031308\\
&(1+a)(\mathbf{y}_{t})^{1/\gamma} -a, &\mathbf{y}_{t} > 0.0031308 \\
\end{aligned}
\right.
\end{align}
where the gamma value and $a$ are empirically set to 2.4 and 0.055, respectively.

\subsection{Dataset Characteristics}
 Compared to existing datasets for low-light image enhancement, our synthetic dataset offers the following two advantages:

\begin{itemize}
\item \textbf{Practical low-light degradation synthesis.} Traditional methods adjust illumination using Gamma transformation and add noise, like AWGN and Poisson, in the sRGB domain \cite{chen2018retinex, lv2021attention, zhu2024temporally}. In contrast, our pipeline synthesizes low-light illuminance and noise directly in the RAW domain for a more accurate representation of natural low-light conditions. Additionally, we vary white balance, Gamma correction, and tone mapping to simulate different camera ISP pipelines, generating synthetic images with a wider range of low-light degradations.

\item \textbf{Unlimited dataset size.} The size of our synthetic dataset surpasses that of any existing LLIE datasets, which typically depend on manual image collection or editing, posing challenges for large-scale data acquisition. Our data generation method, however, can synthesize paired low-light and normal-light images in large quantities across diverse scenes. The vast scale of our dataset also eliminates the need to restrict input to only high-quality, normally exposed images.
\end{itemize}
Fig. \ref{fig:curve} compares the low-light range of our synthetic dataset with existing low-light datasets, including LOL-v1 \cite{chen2018retinex}, LOL-v2 \cite{yang2021sparse}, and the synthetic dataset used for training AGLLNet \cite{lv2021attention} (denoted as AGLLNet's). The shape of the curves represents the degree of underexposure, while their coverage reflects the diversity of lighting conditions. Our synthetic dataset covers a broader range of underexposure levels, enhancing the practicality and generalizability of SOTA LLIE methods.

\section{Experiment and Results}
\label{sec:experiment_and_results}
We conduct comprehensive experiments utilizing the extensive dataset of low-/normal-light pairs synthesized through our proposed pipeline. These experiments employ a vanilla U-Net model to evaluate the effectiveness of our approach, encompassing both perceptual assessments and high-level vision tasks.

\begin{figure}[t]
    \centering
    \includegraphics[width=1.0\linewidth]{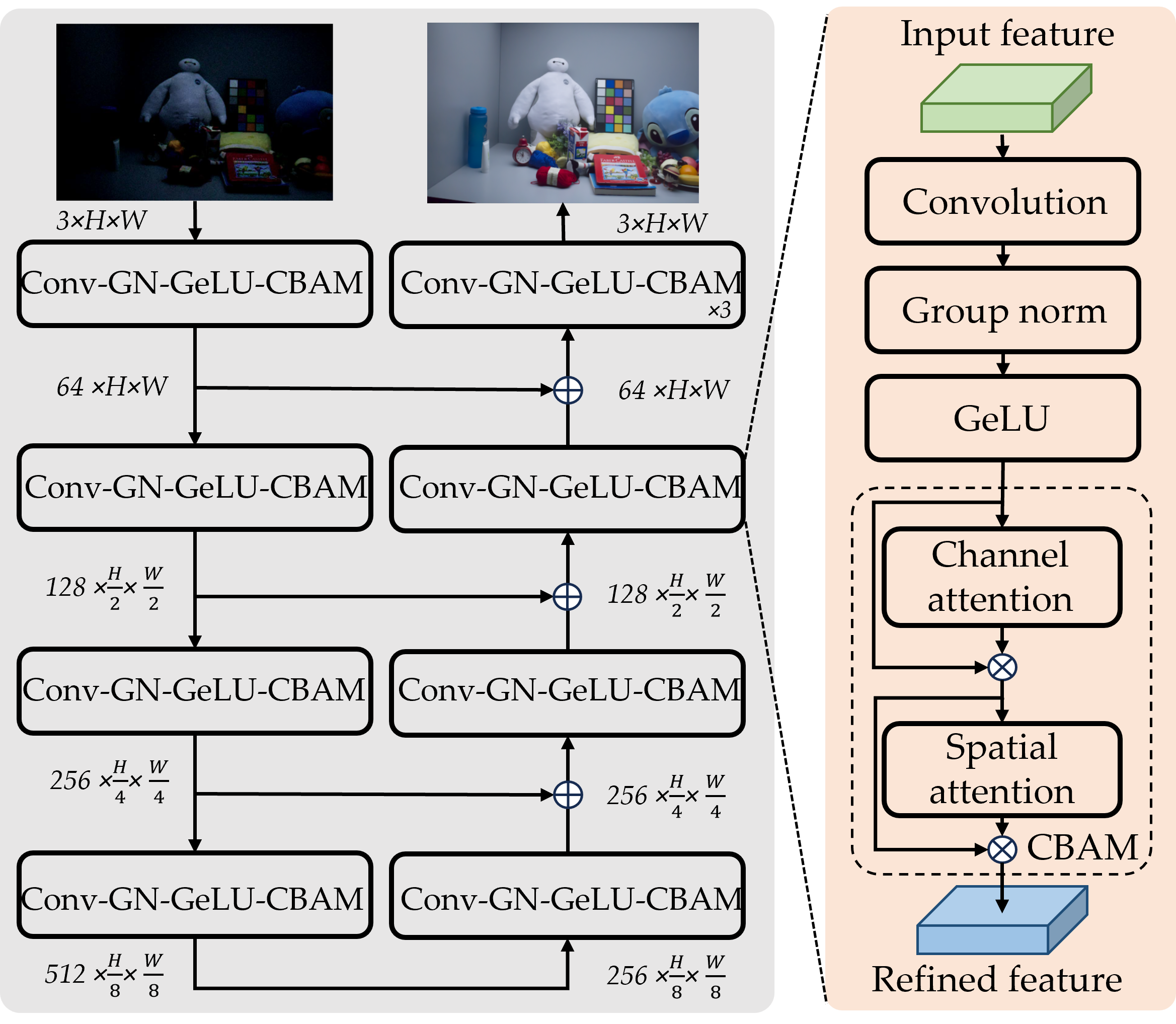}
    \caption{The overview of the vanilla U-Net, where Conv-GN-GeLU-CBAM represents a sequential combination of convolution, group normalization, GeLU activation function, and the CBAM block. $\oplus$ and $\otimes$ denote tensor addition and multiplication, respectively.} 
    \label{fig:network}
\end{figure}

\subsection{Models}
As illustrated in Fig. \ref{fig:network}, we adopt a vanilla U-Net framework \cite{ronneberger2015unet} in our experiments for its efficient feature learning via the encoder-decoder structure. The U-Net model incorporates convolutional layers, group normalization, GeLU activation, and CBAMs \cite{woo2018cbam}, which enhance features by generating attention maps across both channel and spatial dimensions. To evaluate the generalizability of our synthetic pipeline, we also train SOTA models, including SNR-Net \cite{xu2022snrnet}, RetinexFormer \cite{cai2023retinexformer}, and RetinexMama \cite{bai2024retinexmamba}.

\begin{table*}[t]
\caption{Quantitative comparisons of LLIE models on LOL-v1 \cite{chen2018retinex} and LOL-v2 \cite{yang2021sparse}. The table is organized as follows: the top section displays GT images; the second section lists LLIE methods not trained on LOL-v1/LOL-v2; and the third section presents supervised LLIE methods trained or fine-tuned on LOL-v1/LOL-v2.  The vanilla U-Net method is highlighted with a \colorbox{gray!20}{gray background}. FT indicates that vanilla U-Net first trained on synthetic data and then fine-tuned on LOL-v1 or LOL-v2. $\uparrow$ ($\downarrow$) denotes that higher (lower) values indicate better quality. \textbf{The best result in each section is marked in \textcolor{red}{red}}.}
\label{tab:quantitative_comparision}
\centering
\begin{tabular}{l|cccc|cccc|c}
\toprule
\textbf{LOL-v1 \cite{chen2018retinex}} & PSNR$\uparrow$   & SSIM$\uparrow$  & LPIPS$\downarrow$  & DISTS$\downarrow$ & MUSIQ$\uparrow$  & LIQE$\uparrow$ & CLIPIQA$\uparrow$ & Q-Align$\uparrow$ & User Study\\ 
\midrule
GT & ---& ---& ---& --- & 72.204 & 4.731 & 0.687 & 80.338 & 3.913\\
\hline
Zero-DCE \cite{guo2020zero}                & 14.861 & 0.559 & 0.335 & 0.216 & 55.854 & 2.468 & 0.587 & 47.862 & 2.512\\
EnlightenGAN \cite{jiang2019enlightengan}  & 15.308 & 0.531 & 0.489 & 0.196 & 56.604 & 2.420 & 0.481 & 50.607 & 2.324 \\   
RUAS   \cite{liu2021ruas}                  & 16.405 & 0.500 & 0.427 & 0.189 & 59.345 & 2.584 & 0.481 & 58.539 & 2.314\\
SCI  \cite{ma2022toward}                   & 17.784 & 0.522 & 0.339 & 0.211& 57.118 & 2.679 & 0.578 & 54.157 & 2.296\\
PairLIE   \cite{fu2023pairlie}            & 18.468 & 0.748 & 0.244 & 0.182& 56.861 & 2.191 & 0.460 & 43.045 & 2.212\\
NeRCo      \cite{yang2023nerco}            & \textcolor{red}{19.411} & 0.668 & 0.322 & 0.358& 68.969 & 3.657 & 0.502 & 68.963  & 3.107\\
% GDP \cite{fei2023gdp} \\
CLIP-LIT \cite{liang2023cliplit}  &12.394  &0.493  & 0.382 &0.247  & 57.101 & 1.615 & 0.534 &37.632 & 1.879 \\
AGLLNet \cite{lv2021attention} & 17.527 & 0.733 & 0.210 & 0.138 & 66.395 & 3.287 & 0.561 &65.793 & 3.079\\
\rowcolor{gray!20} U-Net & 18.678 &	\textcolor{red}{0.801} & \textcolor{red}{0.172} &	\textcolor{red}{0.122} &	\textcolor{red}{72.043} &	\textcolor{red}{4.394} &	\textcolor{red}{0.653} &	\textcolor{red}{76.480} & 
\textcolor{red}{3.452}\\
\hline

RQ-LLIE \cite{liu2023rqllie} & \textcolor{red}{25.241} & 0.855 & 0.121 & 0.112 & 70.598 & 3.754 & 0.544 &71.186 & 3.625\\
MBLLEN  \cite{lv2018mbllen}                & 17.563 & 0.734 & 0.174 & 0.147 & 69.287 & 3.648 & 0.520 & 69.516 & 2.765\\ 
Retinex-Net \cite{chen2018retinex}         & 14.977 & 0.330 & 0.624 & 0.397 & 57.254 & 1.457 & 0.502 & 32.514  & 1.474\\
KinD++  \cite{zhang2019kindpp}           & 15.724 & 0.637 & 0.374 & 0.350 & 70.539 & 3.634 & 0.562 & 70.407 & 2.639\\
MIRNet  \cite{zamir2020mirnet}             & 24.138 & 0.845 & 0.131 & 0.129 & 67.278 & 3.117 & 0.518 & 64.456 & 3.283\\
SNR-Net \cite{xu2022snrnet}                & 22.593  & 0.832 &0.156 & 0.138& 64.979 & 2.897& 0.500& 61.903 & 3.255\\  
IAT  \cite{cui2022iat}                     & 23.272 & 0.808 & 0.216 & 0.161& 55.453 & 2.434 & 0.483 & 48.740 & 2.655\\
LLFlow \cite{wang2022llflow}               & 21.133 & 0.854 & 0.119 & 0.118 & \textcolor{red}{74.142} & 4.571 & 0.599 & 78.973 & 4.109 \\
Retinexformer \cite{cai2023retinexformer}  & 23.622 & 0.828 & 0.148 & 0.143 & 62.910 & 2.914 & 0.523 & 57.413 & 3.089\\
RetinexMamba  \cite{bai2024retinexmamba}   & 23.527  & 0.829 & 0.143 & 0.139 & 63.568 & 3.038 & 0.527 & 58.933  & 3.123\\
LLFormer \cite{wang2023llformer} & 23.649 & 0.819 &0.169& 0.149&  60.764& 2.603 & 0.481 & 53.558  & 3.288\\
HVI-CIDNet \cite{yan2025hvi} & 23.809 & 0.857 & 0.086 & 0.078 & 71.907 & 4.385 & 0.608 & 77.244 & 4.075 \\
\rowcolor{gray!20} U-Net (FT)  &23.910 &\textcolor{red}{0.861}  &\textcolor{red}{0.085} &\textcolor{red}{0.067}  &73.285  &\textcolor{red}{4.586}  &\textcolor{red}{0.654}  & \textcolor{red}{80.260} &  \textcolor{red}{4.153}
\\

\hline \hline
\textbf{LOL-v2 \cite{yang2021sparse}} & PSNR$\uparrow$  & SSIM$\uparrow$  & LPIPS$\downarrow$ & DISTS$\downarrow$ & MUSIQ$\uparrow$ & LIQE$\uparrow$ & CLIPIQA+$\uparrow$  & Q-Align$\uparrow$ & User Study\\ 
\midrule
GT & ---& ---& ---& ---& 68.568 & 4.298 & 0.649 & 75.218 & 3.610\\
\hline
Zero-DCE      & 18.059 & 0.574 & 0.313 & 0.220 & 53.974 & 2.500 & 0.559 & 49.168 & 2.311\\
EnlightenGAN         & 18.640 & 0.676 & 0.309 & 0.192 & 53.646 & 2.424 & 0.482 & 52.879 & 2.098\\   
RUAS          & 15.325 & 0.488 & 0.310 & 0.215 & 56.177 & 2.540 & 0.450 & 56.664 & 2.127\\
SCI           & 17.300 & 0.533 & 0.308 & 0.202 & 54.443 & 2.762 & 0.544 & 55.031 & 2.211\\
PairLIE       & 19.880 & 0.777 & 0.234 & 0.191 & 54.935 & 2.056 & 0.450 & 40.794 & 1.940\\
NeRCo         & 17.423 & 0.630 & 0.419 & 0.403 & 47.904 & 1.245 & 0.381 & 34.018 & 1.528 \\ 
CLIP-LIT & 15.182 & 0.529 & 0.369 & 0.227 & 55.304 & 1.894 & 0.525 &40.363 &1.754 \\
AGLLNet & \textcolor{red}{20.693} & 0.758 &0.199& 0.150&  65.217 & 3.404 & 0.545 & 66.045 & 2.812\\
\rowcolor{gray!20} U-Net & 20.289 &	\textcolor{red}{0.822} &	\textcolor{red}{0.152} &	\textcolor{red}{0.118} &	\textcolor{red}{68.857} &	\textcolor{red}{3.914} & \textcolor{red}{0.619}	& \textcolor{red}{72.470} &
\textcolor{red}{3.346}\\
\hline
RQ-LLIE & 22.371 & 0.854 & 0.142 & 0.125 & 66.861 & 3.555 & 0.496 & 67.919 & 3.305\\
MBLLEN        & 17.297 & 0.685 & 0.222 & 0.193 & 66.767 & 3.648 & 0.514 & 66.959 & 2.584\\
Retinex-Net   & 16.097 & 0.401 & 0.543 & 0.413 & 56.672 & 1.664 & 0.489 & 30.487 & 1.249\\
KinD++        & 17.660 & 0.770 & 0.217 & 0.382 & 68.061 & 3.694 & 0.558 & 68.327 & 2.446\\
MIRNet        & 21.643 & 0.813 & 0.268 & 0.187 & 51.028 & 1.831 & 0.454 & 41.174 & 2.303\\
SNR-Net       & 21.229 & 0.850 &0.156 & 0.131 & 63.873& 2.800 & 0.488 & 62.208 & 3.106\\  
IAT           & 23.498 & 0.823 & 0.223 & 0.384 & 47.374 & 1.670 & 0.400 & 42.486 & 2.173\\
LLFlow        & 20.236 & 0.832 & 0.176 & 0.146 & \textcolor{red}{70.870} & 3.948 & 0.544 & 71.820 & 3.375\\ 
Retinexformer  & 22.236 & 0.849 &0.161 & 0.149 & 61.209 & 2.920 & 0.490 & 59.319 & 2.924\\
RetinexMamba   & 22.010 & 0.841 &0.159& 0.148&  61.402 & 2.998 & 0.489 & 58.086 & 2.880\\
LLFormer &  21.731& 0.821 &0.195&0.174 & 54.539  &2.297  &0.444  &47.744 & 3.038\\
HVI-CIDNet \cite{yan2025hvi} & \textcolor{red}{24.111} & 0.868 & 0.116 & 0.101 & 67.655 & 3.912 & 0.569 & 69.290 & 3.360 \\
\rowcolor{gray!20} U-Net (FT)  &23.096	&\textcolor{red}{0.880}	& \textcolor{red}{0.096} & 	\textcolor{red}{0.085}	 &  70.342	& \textcolor{red}{4.108}	  & \textcolor{red}{0.620} & 	\textcolor{red}{72.296}  & 
\textcolor{red}{3.526}
\\
\bottomrule 
\end{tabular}
\end{table*}

\subsection{Datasets}
\noindent\textbf{Datasets for Low-light Synthesis.} We utilize a combination of the ImageNet \cite{deng2009imagenet} and COCO \cite{lin2014microsoft} datasets, comprising nearly ten million images, as normal-light references for synthesizing low-light images. To improve simulation efficiency and image quality, we first resize and downscale the images by a factor of two, considering the varying quality and high resolutions within these datasets. Subsequently, we randomly crop image patches of size $156 \times 156 \times 3$ for simulation.
\begin{table}[t]
\caption{Correlations between model predictions and human-rated MOSs on LOL-v1 \cite{chen2018retinex}, LOL-v2 \cite{yang2021sparse}, and LIEQ \cite{zhang2021no}. The best result in each section is highlighted in \textcolor{red}{red}, while the second-best result is shown in \textcolor{blue}{blue}.}
\setlength\tabcolsep{1.5pt}
\centering
\label{tab:correlation}
\begin{tabular}{l|ccc|ccc}
\toprule
Dataset & \multicolumn{3}{c|}{LOL-v1 \& LOL-v2} & \multicolumn{3}{c}{LIEQ \cite{zhang2021no}} \\
\hline
Metric & SRCC $\uparrow$ & PLCC $\uparrow$  & KRCC $\uparrow$ & SRCC $\uparrow$ & PLCC $\uparrow$  & KRCC $\uparrow$\\ 
\hline
PSNR & 0.503 & 0.420 & 0.349 & 0.378 &   0.387 &   0.254\\
SSIM \cite{wang2004ssim} & 0.688 & 0.743 & 0.505 & 0.583 &   0.585 &   0.403\\
FSIM \cite{zhang2011fsim} & 0.630 & 0.600 & 0.454 & 0.693 &   0.680 &   0.499 \\
VIF \cite{sheikh2006vif} & 0.408 & 0.424 & 0.285 & 0.602 &   0.619 &   0.431\\
VSI \cite{zhang2014vsi} & 0.559 & 0.531 & 0.394 & 0.669 &   0.643 &   0.477\\
GMSD \cite{xue2013gmsd} & 0.560 & 0.539 & 0.399 & 0.647 &   0.655 &  0.458\\
PieAPP \cite{prashnani2018pieapp} & 0.459 & 0.457 & 0.318 & 0.595 &   0.608 &  0.414\\
LPIPS \cite{zhang2018lpips} & \textcolor{red}{0.819} & \textcolor{red}{0.840} & \textcolor{red}{0.630} & \textcolor{red}{0.847} &   \textcolor{red}{0.817} &  \textcolor{red}{0.651} \\
DISTS \cite{ding2020dists} & \textcolor{blue}{0.807} & \textcolor{blue}{0.801} & \textcolor{blue}{0.619} & \textcolor{blue}{0.844} &   \textcolor{blue}{0.812} &  \textcolor{blue}{0.645}\\
LOE \cite{guo2016lime} & 0.513 & 0.457 & 0.359 & 0.427 & 0.423 & 0.295\\
CIELAB \cite{robertson1977cielab} & 0.596 & 0.648 & 0.424 & 0.631 & 0.622 & 0.441\\
\hline
BRISQUE \cite{mittal2012brisque} & 0.066 & 0.290 & 0.042 & 0.492 &   0.637 &  0.341\\
NIQE \cite{mittal2012making} & 0.331 & 0.541 & 0.220 & 0.603 &   0.631 &  0.432 \\
PI \cite{blau20182018} & 0.256 & 0.257 & 0.173 & 0.484 &   0.494 &  0.335\\
MUSIQ \cite{ke2021musiq} & 0.672 & 0.669 & 0.480 & 0.743 &   0.723 &   0.546\\
LIQE \cite{zhang2023liqe} & \textcolor{blue}{0.723} & \textcolor{blue}{0.729} & \textcolor{blue}{0.529} & \textcolor{blue}{0.827} &   \textcolor{blue}{0.818} &   \textcolor{blue}{0.636}\\
CLIPIQA \cite{wang2022clipiqa} & 0.451 & 0.465 & 0.312 & 0.754 &   0.754 &   0.556\\
Q-Align \cite{wu2023qalign} & \textcolor{red}{0.753} & \textcolor{red}{0.781} & \textcolor{red}{0.560} & \textcolor{red}{0.888} & \textcolor{red}{0.896} & \textcolor{red}{0.708}\\
\bottomrule
\end{tabular}
\end{table}

\begin{table*}[t]
\caption{Quantitative comparisons conducted on the unpaired NPE \cite{wang2013naturalness}, MEF \cite{ma2015perceptual}, DICM \cite{lee2013contrast}, LIME \cite{guo2017lime} and VV datasets to evaluate the generalizability of LLIE models. The table is organized as follows: unsupervised LLIE methods (the top section) and supervised LLIE methods (the second section). $\uparrow$ ($\downarrow$) indicates that larger (smaller) values correspond to better quality. The best result is in \textcolor{red}{red} color whereas the second best one is in \textcolor{blue}{blue} color. }
\centering
\label{tab:generalization_comparision}
\begin{threeparttable}
\begin{tabular}{l|cc|cc|cc|cc|cc}
\toprule
 Dataset & \multicolumn{2}{c|}{NPE \cite{wang2013naturalness}} & \multicolumn{2}{c|}{MEF \cite{ma2015perceptual}} & \multicolumn{2}{c|}{DICM\cite{lee2013contrast}} & \multicolumn{2}{c|}{LIME \cite{guo2017lime}} & \multicolumn{2}{c}{VV\tnote{1}} \\ 
 \hline
Metric & Q-Align$\uparrow$ & LIQE$\uparrow$  & Q-Align$\uparrow$ & LIQE$\uparrow$  & Q-Align$\uparrow$ & LIQE$\uparrow$  & Q-Align$\uparrow$ & LIQE$\uparrow$ & Q-Align$\uparrow$ & LIQE$\uparrow$\\ 
\midrule
Zero-DCE  \cite{guo2020zero}  & 59.132 & 3.842 & 65.864 & 3.786 & 60.140 & 3.093 &  55.220 & \textcolor{blue}{3.062} & 64.719 & 1.227
   \\
EnlightenGAN \cite{jiang2019enlightengan} &59.241 & 3.415 & 61.604 &3.382  & 65.083 & 3.022 &  49.985 & 2.889 & 65.321 & 1.259
\\   
RUAS \cite{liu2021ruas} &40.023 & 2.784 & 57.066 &3.135  & 45.507 & 2.398 &  27.838 & 1.191 & 47.312 & 1.097
        \\
SCI     \cite{ma2022toward}   &40.545 & 3.203 & 62.583 &3.527  & 47.148 & 2.618 &  50.843 & 1.738 & 33.146 & 1.173  \\
PairLIE  \cite{fu2023pairlie}  &64.629 & 2.761 & 62.165 &2.837  & 58.259 & 2.293 &  58.857 & 2.114 & 51.602 & 1.199
    \\
NeRCo  \cite{yang2023nerco} &61.594 & 3.245 & 52.708 &2.849  & 64.987& 2.777 & 46.455&1.475&   46.513 &  1.086 
      \\ 
% GDP \cite{fei2023gdp} \\
CLIP-LIT \cite{liang2023cliplit} &58.043 &2.932  &59.459 &3.231  & 61.456 &2.586  &52.344  &2.945  &64.344 &1.318 
\\

\hline
RQ-LLIE \cite{liu2023rqllie} & \textcolor{blue}{67.558}	& 3.537&	\textcolor{blue}{68.212} & 3.630& 68.121 & 3.001 & 	\textcolor{blue}{62.392} & 2.999   & 	46.896 & 1.437 \\

MBLLEN   \cite{lv2018mbllen}              & 57.544 & 3.695 & 67.530 &\textcolor{red}{3.953} & \textcolor{blue}{71.529} & 3.129 &  28.917 & 2.641 & 58.569 & 1.513   \\
Retinex-Net  \cite{chen2018retinex}       & 28.473 & 3.110 & 41.124 &3.037  & 42.929 & 2.628 &  26.940 & 2.285 & 28.833 & 1.329 \\
KinD++   \cite{zhang2019kindpp}         & 61.691 & 3.563 &  67.538 &3.872  & 69.125 & \textcolor{blue}{3.477} &  32.158& 2.850 & 54.632 & \textcolor{blue}{1.642}  \\
MIRNet    \cite{zamir2020mirnet}          & 57.347 & 2.932 & 57.812 &3.231  & 56.752 & 2.586 &  36.286& 2.410 &50.339 & 1.154  \\
SNR-Net  \cite{xu2022snrnet}              & 62.444 & 2.955 & 46.011 &2.804  & 57.147 & 2.329 &  58.947 & 2.354 & 55.092 & 1.003  \\  
IAT     \cite{cui2022iat}                 & 41.218 & 2.233 & 39.007 &2.130  & 39.982 & 1.670 &  44.798& 2.017 & 49.980 & 1.008 \\
LLFlow \cite{wang2022llflow}              & 64.701 & \textcolor{blue}{3.899} & 64.523 & 3.756 & 71.238 & 3.187 &  52.568 & 3.021 & 63.892 & 1.371\\
Retinexformer \cite{cai2023retinexformer} & 61.258 & 2.909 & 52.570 & 2.911 &  57.610 & 2.482  & 57.160 & 2.821 & 59.193 & 1.261\\
RetinexMamba \cite{bai2024retinexmamba}   & 63.692 & 3.266 & 55.391  & 3.266 & 57.782  & 2.809  & 58.282  & 2.688 & 36.790 & 1.145\\
LLFormer \cite{wang2023llformer} & 	50.494 & 3.735 & 	47.635 & 3.836& 51.113 & 2.856 & 	47.777 & 2.412  & 	56.222 & 1.526 \\
HVI-CIDNet \cite{yan2025hvi} & 65.390 & 3.856 &64.399 & 3.523 &68.543& 3.223&57.403&2.981 &\textcolor{blue}{66.457} & 1.305 \\
\hline
\rowcolor{gray!20} U-Net 
&\textcolor{red}{69.364} & \textcolor{red}{3.940} & \textcolor{red}{70.111} & \textcolor{blue}{3.933} & \textcolor{red}{72.947}& \textcolor{red}{3.561}& \textcolor{red}{65.421} & \textcolor{red}{3.648} & \textcolor{red}{66.513}& \textcolor{red}{2.004} \\
\bottomrule
\end{tabular}
\begin{tablenotes}
\item[1] Since the images in the VV dataset are compressed in JPEG format and contain both correctly exposed regions and severely under- or overexposed areas, we incorporate JPEG noise into the low-light synthetic pipeline \cite{zhang2023practical} and apply exposure control loss \cite{guo2020zero} to adjust local exposure.
\end{tablenotes}
\end{threeparttable}
\end{table*}

\noindent\textbf{Datasets for Performance Evaluation.} The perceptual performance of LLIE methods is evaluated using two widely recognized paired datasets, LOL-v1 \cite{chen2018retinex} and LOL-v2 \cite{yang2021sparse}, for enhancement fidelity comparison. Additionally, five unpaired benchmarks—NPE \cite{wang2013naturalness}, MEF \cite{ma2015perceptual}, DICM \cite{lee2013contrast}, LIME \cite{guo2017lime}, and VV—are employed to assess generalization. For high-level vision tasks, we conduct experiments on the ExDark \cite{loh2019getting}, the low-light subset of the ACDC dataset \cite{sakaridis2021acdc}, and the DARK FACE dataset \cite{darkface}. A detailed description of each dataset is provided in Table \ref{tab:comparision_synthetic_data}.

\subsection{Implementation Details} 
We implement our experiments by PyTorch on one NVIDIA RTX 4090 GPUs. To enhance image quality both qualitatively and quantitatively, we employ a weighted loss combination that includes L1 loss, SSIM loss \cite{wang2004ssim}, VGG perceptual loss, and UNetGAN loss \cite{wang2022realesrgan}. During training, the batch size is set to $16$, with random clipping patches sized at $128\times128\times3$ pixels. We adopt the Adam optimizer with parameters: $\alpha = 0.0001$, $\beta_1 = 0.9$, $\beta_2 = 0.999$, and $\epsilon = 10^{-8}$. We also implement a learning rate decay strategy, reducing the learning rate by a factor of 0.5 every 200k iterations. For paired datasets, a two-stage training strategy is adopted. The initial training phase serves as pre-training. Due to potential image style biases from the pre-trained dataset, a second fine-tuning stage is conducted to adapt the models to the target test sets.

\subsection{Evaluation Metrics}

We utilize four widely-used full-reference IQA metrics—PSNR, SSIM \cite{wang2004ssim}, LPIPS \cite{zhang2018lpips}, and DISTS \cite{ding2020dists}—to quantify the similarity between enhanced images and their corresponding GT images, and four SOTA BIQA metrics, including MUSIQ \cite{ke2021musiq}, LIQE \cite{zhang2023liqe}, CLIPIQA \cite{wang2022clipiqa}, and Q-Align \cite{wu2023qalign}, for assessing image quality without reference to GT images.
We also assess the impact of low-light enhancement methods on high-level vision tasks using two evaluation metrics: mean Average Precision (mAP) for object detection and Mean Intersection over Union (mIoU) for segmentation.

\subsection{Results}

\noindent\textbf{Comparison Methods.} The comparison methods are classified into two main categories: (1) Unsupervised LLIE methods or methods trained on synthetic data, including Zero-DCE \cite{guo2020zero}, EnlightenGAN \cite{jiang2019enlightengan}, RUAS \cite{liu2021ruas}, SCI \cite{ma2022toward}, PairLIE \cite{fu2023pairlie}, NeRCo \cite{yang2023nerco}, CLIP-LIT \cite{liang2023cliplit}, and AGLLNet \cite{lv2021attention}; (2) Supervised LLIE methods, including RQ-LLIE \cite{liu2023rqllie}, MBLLEN \cite{lv2018mbllen}, RetinexNet \cite{chen2018retinex}, KinD++ \cite{zhang2019kindpp}, MIRNet \cite{zamir2020mirnet}, SNR-Net \cite{xu2022snrnet}, IAT \cite{cui2022iat}, LLFlow \cite{wang2022llflow}, Retinexformer \cite{cai2023retinexformer}, RetinexMamba \cite{bai2024retinexmamba}, LLFormer \cite{wang2023llformer}, and HVI-CIDNet \cite{yan2025hvi}. For the unsupervised methods, we directly use the models provided by the authors. For the supervised methods, we either utilize the released models trained on the LOL-v1 and LOL-v2 datasets or retrain them on these datasets using the default configurations provided by the authors. To ensure a fair comparison, the global brightness fine-tuning step\footnote{https://github.com/wyf0912/LLFlow/blob/main/code/test.py} in KinD++ and LLFlow is omitted.

\noindent\textbf{Quantitative Results on LOL-v1 and LOL-v2.}  We first evaluate the models' performance on paired datasets, namely LOL-v1 and LOL-v2. Table \ref{tab:quantitative_comparision} presents quantitative results across various metrics for models trained exclusively on synthetic data and those fine-tuned on LOL-v1 or LOL-v2 after initial training on synthetic data. Several key insights emerge from this analysis.
Firstly, supervised LLIE methods trained on paired datasets generally outperform unsupervised methods, particularly in full-reference IQA metrics. Furthermore, compared to vanilla U-Net trained only on synthetic data, additional fine-tuning on paired datasets enhances model performance significantly across most metrics. These findings underscore the critical importance of constructing paired datasets for training, as they provide a reliable foundation for improving model performance.
Moreover, even without a complex model design, the vanilla U-Net outperforms SOTA methods across most quality metrics. These results highlight the effectiveness of our synthetic data generation approach in simulating real-world low-light conditions and demonstrate the viability of our innovative network architecture.

\noindent\textbf{User Study.} We conduct a user study to assess the subjective quality of enhanced low-light images. Twenty participants (10 males, 10 females) take part in a questionnaire survey, each rating images on a five-point Absolute Category Rating (ACR) scale as defined by the ITU-R \cite{bt2002methodology}: 5 (Excellent), 4 (Good), 3 (Fair), 2 (Poor), and 1 (Very Poor). The study spans 30 minutes of evaluation, followed by a 10-minute break to reduce visual fatigue and help participants maintain focus.
We collect 43,913 annotations for 2,530 images, including 115 test images (15 from LOL-v1 and 100 from LOL-v2) enhanced by 21 models, along with their corresponding ground-truth (GT) images. Each image is rated by at least 15 subjects. Following recommendations in \cite{bt2002methodology}, we identify and remove outlier ratings that exceed 1.96 standard deviations from the mean (95\% confidence interval). Subjects with an outlier rate above 5\% are deemed invalid. After data purification, 2.35\% of the ratings are classified as outliers and subsequently removed.

The results of user comparisons are presented in the last column of Table \ref{tab:quantitative_comparision}, and several key observations that emerge from the data. First, there is a notable lack of correlation between the user study scores and CLIPIQA scores, suggesting that CLIPIQA performs sub-optimally in assessing low-light enhanced images (see TABLE \ref{tab:correlation}). 
In contrast, the vanilla U-Net consistently achieves the highest average scores across all sections, indicating that the enhanced images it produces are visually more appealing than those generated by other methods. Interestingly, the user study scores for U-Net enhanced images even exceed those of the GT images, suggesting that the enhanced images may offer superior visual quality or appeal compared to the original GT images. This highlights the effectiveness of the pre-training using our synthetic pipeline.

We also evaluate the correlations, \ie, the Spearman Rank Correlation Coefficient (SRCC), Pearson Linear Correlation Coefficient (PLCC), and Kendall Rank Correlation Coefficient (KRCC), between various IQA methods and the MOS from our user study to identify suitable metrics for quality assessment in LLIE. 
The results show that among full-reference metrics, LPIPS \cite{zhang2018lpips} and DISTS \cite{ding2020dists} best align with human perception, while PSNR and SSIM \cite{wang2004ssim} show only moderate correlation. Traditional BIQA methods like NIQE \cite{mittal2012making} and BRISQUE \cite{mittal2012making} perform poorly, whereas DNN-based BIQA methods such as Q-Align \cite{wu2023qalign} and LIQE \cite{zhang2023liqe} exhibit stronger correlations. These findings underscore the importance of selecting appropriate IQA metrics for LLIE tasks, recommending LPIPS and DISTS when ground truth images are available, and advanced BIQA methods like Q-Align and LIQE when they are not.

\noindent\textbf{Quantitative Results on NPE, MEF, DICM, LIME and VV.} To further validate the effectiveness of our proposed low-light synthesis pipeline, we conducted tests on five unpaired, real-captured benchmarks: NPE \cite{wang2013naturalness}, MEF \cite{ma2015perceptual}, DICM \cite{lee2013contrast}, LIME \cite{guo2017lime} and VV. We employ Q-Align \cite{wu2023qalign} and LIQE \cite{zhang2023liqe} as evaluation metrics, as they achieve superior performance for assessing the image quality of authentic distortions (see TABLE \ref{tab:correlation}). The results are summarized in TABLE \ref{tab:generalization_comparision}. 
Our analysis indicates that no single LLIE method outperforms all others across every assessment metric and dataset. However, the vanilla U-Net, trained using our synthetic pipeline, achieves nine top and one second-place results, outperforming all other models. Overall, the vanilla U-Net trained with our synthetic pipeline demonstrates stable, superior performance and better generalization to other datasets compared to other LLIE methods.

\begin{figure*}[!h]
    \begin{minipage}[]{\textwidth}
        \centering
             \subfigure[\scriptsize Input]{\includegraphics[width=0.135\linewidth]{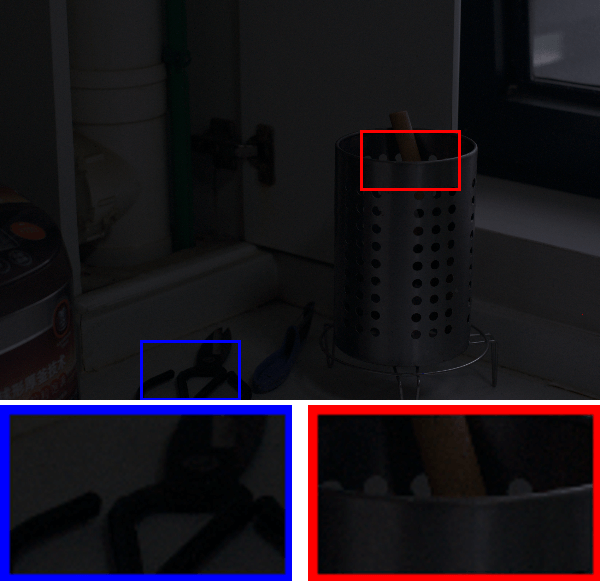}}\hskip.3em
            \subfigure[\scriptsize Zero-DCE] {\includegraphics[width=0.135\linewidth]{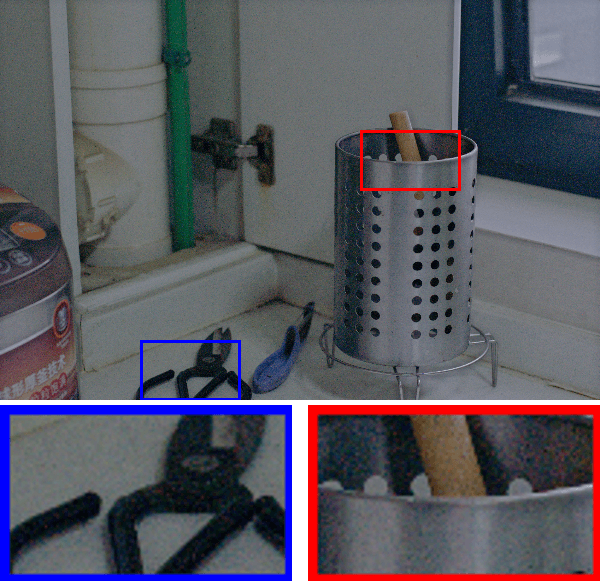}}\hskip.3em
            \subfigure[\scriptsize EnlightenGAN]{\includegraphics[width=0.135\linewidth]{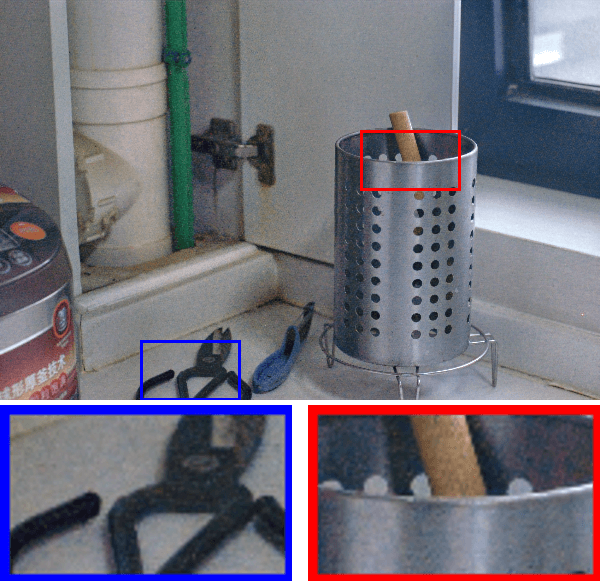}}
             \subfigure[\scriptsize RUAS]{\includegraphics[width=0.135\linewidth]{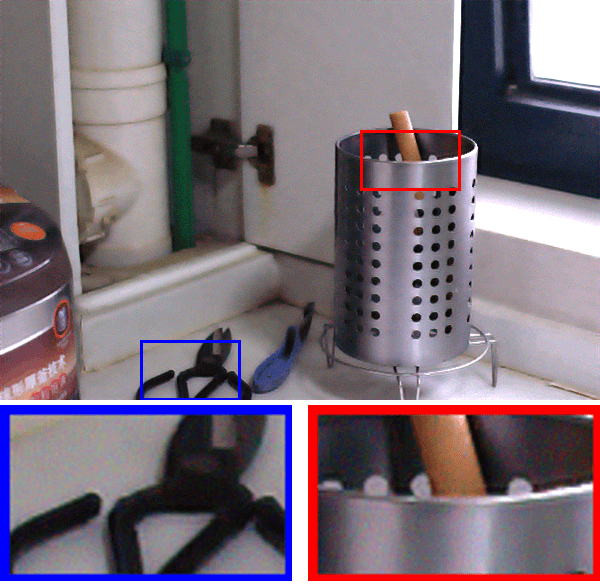}}\hskip.3em
             \subfigure[\scriptsize SCI]{\includegraphics[width=0.135\linewidth]{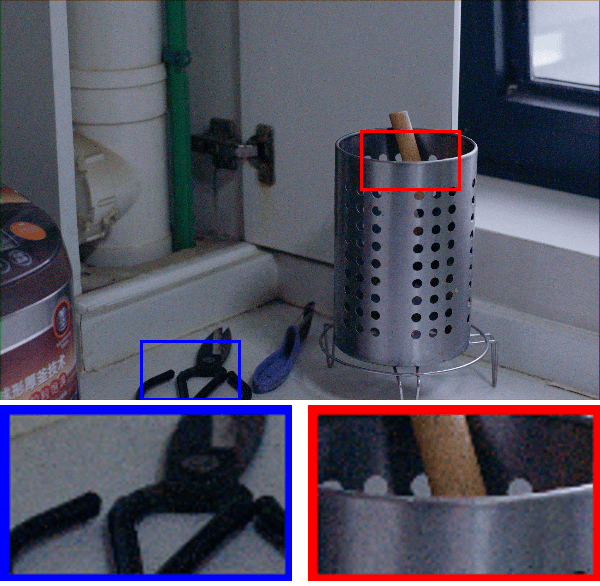}}\hskip.3em
             \subfigure[\scriptsize PairLIE]{\includegraphics[width=0.135\linewidth]{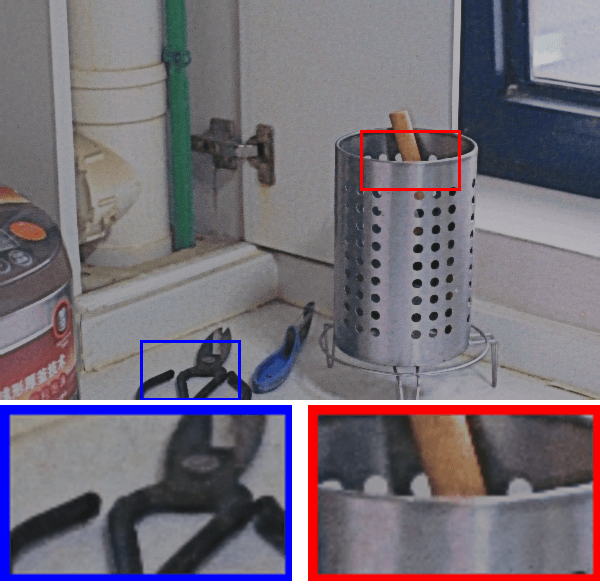}}\hskip.3em
             \subfigure[\scriptsize NeRCo]{\includegraphics[width=0.135\linewidth]{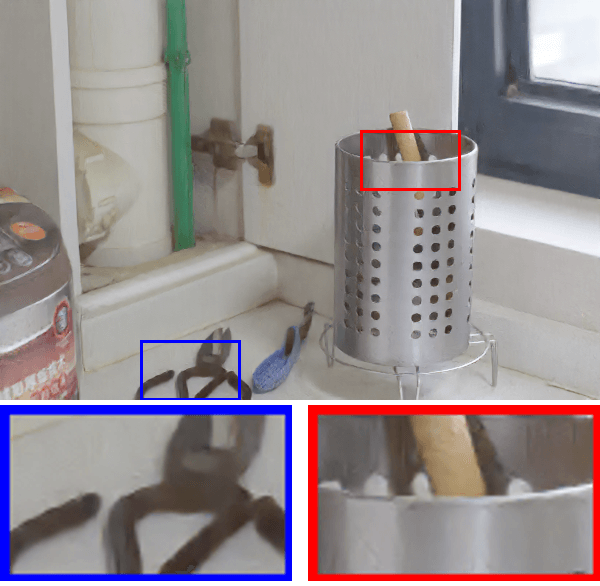}}
    \end{minipage}

    \begin{minipage}[]{\textwidth}
        \centering
         \subfigure[\scriptsize CLIP-LIT]{\includegraphics[width=0.135\linewidth]{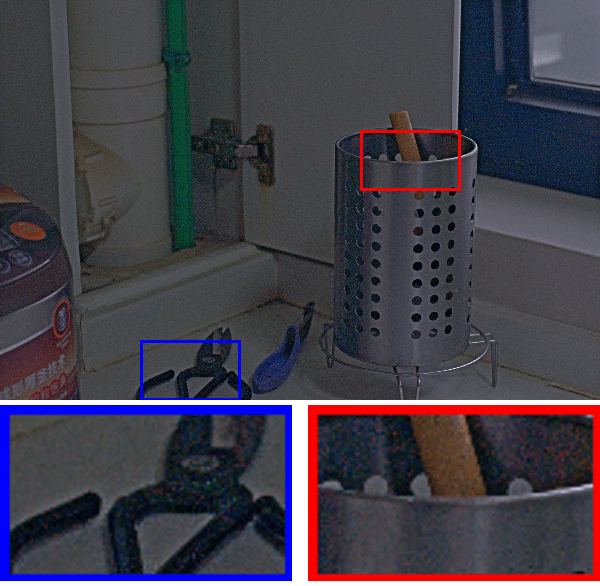}}\hskip.3em
        \subfigure[\scriptsize AGLLNet]{\includegraphics[width=0.135\linewidth]{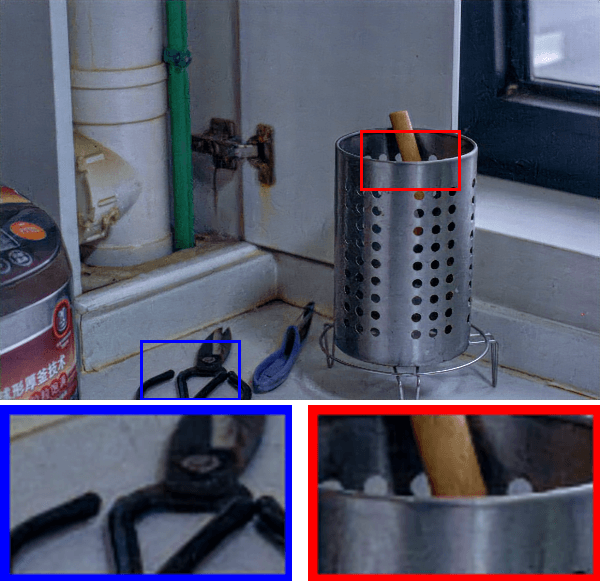}}\hskip.3em
        \subfigure[\scriptsize RQ-LLIE]{\includegraphics[width=0.135\linewidth]{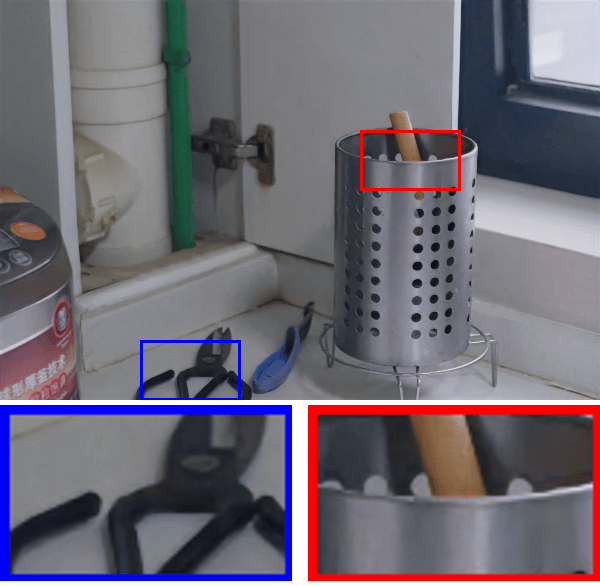}}
         \subfigure[\scriptsize MBLLEN]{\includegraphics[width=0.135\linewidth]{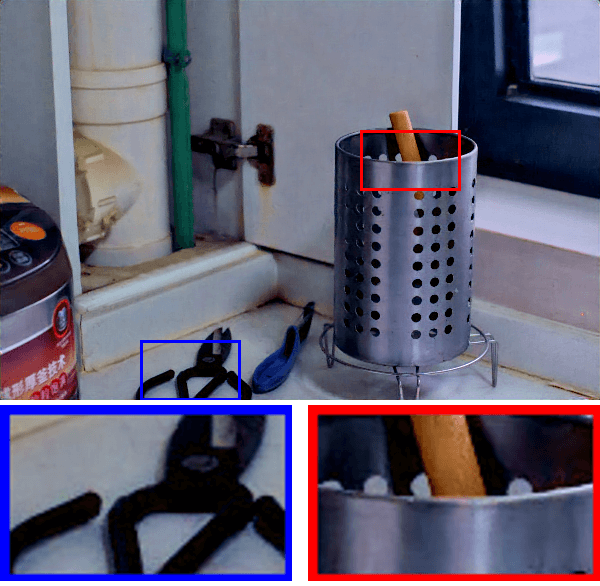}}\hskip.3em
         \subfigure[\scriptsize Retinex-Net]{\includegraphics[width=0.135\linewidth]{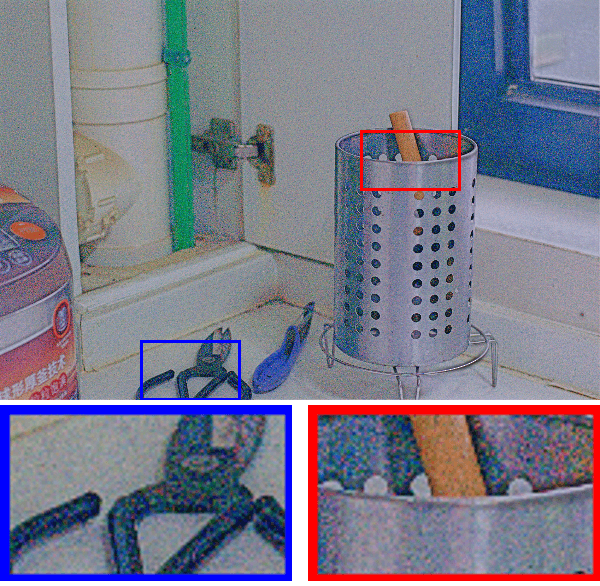}}\hskip.3em
         \subfigure[\scriptsize KinD++]{\includegraphics[width=0.135\linewidth]{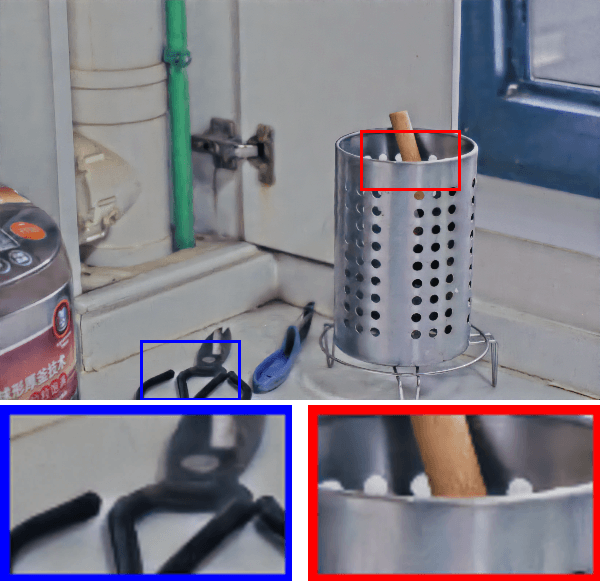}}\hskip.3em
         \subfigure[\scriptsize SNR-Net]{\includegraphics[width=0.135\linewidth]{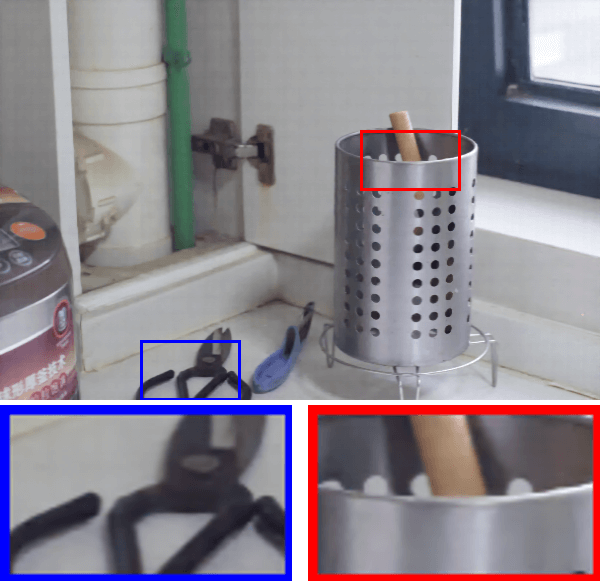}}
    \end{minipage}

    \begin{minipage}[]{\textwidth}
        \centering
         \subfigure[\scriptsize IAT]{\includegraphics[width=0.135\linewidth]{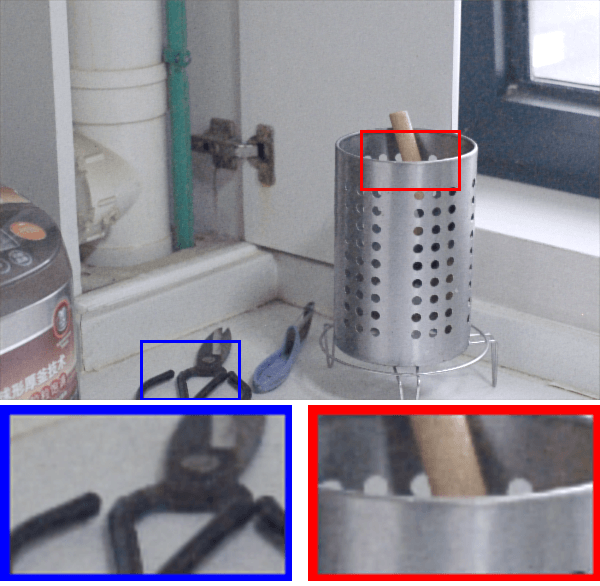}}\hskip.3em
        \subfigure[\scriptsize LLFlow]{\includegraphics[width=0.135\linewidth]{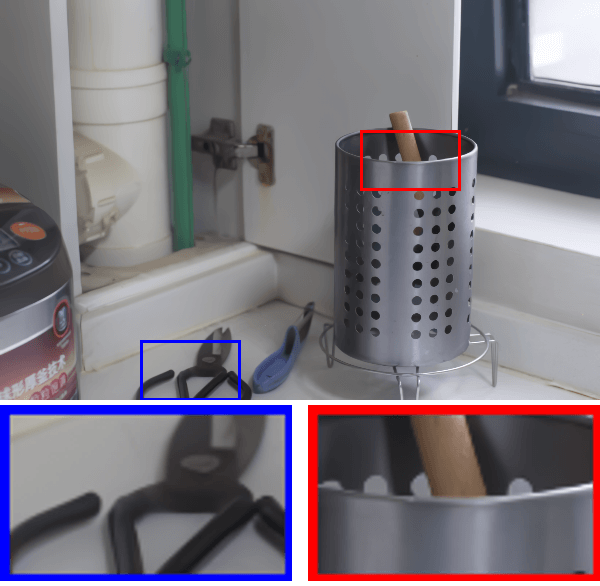}}\hskip.3em
        \subfigure[\scriptsize Retinexformer]{\includegraphics[width=0.135\linewidth]{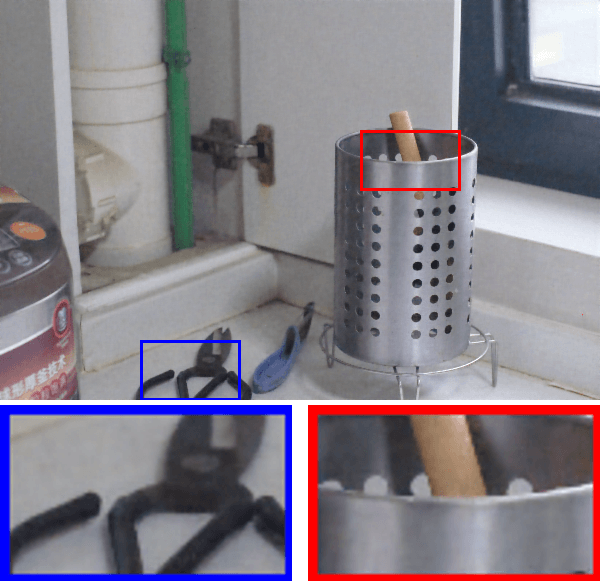}}
         \subfigure[\scriptsize RetinexMamba]{\includegraphics[width=0.135\linewidth]{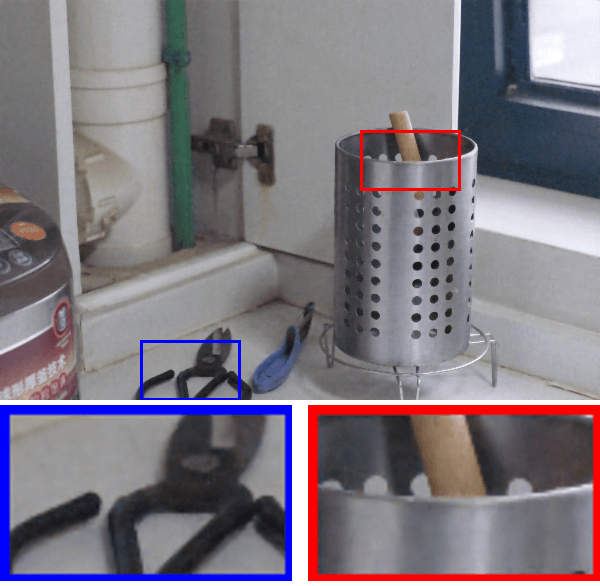}}\hskip.3em
         \subfigure[\scriptsize LLFormer]{\includegraphics[width=0.135\linewidth]{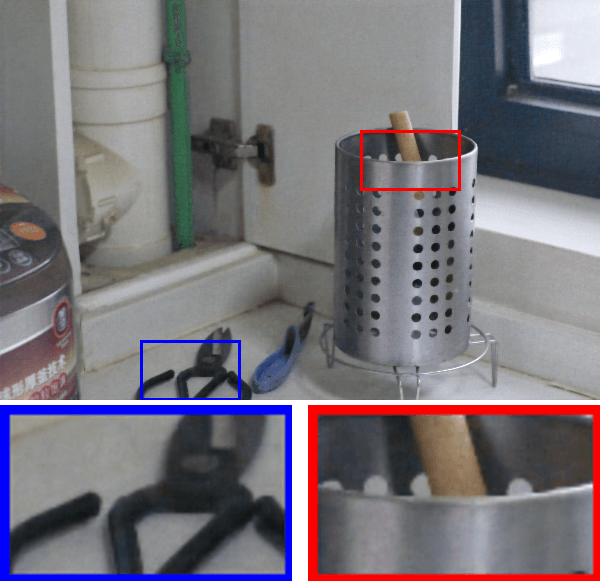}}\hskip.3em
         \subfigure[\scriptsize U-Net (FT)]{\includegraphics[width=0.135\linewidth]{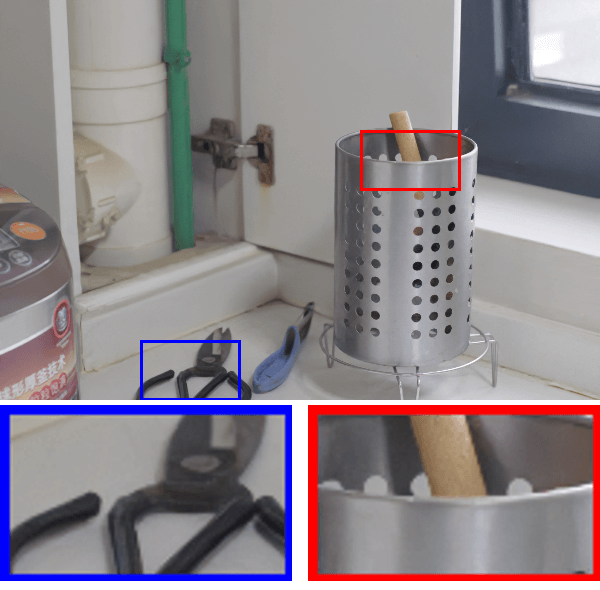}}\hskip.3em
         \subfigure[\scriptsize GT]{\includegraphics[width=0.135\linewidth]{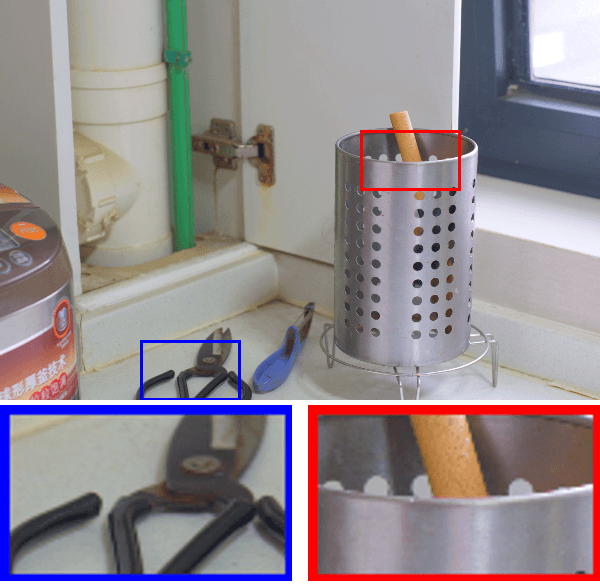}}
    \end{minipage}
\caption{Qualitative comparison of various LLIE methods on a representative sample from the LOL-v1 \cite{chen2018retinex}. FT indicates that vanilla U-Net first trained on synthetic data and then fine-tuned on LOL-v1. We observe vanilla U-Net significantly enhances visibility, reducing noise, and preserving color fidelity.  Please zoom in for details.}
\label{fig:qc_lolv1}
\end{figure*}

\begin{figure*}[!h]
    \centering
    \begin{minipage}[]{0.95\textwidth}
    \hskip -1em
        \centering
              \subfigure[\scriptsize Input]{\includegraphics[width=0.110\linewidth, height=2.2cm]{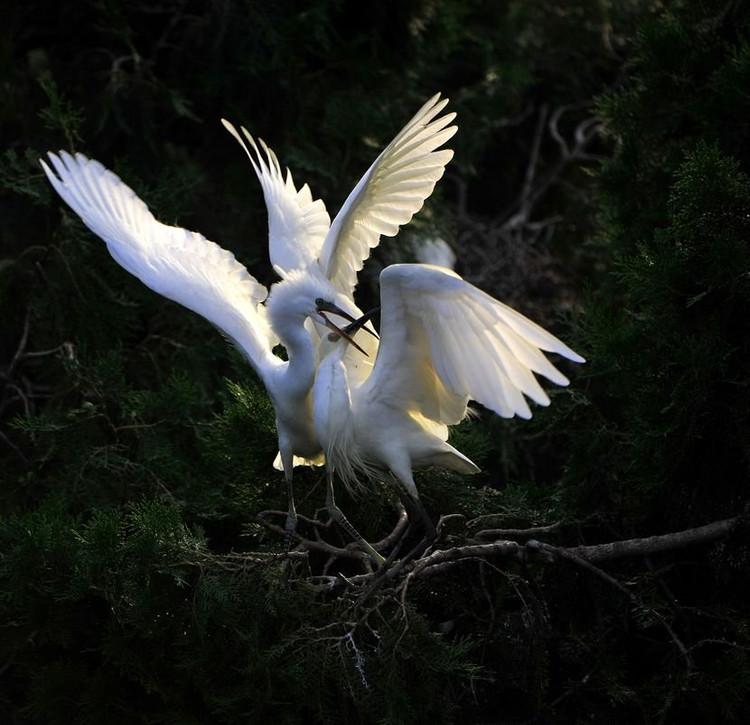}}\hskip.2em
             \subfigure[\scriptsize MIRNet]{\includegraphics[width=0.110\linewidth, height=2.2cm]{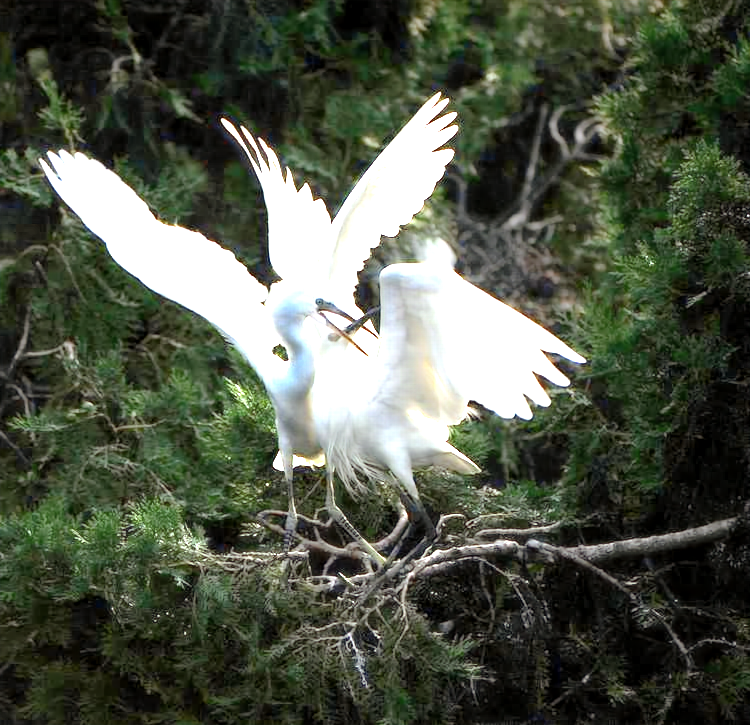}}\hskip.2em
             \subfigure[\scriptsize U-Net]{\includegraphics[width=0.110\linewidth, height=2.2cm]{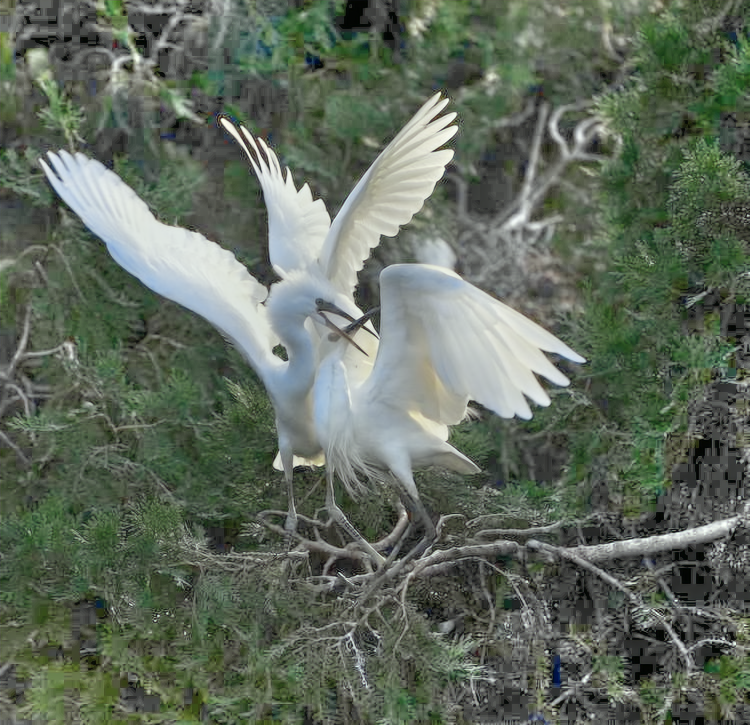}}\hskip.5em
            \subfigure[\scriptsize Input]{\includegraphics[width=0.090\linewidth, height=2.2cm]{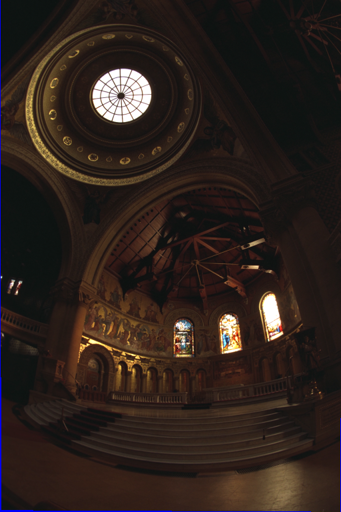}}\hskip.2em
             \subfigure[\scriptsize Zero-DCE]{\includegraphics[width=0.090\linewidth, height=2.2cm]{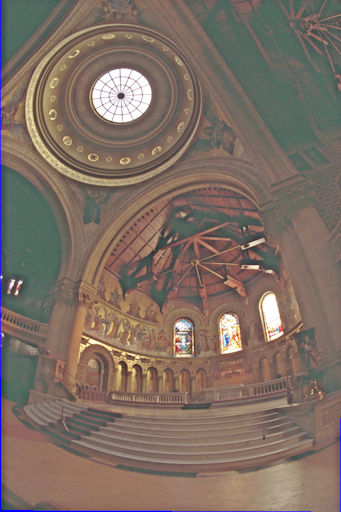}}\hskip.2em
             \subfigure[\scriptsize U-Net]{\includegraphics[width=0.090\linewidth, height=2.2cm]{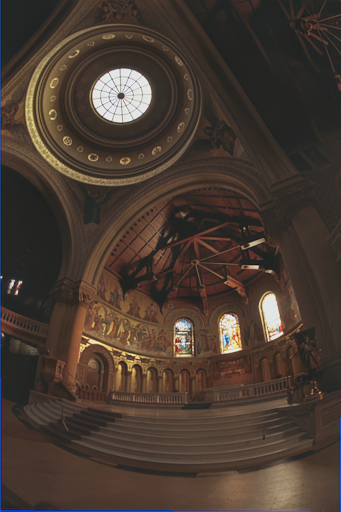}}\hskip.5em
            \subfigure[\scriptsize Input]{\includegraphics[width=0.120\linewidth, height=2.2cm]{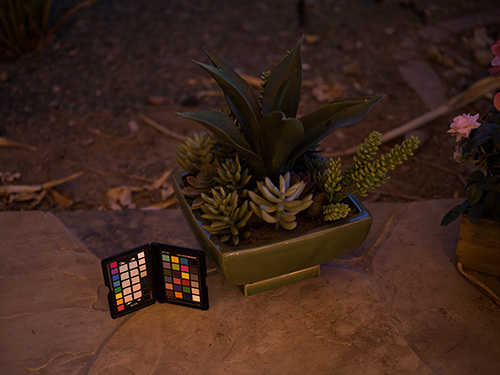}}\hskip.2em
             \subfigure[\scriptsize LLFlow]{\includegraphics[width=0.125\linewidth, height=2.2cm]{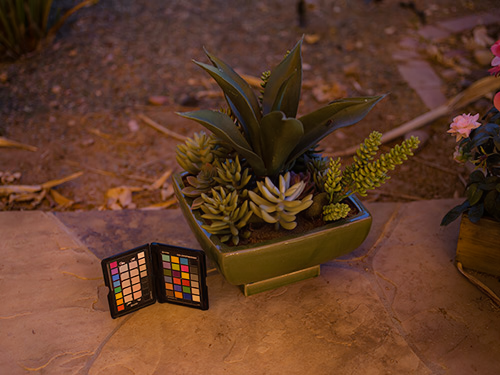}}\hskip.2em
             \subfigure[\scriptsize U-Net]{\includegraphics[width=0.120\linewidth, height=2.2cm]{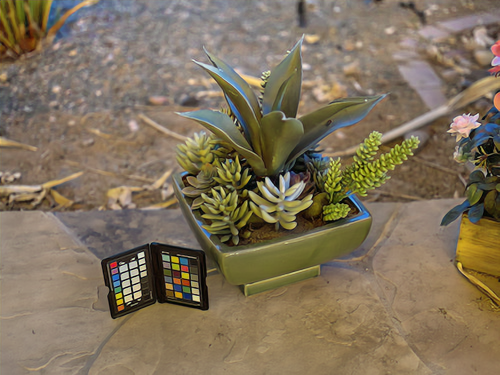}}
    \end{minipage}
    \begin{minipage}[]{0.95\textwidth}
    \hskip -1em
        \centering
        \subfigure[\scriptsize Input]{\includegraphics[width=0.172\linewidth, height=1.9cm]{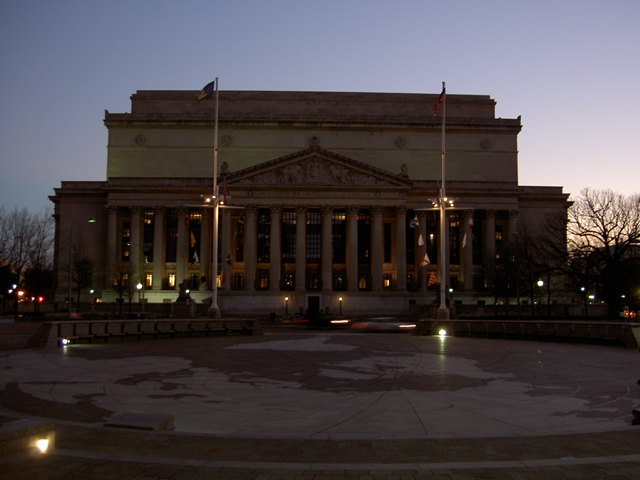}}\hskip.2em
        \subfigure[\scriptsize MBLLEN]{\includegraphics[width=0.172\linewidth, height=1.9cm]{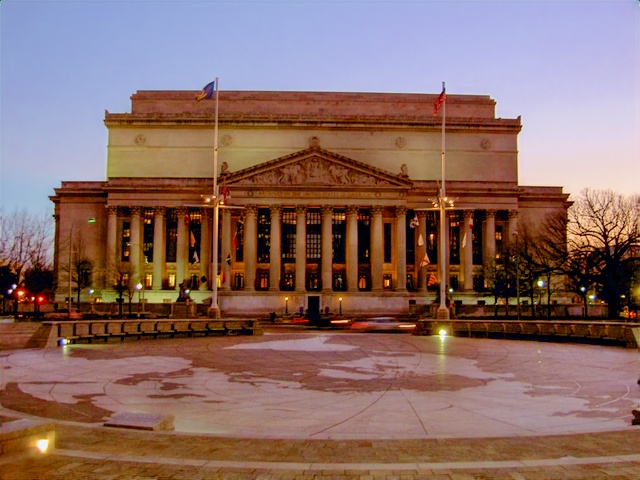}}\hskip.2em
        \subfigure[\scriptsize U-Net]{\includegraphics[width=0.172\linewidth, height=1.9cm]{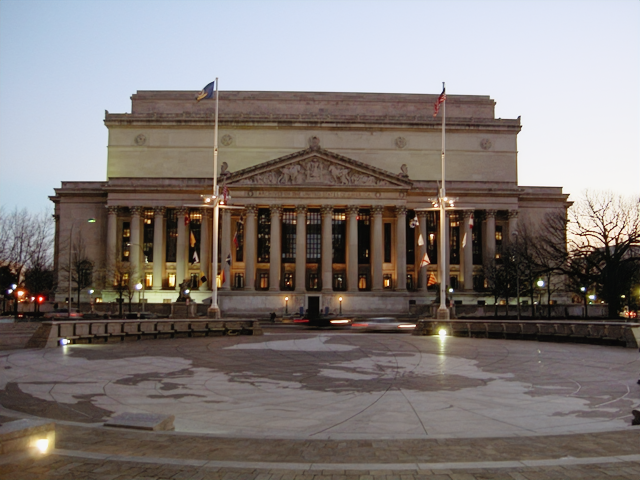}} \hskip.2em
        \subfigure[\scriptsize Input]{\includegraphics[width=0.158\linewidth, height=1.9cm]{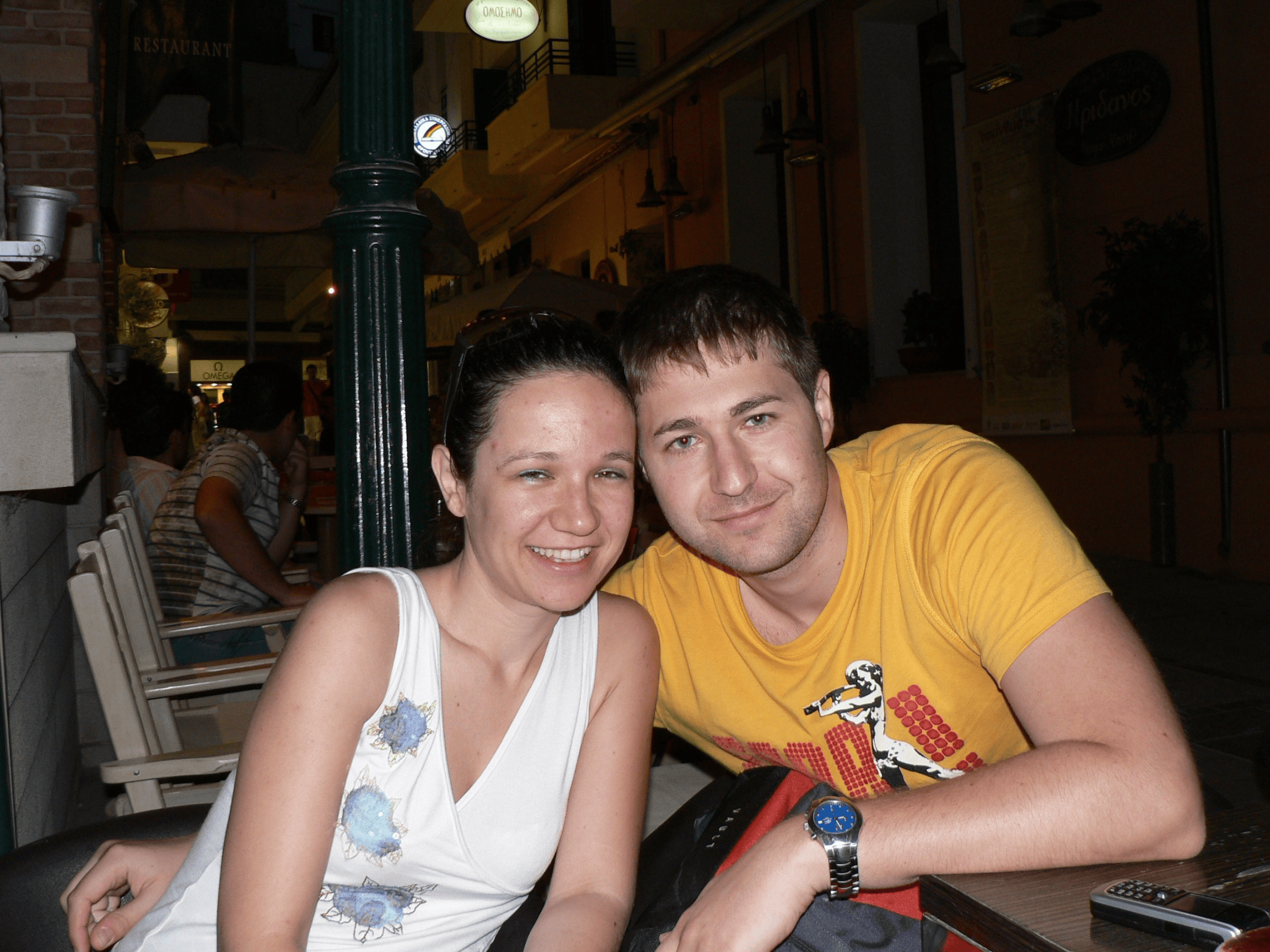}}\hskip.2em
        \subfigure[\scriptsize RetinexMamba]{\includegraphics[width=0.158\linewidth, height=1.9cm]{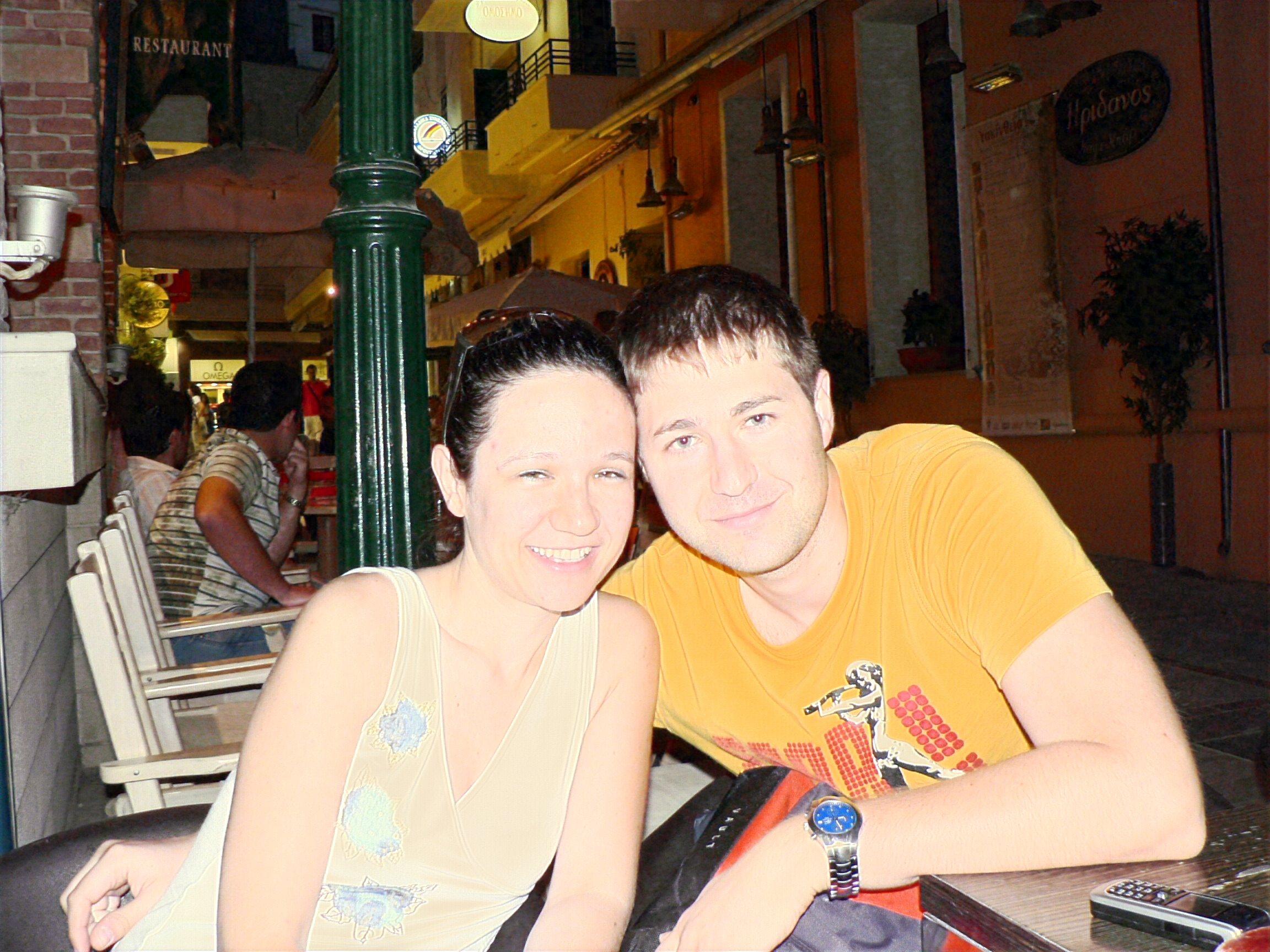}}\hskip.2em
        \subfigure[\scriptsize U-Net]{\includegraphics[width=0.158\linewidth, height=1.9cm]{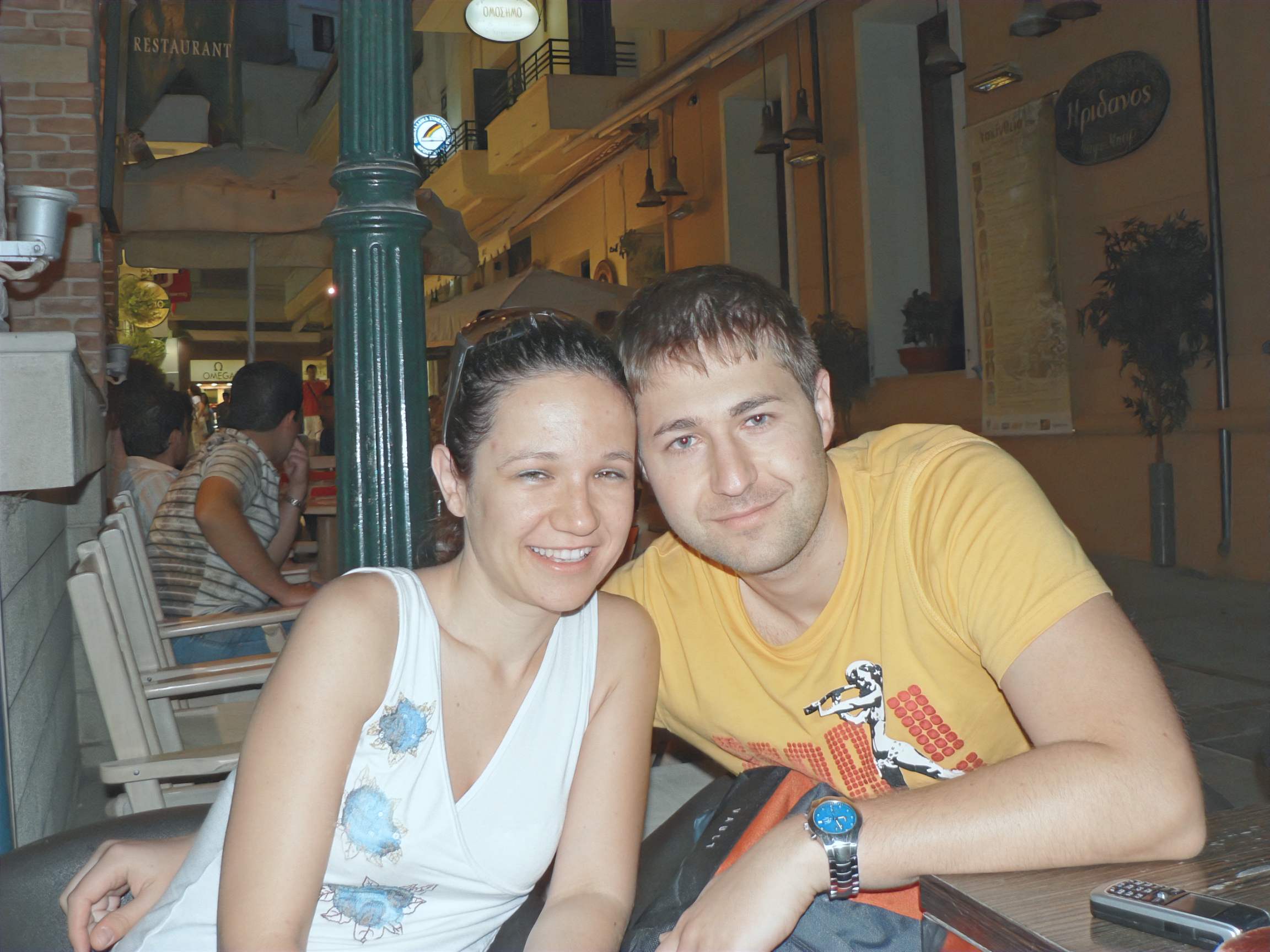}}
    \end{minipage}
\caption{Visual comparison on (a) NPE, (d) MEF, (g) LIME, (j) DICM, and (m) VV. We observe vanilla U-Net can achieve superior performance, producing more natural and vivid colors. Please zoom in for details.}
\label{fig:qc_unsupervised}
\end{figure*}

\begin{figure*}[t]
    \centering
    \begin{minipage}[]{\textwidth}
    \centering
         \subfigure[\scriptsize Input]{\includegraphics[width=0.135\linewidth]{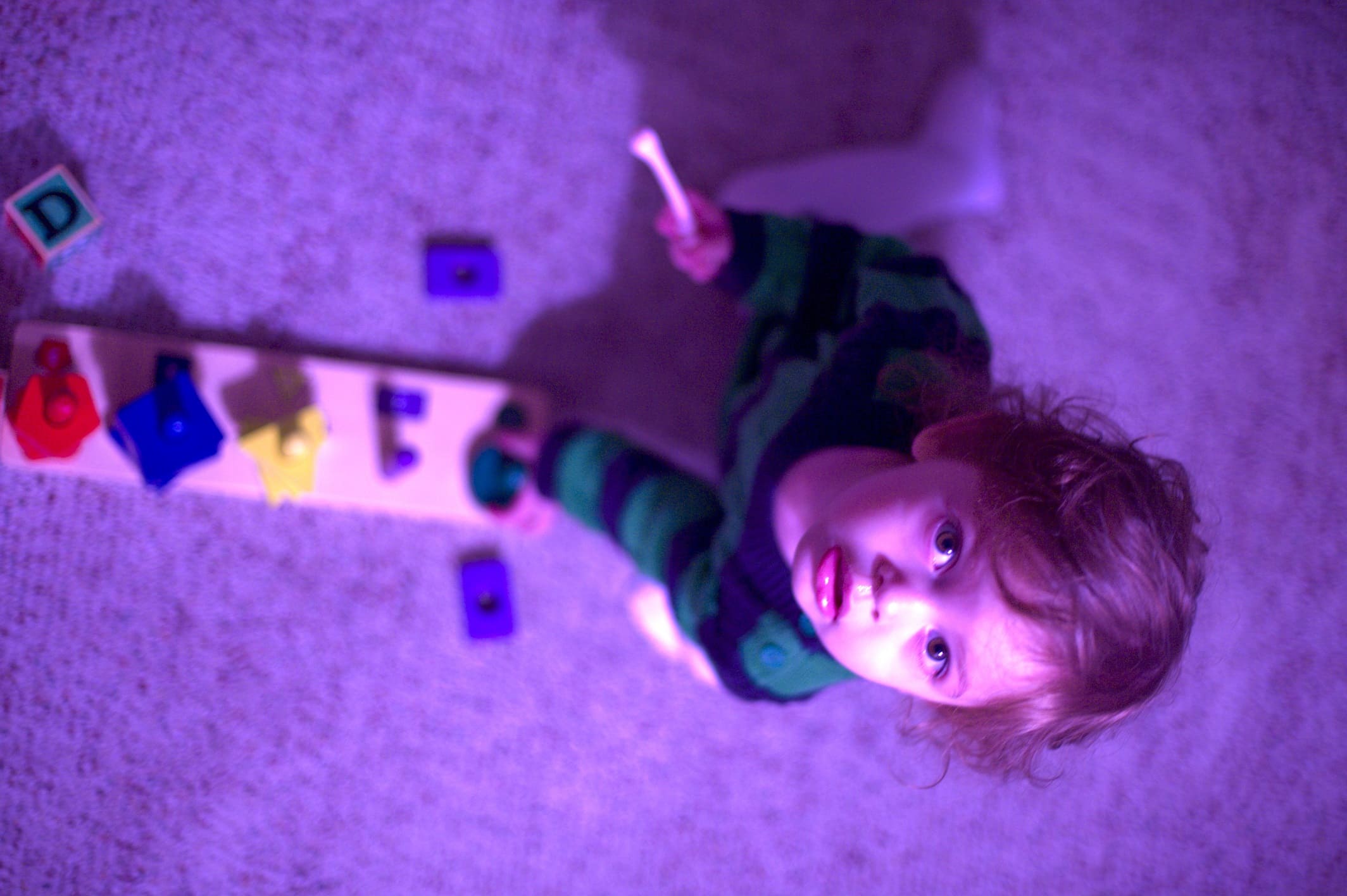}}\hskip.3em
        \subfigure[\scriptsize Zero-DCE]{\includegraphics[width=0.135\linewidth]{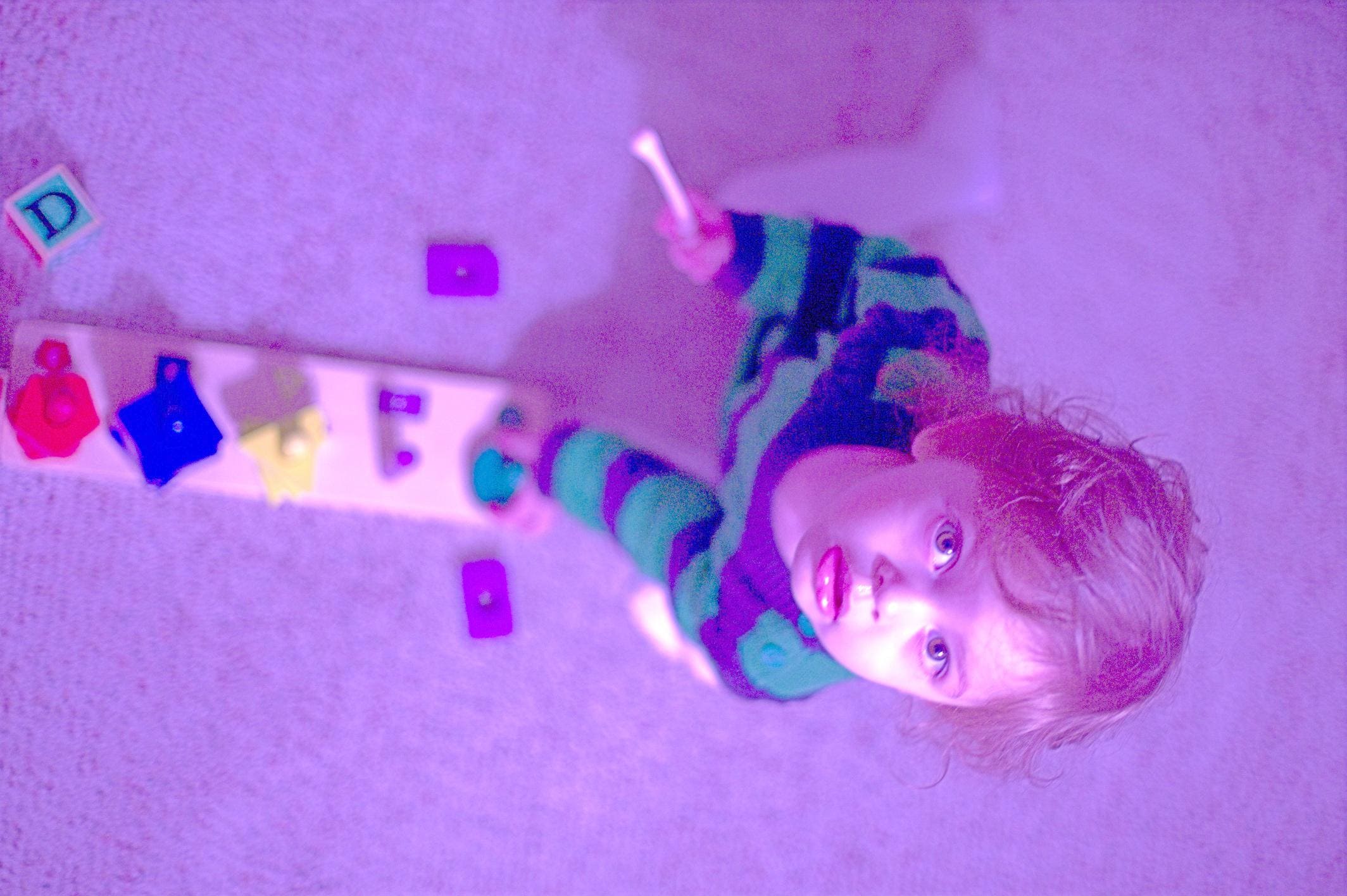}}\hskip.3em
        \subfigure[\scriptsize EnlightenGAN]{\includegraphics[width=0.135\linewidth]{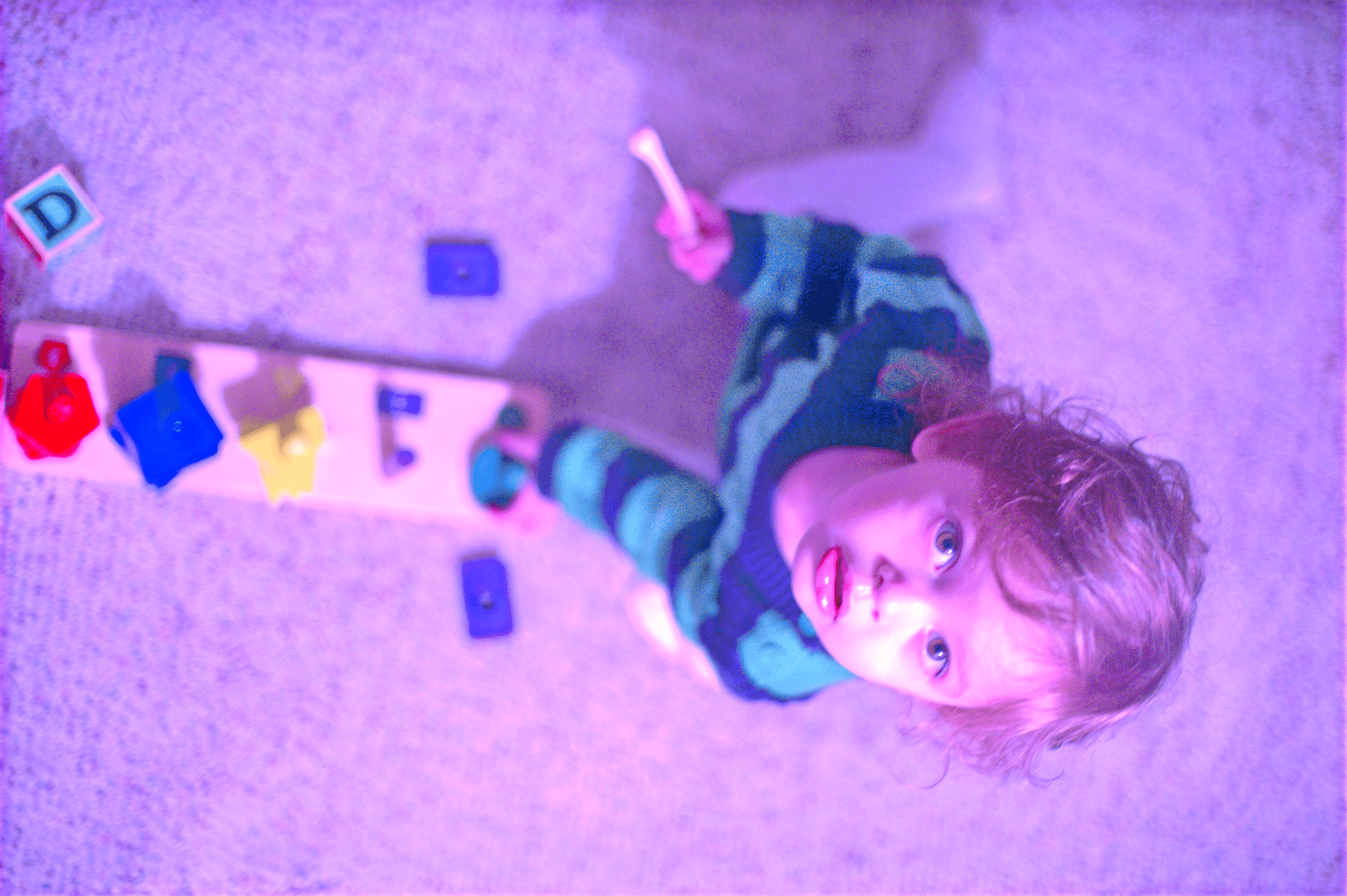}}
         \subfigure[\scriptsize RUAS]{\includegraphics[width=0.135\linewidth]{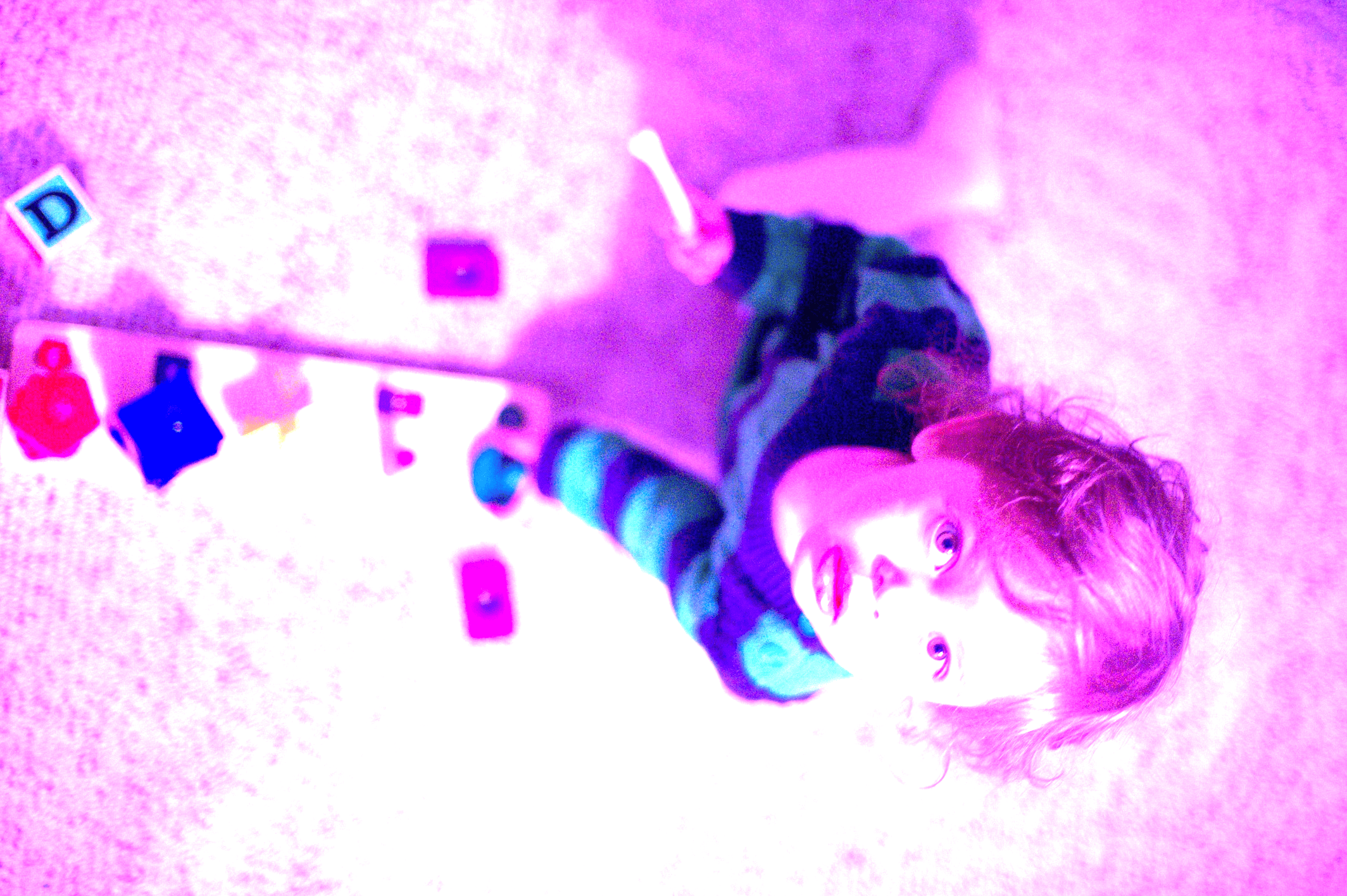}}\hskip.3em
         \subfigure[\scriptsize SCI]{\includegraphics[width=0.135\linewidth]{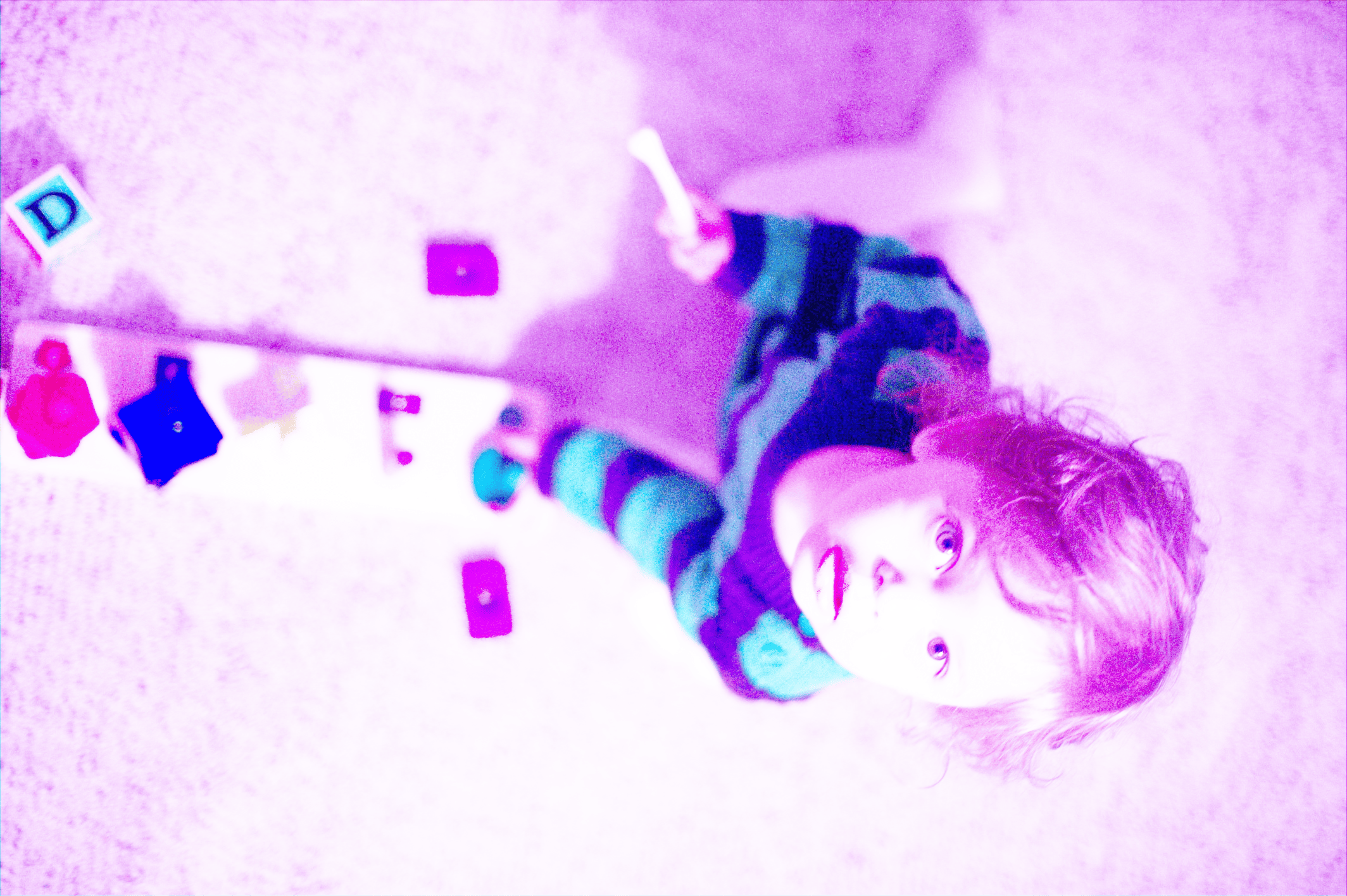}}\hskip.3em
         \subfigure[\scriptsize AGLLNet]{\includegraphics[width=0.135\linewidth]{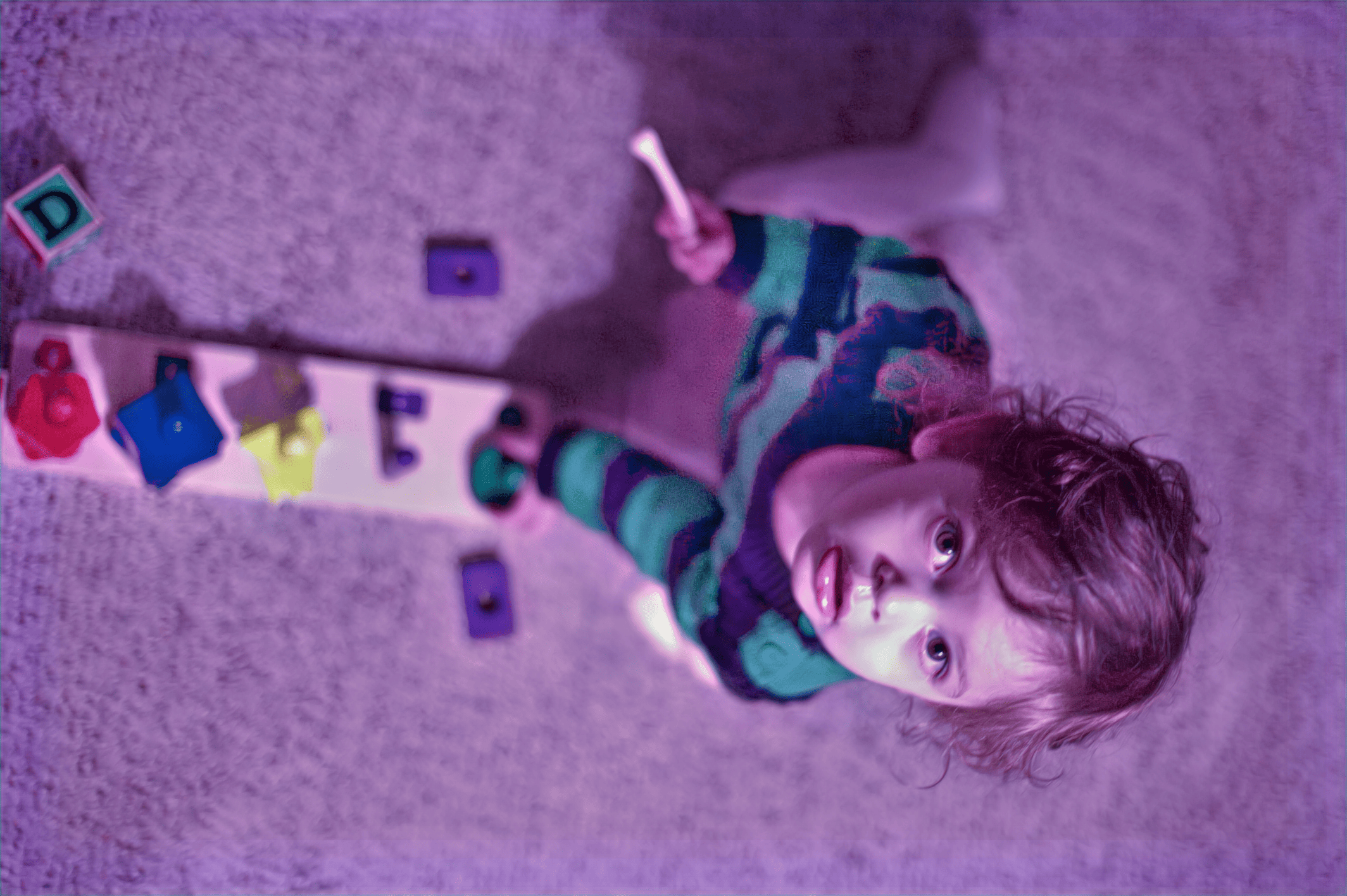}}\hskip.3em
         \subfigure[\scriptsize Retinex-Net]{\includegraphics[width=0.135\linewidth]{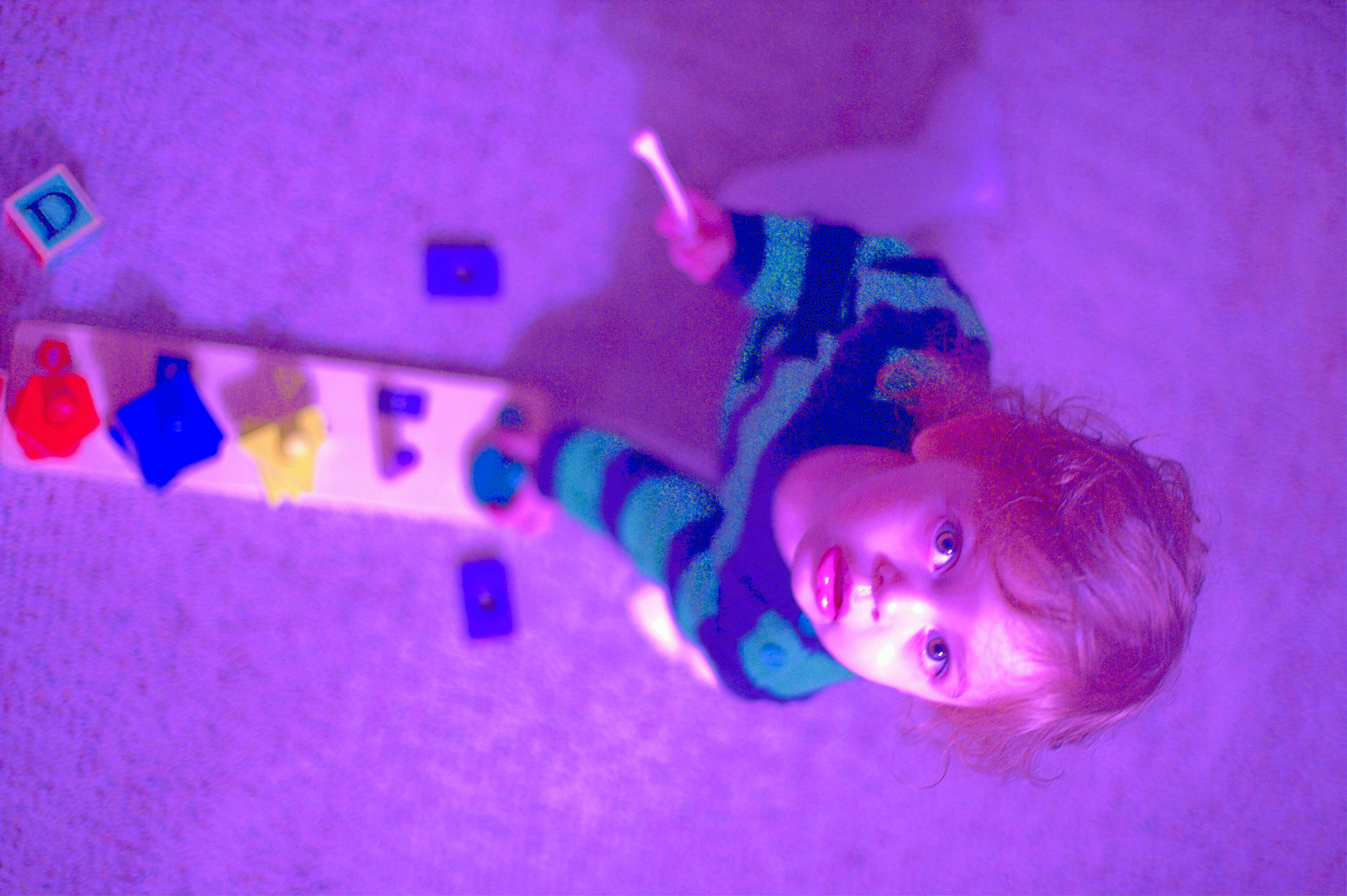}}
    \end{minipage}

    \begin{minipage}[]{\textwidth}
    \centering
        \subfigure[\scriptsize RQ-LLIE]{\includegraphics[width=0.135\linewidth]{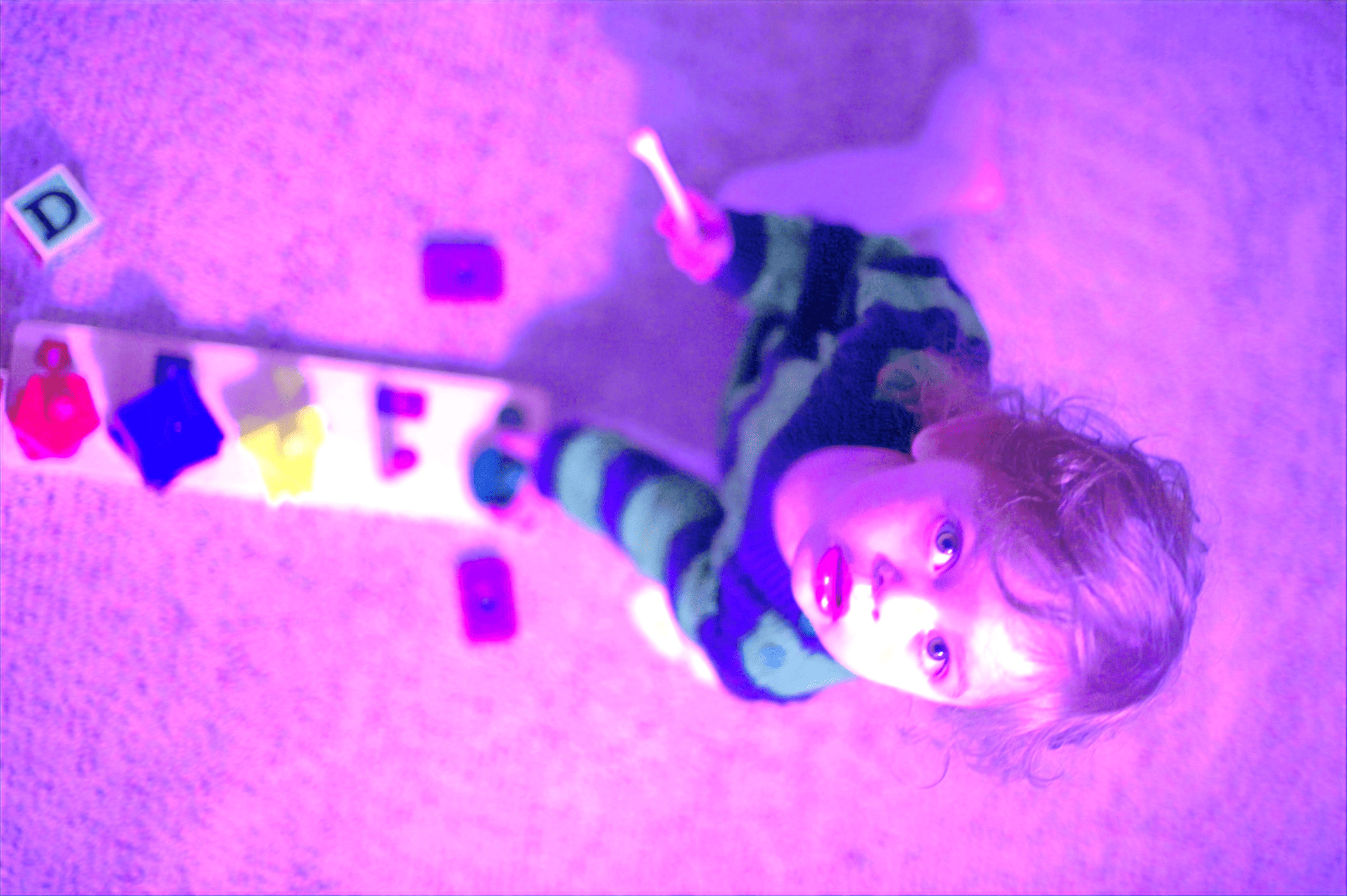}}\hskip.3em
        \subfigure[\scriptsize SNR-Net]{\includegraphics[width=0.135\linewidth]{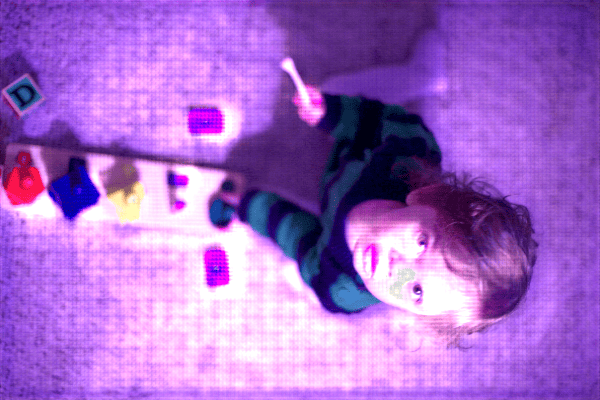}}
        \subfigure[\scriptsize LLFlow]{\includegraphics[width=0.135\linewidth]{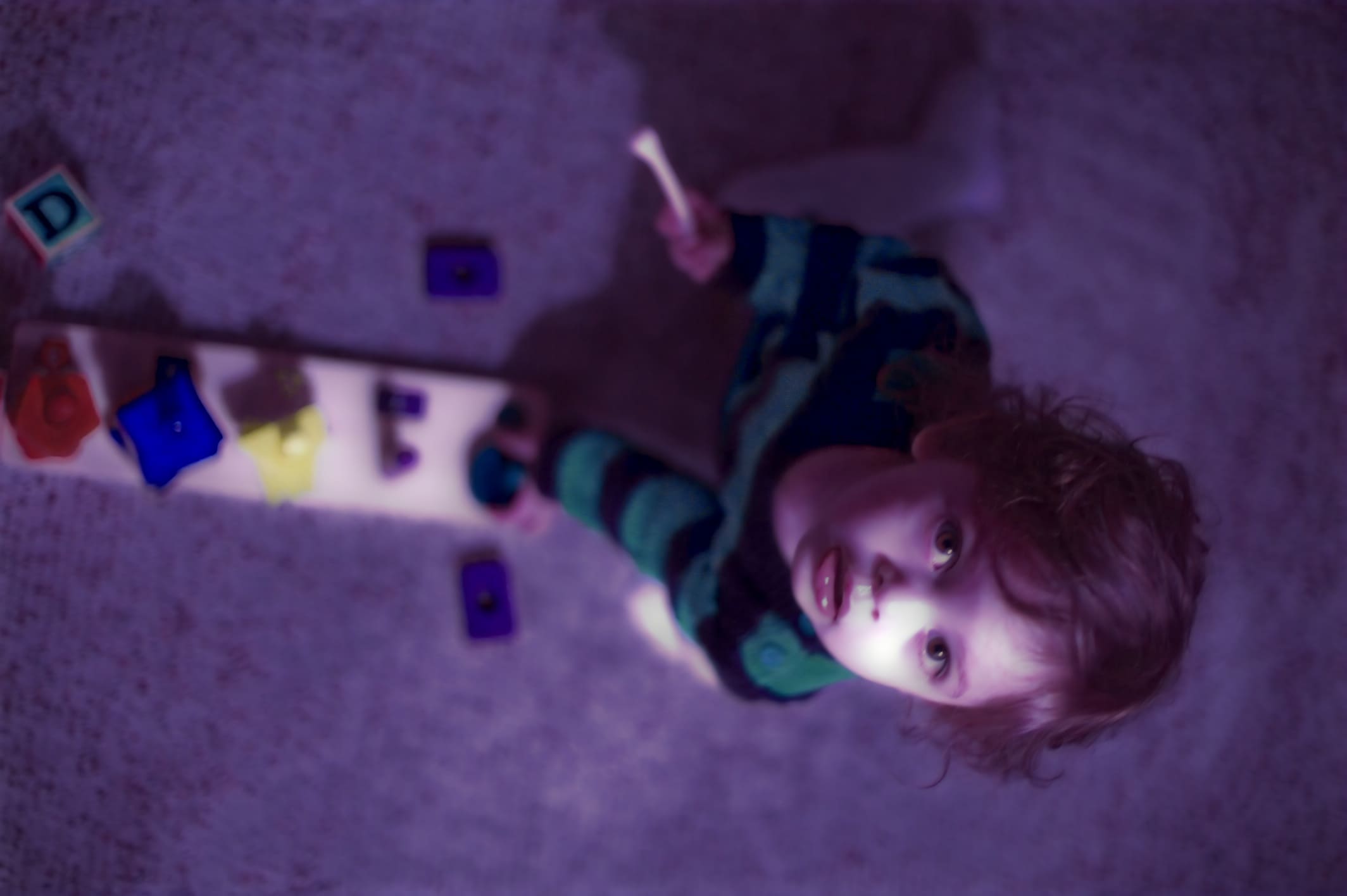}}\hskip.3em
        \subfigure[\scriptsize Retinexformer]{\includegraphics[width=0.135\linewidth]{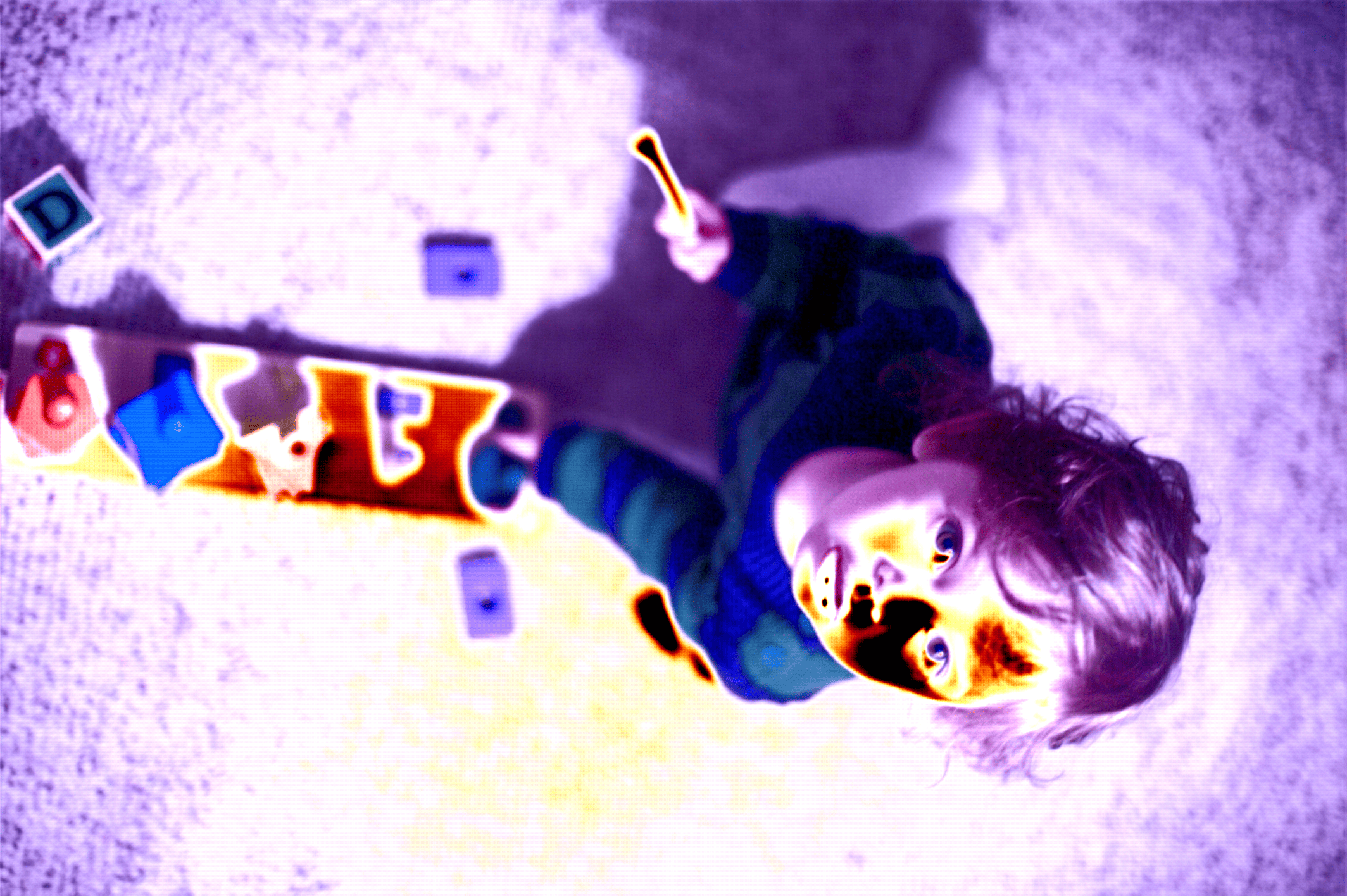}}
         \subfigure[\scriptsize RetinexMamba]{\includegraphics[width=0.135\linewidth]{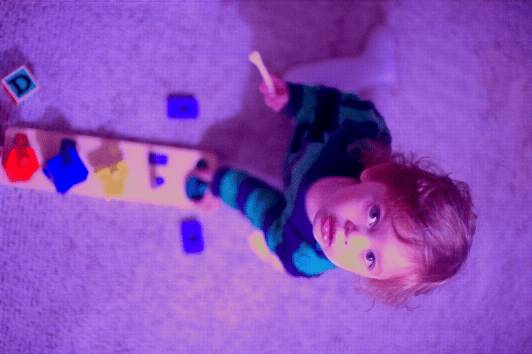}}\hskip.3em
         \subfigure[\scriptsize LLFormer]{\includegraphics[width=0.135\linewidth]{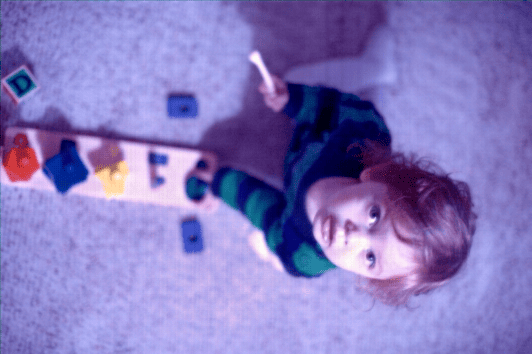}}\hskip.3em
         \subfigure[\scriptsize U-Net]{\includegraphics[width=0.135\linewidth]{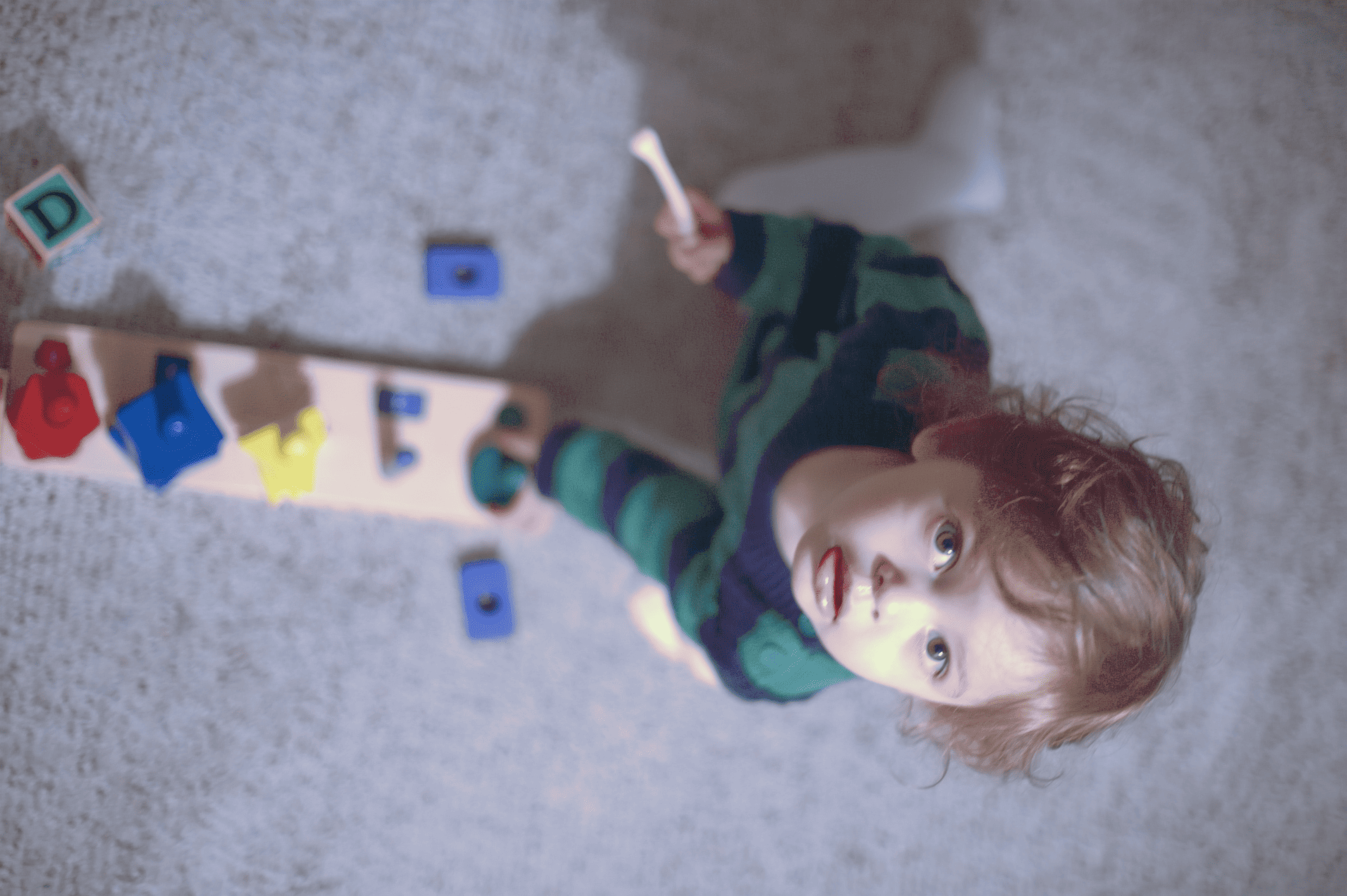}}
    \end{minipage}
\caption{Visual comparison of white balance correction. The input image is sourced from the rendered WB dataset \cite{afifi2019color}. Our proposed vanilla U-Net, trained solely on synthetic datasets, generates more visually pleasing and color-accurate images.  Please zoom in for details.}
\label{fig:qc_whitebalance}
\end{figure*}

\begin{table*}[!h]
\caption{Quantitative comparisons of LLIE models initially trained on our synthetic data and subsequently fine-tuned on paired datasets, compared to models trained exclusively on paired datasets: LOL-v1 \cite{chen2018retinex}, and LOL-v2 \cite{yang2021sparse}. $\uparrow$ ($\downarrow$) indicates that larger (smaller) values correspond to better quality. The improvement (reduction) of performance is in \textcolor{red}{red} (\textcolor{blue}{blue}) color, corresponding to $\uparrow$ ($\downarrow$).
}
\label{tab:other_method_comparision}
\centering
\begin{tabular}{l|cccc|cccc}
\toprule
LOL-v1 \cite{chen2018retinex} & PSNR$\uparrow$  & SSIM$\uparrow$  & LPIPS$\downarrow$ & DISTS$\downarrow$ & MUSIQ$\uparrow$ & LIQE$\uparrow$ & CLIPIQA+$\uparrow$  & Q-Align$\uparrow$\\ 
\midrule
%ReSCUE  & 24.084 & 0.857 & 0.094 &0.081 &72.524 & 4.322&0.630 & 77.960\\
SNR-Net  \cite{xu2022snrnet} &23.193 {\scriptsize \textcolor{red}{(1.070)}}   &0.846 {\scriptsize \textcolor{red}{(0.024)}}  &0.144 {\scriptsize \textcolor{red}{(0.012)}} &0.120 {\scriptsize \textcolor{red}{(0.018)}} &68.140 {\scriptsize \textcolor{red}{(3.161)}}  & 3.275 {\scriptsize \textcolor{red}{(0.024)}} &0.537 {\scriptsize \textcolor{red}{(0.378)}} &68.934 {\scriptsize \textcolor{red}{(~7.034~)}}   \\
Retinexformer \cite{cai2023retinexformer}  & 23.999 {\scriptsize \textcolor{red}{(0.377)}}   & 0.854 {\scriptsize \textcolor{red}{(0.026)}}  &0.100 {\scriptsize \textcolor{red}{(0.048)}}  &0.085 {\scriptsize \textcolor{red}{(0.058)}}  & 72.287 {\scriptsize \textcolor{red}{(9.377)}} & 4.261  {\scriptsize \textcolor{red}{(1.347)}} & 0.622  {\scriptsize \textcolor{red}{(0.099)}} & 77.490 {\scriptsize \textcolor{red}{(20.077)}} \\
RetinexMamba \cite{bai2024retinexmamba}   &24.133 {\scriptsize \textcolor{red}{(0.606)}}    &0.855 {\scriptsize \textcolor{red}{(0.026)}}   &0.100 {\scriptsize \textcolor{red}{(0.043)}} &0.083 {\scriptsize \textcolor{red}{(0.056)}} &72.239 {\scriptsize \textcolor{red}{(9.671)}}  &4.253 {\scriptsize \textcolor{red}{(1.215)}} &0.631 {\scriptsize \textcolor{red}{(0.104)}}   & 77.940 {\scriptsize \textcolor{red}{(19.007)}} \\

\hline \hline
LOL-v2 \cite{yang2021sparse} & PSNR$\uparrow$  & SSIM$\uparrow$  & LPIPS$\downarrow$ & DISTS$\downarrow$ & MUSIQ$\uparrow$ & LIQE$\uparrow$ & CLIPIQA+$\uparrow$  & Q-Align$\uparrow$ \\ 
\midrule
%ReSCUE & 22.611 & 0.867 & 0.105 & 0.091 & 69.814 &4.017 & 0.594 &73.767\\
SNR-Net  &21.645 {\scriptsize \textcolor{red}{(0.416)}}  &0.856 {\scriptsize \textcolor{red}{(0.006)}} &0.160  {\scriptsize \textcolor{blue}{(0.004)}}& 0.133 {\scriptsize \textcolor{blue}{(0.002)}} & 64.303 {\scriptsize \textcolor{red}{(0.430)}}& 2.897 {\scriptsize \textcolor{red}{(0.097)}} &0.496  {\scriptsize \textcolor{red}{(0.008)}}& 65.418  {\scriptsize \textcolor{red}{(~3.210~)}}\\
Retinexformer   & 22.514   {\scriptsize \textcolor{red}{(0.278)}} & 0.874  {\scriptsize \textcolor{red}{(0.025)}} & 0.106 {\scriptsize \textcolor{red}{(0.055)}}&0.089   {\scriptsize \textcolor{red}{(0.060)}} & 69.030 {\scriptsize \textcolor{red}{(7.821)}} & 3.817 {\scriptsize \textcolor{red}{(0.897)}} & 0.590 {\scriptsize \textcolor{red}{(0.100)}}& 77.940 {\scriptsize \textcolor{red}{(18.621)}} \\
RetinexMamba    & 22.345 {\scriptsize \textcolor{red}{(0.335)}}  & 0.870  {\scriptsize \textcolor{red}{(0.029)}} & 0.113  {\scriptsize \textcolor{red}{(0.046)}} & 0.094  {\scriptsize \textcolor{red}{(0.054)}} & 68.446  {\scriptsize \textcolor{red}{(7.044)}} & 3.825  {\scriptsize \textcolor{red}{(0.827)}} & 0.592  {\scriptsize \textcolor{red}{(0.103)}} & 73.572  {\scriptsize \textcolor{red}{(15.486)}}  \\
\bottomrule 
\end{tabular}
\end{table*}

\noindent\textbf{Qualitative Comparison.} Fig. \ref{fig:qc_lolv1} presents the visual comparison of different LLIE methods on the LOL-v1 dataset. While all methods improve visibility, several, such as Zero-DCE, EnlightenGAN, RUAS, SCI, PairLIE, CLIP-LIT, AGLLNet, Retinex-Net, and IAT, struggle with noise suppression. Furthermore, MBLLEN, SNR-Net, LLFlow, Retinexformer, RetinexMamba, and LLFormer display somewhat dark when compared to the GT image. In contrast, vanilla U-Net achieves a more visually pleasing result by significantly enhancing lightness, reducing noise, and preserving accurate color fidelity. Fig. \ref{fig:qc_unsupervised} illustrates the quantitative results across five unpaired, real-captured benchmarks. We observe that existing SOTA LLIE methods may result in images with overexposure (see (b) and (n)) and unnatural colors (see (e), (h) and (k)) . In contrast, vanilla U-Net  generalizes more effectively across diverse scenes, producing images with natural and vivid colors.

\noindent\textbf{White Balance Correction.}
Low-light environments often cause improper color balance, resulting in images with distorted hues and unnatural tones. White balance correction, which adjusts color temperature, is essential for LLIE methods to mitigate the effects of ambient lighting and improve color accuracy. Fig. \ref{fig:qc_whitebalance} presents a visual comparison of white balance correction capabilities across different LLIE methods. The results demonstrate that the vanilla U-Net trained with our synthetic pipeline produces visually superior and color-accurate images, while other methods exhibit color deviation and introduce noticeable artifacts.

\begin{table*}[t]
    \caption{Quantitative low-light detection results on ExDark \cite{loh2019getting}.  The table is organized as follows: unsupervised LLIE methods and methods trained on synthetic data (the top section); supervised LLIE methods (the second section) and the vanilla U-Net (the third section). $\uparrow$ indicates that a larger value corresponds to better performance. In each class (\ie, column), the best result is in \textcolor{red}{red} color whereas the second best one is in \textcolor{blue}{blue} color.}
    \label{tab:exdark}
    \centering
    \begin{tabular}{l|cccccccccccc|c}
    \toprule
    Class & Bicycle & Boat & Bottle & Bus & Car & Cat & Chair & Cup & Dog & Motor & People & Table & Mean $\uparrow$ \\
    \midrule
    Zero-DCE  \cite{guo2020zero}& \textcolor{blue}{0.859}&	0.718	&0.674	&\textcolor{blue}{0.903}&	0.795	&0.695&	0.656	&0.627&	\textcolor{red}{0.782}&	0.761	&0.764&	0.562	&0.733\\
    EnlightenGAN \cite{jiang2019enlightengan} & 0.795&	\textcolor{blue}{0.762}	& 0.685	&0.872&	\textcolor{red}{0.836}	&0.738&	0.682	& {0.697}&	0.736&	0.734	&0.753&	0.579	&0.739  \\   
    RUAS \cite{liu2021ruas} & 0.727&	0.578	&0.555	&0.734&	0.674	&0.501&	0.541	&0.471&	0.557&	0.593	&0.566&	0.484	&0.582 \\
    SCI     \cite{ma2022toward}   & 0.799&	0.711	&0.668	&0.861&	0.805	&\textcolor{red}{0.752}&	\textcolor{blue}{0.686}	&0.686&	0.733&	0.683	&0.727&	0.581	&0.724  \\
    PairLIE  \cite{fu2023pairlie}& 0.858 &	0.710	&0.630	&0.866&	0.809	&0.669&	0.641	&0.601&	0.719&	0.727	&0.736&	0.561	&0.711 \\
    NeRCo  \cite{yang2023nerco}& 0.729&	0.678	&0.598	&0.844&	0.757	&0.699&	0.601	&0.560&	0.674&	0.608	&0.678&	0.543	&0.664\\ 
    % GDP \cite{fei2023gdp} \\
    CLIP-LIT \cite{liang2023cliplit} & 0.795&	0.711	&0.614	&0.845&	0.784	&0.714&	0.675	&0.669&	0.743&	0.768	&0.730&	0.541	&0.716\\
    AGLLNet \cite{lv2021attention}    & 0.833&	 0.746&	0.684&	0.890	&0.828	&\textcolor{blue}{0.742}	&\textcolor{red}{0.692}	&\textcolor{red}{0.706}&	0.740	&0.703	&0.750	&0.570&	\textcolor{blue}{0.740}\\
    %\rowcolor{gray!20} ReSCUE \\
    \hline
    RQ-LLIE  & 0.858&	0.710	&0.655	&0.898&	0.813	&0.646&	0.675	&0.636&	0.717&	0.735	&\textcolor{blue}{0.767}&	0.598	&0.726\\
    MBLLEN   \cite{lv2018mbllen}  & 0.845&	0.733	&0.654	&0.894&	0.811	&0.632&	0.681	&0.633&	0.721&	0.720	&0.758&	0.550	&0.719   \\
    Retinex-Net  \cite{chen2018retinex} & 0.850&	0.738	&0.670	&0.890&	0.800	&0.696&	0.650	&0.619&	0.757&	\textcolor{blue}{0.770}	&0.753&	0.536	&0.727\\
    KinD++   \cite{zhang2019kindpp}   & 0.794&	0.744	&0.667	&0.898&	0.831	&0.724&	0.642	&\textcolor{blue}{0.692}&	0.726&	0.727	&0.740&	0.564	&0.729  \\
    MIRNet    \cite{zamir2020mirnet}  & 0.837&	0.647	&0.497	&0.855&	0.726	&0.730&	0.581	&0.502&	0.676&	0.614	&0.648&	0.465	&0.648  \\
    SNR-Net  \cite{xu2022snrnet} & 0.853&	0.725	&0.645	&0.877&	0.795	&0.684&	0.647	&0.642&	0.713&	0.762	&0.762&	0.524	&0.719    \\  
    IAT     \cite{cui2022iat} & 0.784&	0.756	&0.647	&0.855&	0.808	&0.731&	0.672	&0.663&	0.741&	0.693	&0.721&	0.585	&0.721     \\
    LLFlow \cite{wang2022llflow} & 0.861&	0.733	&\textcolor{blue}{0.702}	&0.884&	0.812	&0.738&	0.652	&0.654&	0.750&	0.735	&0.757&	0.573	&0.737\\
    Retinexformer \cite{cai2023retinexformer} & 0.824	&0.628	&0.647	&0.884&	0.792	&0.714&	0.609	&0.654&	0.739	&0.689	&0.725&	0.513	&0.702 \\
    RetinexMamba \cite{bai2024retinexmamba} & 0.817& \textcolor{red}{0.802} &	0.606&	\textcolor{red}{0.907}	&0.754	&0.688	&0.621	&0.596&	0.775	&0.720	&0.742	&0.595&	0.719\\
    LLFormer \cite{wang2023llformer} & 0.764&	 0.730&	0.683&	0.888	&0.787	&0.692	&0.650	&0.659&	0.731	&0.643	&0.717	&\textcolor{red}{0.612}&	0.713\\
    HVI-CIDNet \cite{yan2025hvi} & \textcolor{red}{0.881}&	 0.704&	0.664&	0.898	&0.798	&0.668	&0.673	&0.597&	0.726	&0.719	&0.762	&0.569&	0.722 \\
    \hline\rowcolor{gray!20} U-Net & 0.840&	0.739 &	\textcolor{red}{0.713}&	0.878	&\textcolor{blue}{0.817}	&0.716	&0.667	&0.656&	\textcolor{blue}{0.779}	&\textcolor{red}{0.806}	&\textcolor{red}{0.791}	&\textcolor{blue}{0.605}&	\textcolor{red}{0.751}\\
    \bottomrule
    \end{tabular}
\end{table*}

\begin{table*}[t]
\caption{Quantitative results of nighttime semantic segmentation on the ACDC dataset. The symbol set $\{\mathrm{RO}, \mathrm{SI}, \mathrm{BU}, \mathrm{WA}, \mathrm{FE}, \mathrm{PO}, \mathrm{TL}$, TS, VE, TE, SK, PE, RI, CA, TR, MO, BI\} represents \{road, sidewalk, building, wall, fence, pole, traffic light, traffic sign, vegetation, terrain, sky, person, rider, car, train, motorcycle, bicycle\}.  The table is organized as follows: unsupervised LLIE methods and methods trained on synthetic data (the top section); supervised LLIE methods (the second section) and our proposed U-Net (the third section). $\uparrow$ indicates that a larger value corresponds to better performance. In each class (\ie, column), the best result is in \textcolor{red}{red} color whereas the second best one is in \textcolor{blue}{blue} color.}
\label{tab:acdc}
\centering
\setlength{\tabcolsep}{2.5pt}
% \resizebox{\linewidth}{!}{
\begin{tabular}{l|ccccccccccccccccc|c}
\toprule
Class & RO & SI & BU & WA & FE & PO & TL & TS & VE & TE & SK & PE & RI & CA & TR & MO & BI & mIoU $\uparrow$ \\
\midrule
Zero-DCE  \cite{guo2020zero} &92.58 &69.79 &79.58 &46.15 &41.12 &50.14 &51.18 &43.03 &71.25 &\textcolor{red}{16.13} &82.43 &39.51 &5.72 &63.62 &82.42 &18.82 &\textcolor{red}{53.22} &47.72\\
EnlightenGAN \cite{jiang2019enlightengan} &93.01 &70.94 &79.23 &46.20 &41.26 &\textcolor{red}{52.03} &54.43 &42.28 &72.02 &12.30 &82.87 &40.20 &9.02 &64.28 &76.06 &20.05 &50.24 &47.71\\   
RUAS \cite{liu2021ruas} &89.58 &61.52 &78.34 &39.99 &39.38 &48.87 &45.90 &34.35 &70.26 &12.31 &82.99 &30.49 &0.34 &53.02 &81.32 &3.89 &39.77 &42.76\\
SCI \cite{ma2022toward} &92.49 &69.32 &79.50 &\textcolor{red}{47.48} &40.89 &51.18 &53.78 &42.18 &71.65 &13.48 &83.48 &\textcolor{red}{45.81} &14.64 &64.35 &81.30 &18.39 &50.74 &\textcolor{blue}{48.46}\\
PairLIE  \cite{fu2023pairlie} &92.92 &\textcolor{red}{71.22} &78.94 &47.22 &40.26 &51.35 &52.30 &38.49 &71.42 &14.68 &82.77 &39.38 &12.22 &61.03 &76.94 &14.48 &49.89 &47.13\\
NeRCo  \cite{yang2023nerco} &92.61 &69.68 &78.95 &45.70 & 40.44 & 50.56 & 47.88 & 44.24 & 71.12 & 10.20 & 82.61 & \textcolor{blue}{45.56} &7.64 &64.06 &72.95 &14.80 &45.62 &46.56 \\ 
%GDP \cite{fei2023gdp} \\
CLIP-LIT \cite{liang2023cliplit} &93.11 &70.97 &\textcolor{blue}{79.90} &47.12 &41.29 &50.83 &52.62 &44.20 &70.87 &11.42 &\textcolor{red}{83.96} &39.64 &14.46 &62.74 &80.78 &14.36 &\textcolor{blue}{53.09} &48.09 \\
AGLLNet \cite{lv2021attention} &\textcolor{blue}{93.13} &70.00 &79.24 &42.75 &41.16 &49.86 &53.93 &41.58 &71.38 &14.48 &82.56 &42.30 &6.50 &66.41 &79.48 &23.13 &51.03 &47.84 \\
%\rowcolor{gray!20} ReSCUE \\
\hline
RQ-LLIE \cite{liu2023rqllie} &92.32 &69.06 &78.60 &44.36 &\textcolor{blue}{42.53} &50.76 &51.76 &42.44 &\textcolor{blue}{72.10} &12.26 &\textcolor{blue}{83.55} &45.40 &\textcolor{blue}{15.36} &64.18 &78.53 &18.12 &50.02 &47.97 \\
MBLLEN   \cite{lv2018mbllen} &92.77&\textcolor{blue}{71.11}&77.84&42.43&37.63&50.39&\textcolor{blue}{55.09} &39.07&68.94&13.69&80.95&37.61&9.77&64.35&78.68&18.62&48.61&46.71    \\
Retinex-Net  \cite{chen2018retinex} &91.75 &66.98 &77.55 &43.40 &39.09 &48.88 &52.87 &33.96 &69.67 &9.55 &80.69 &34.25 &5.75 &65.53 &80.75 &5.85 &36.81 &44.39\\
KinD++   \cite{zhang2019kindpp} &92.78 &70.93 &78.72 &44.92 &41.01 &48.73 &53.66 &38.53 &70.36 &10.61 &81.98 &38.17 &2.42 &64.72 &76.77 &\textcolor{red}{25.36} &48.64 &46.75 \\
MIRNet    \cite{zamir2020mirnet} &92.64&68.73&79.54&45.10&40.94&50.80&\textcolor{red}{55.70} &39.97&71.48&14.07&82.87&40.84&\textcolor{red}{16.99} &62.50&80.59&21.75&48.67&48.06   \\
SNR-Net  \cite{xu2022snrnet} &91.33 &65.87 &77.17 &39.30 &39.65 &48.63 &50.66 &40.23 &69.77 &12.90 &80.31 &43.96 &11.00 &62.31 & \textcolor{red}{84.46} &15.18 &48.26 &46.37    \\  
IAT     \cite{cui2022iat} &92.74&70.25&\textcolor{red}{80.23}&46.01&\textcolor{red}{42.73}&50.14&52.15&44.22&\textcolor{red}{72.85} &10.17&83.40&41.09&12.81& \textcolor{blue}{66.44} &75.34&12.31&51.89&47.62     \\
LLFlow \cite{wang2022llflow} &92.57&69.83&79.13&42.05&41.09&49.21&54.39&42.51&71.14&12.31&81.81&40.11&11.14&64.52&78.30&16.66&49.42&47.17\\
Retinexformer \cite{cai2023retinexformer} &\textcolor{red}{93.25} &70.45 &79.73 &43.34 &42.27 &\textcolor{blue}{51.99} &52.62 &43.07 &72.08 &13.07 &82.78 &40.50 &10.48 &63.94 &77.07 &20.20 &48.63 &47.66 \\
RetinexMamba \cite{bai2024retinexmamba} &92.80 &69.28 &78.10 &45.72 &42.06 &49.02 &49.38 &41.05 &69.48 &\textcolor{blue}{15.11} &80.70 &40.82 &15.17 &64.59 &78.13 &12.27 &44.41 &46.74\\
%GLARE \cite{GLARE} \\
LLFormer \cite{wang2023llformer} &93.01 &70.48 &79.06 &41.95 &40.20 &50.50 &53.07 &43.95 &71.00 &12.94 &82.05 &44.01 &15.06 &64.94 &81.80 &14.93 &52.01 &47.95\\
HVI-CIDNet \cite{yan2025hvi} &92.76 &69.58 &79.72 &\textcolor{blue}{47.29} &40.87 &50.48 &53.91 &\textcolor{red}{45.79} &72.00 &12.21 &83.44 &40.94 &14.55 &65.96 &82.50 &13.75 &49.82 &48.19\\
\hline
\rowcolor{gray!20} U-Net & 92.42 & 68.90 & 79.02 & 44.24 & 41.47 & 51.43 & 54.00 & \textcolor{blue}{44.39} & 71.41 & 13.50 & 83.04 & 41.00 & 13.67 & \textcolor{red}{68.35} & \textcolor{blue}{84.05} & \textcolor{blue}{23.80} & 49.17 & \textcolor{red}{48.63}\\
\bottomrule

\end{tabular}
%}
\end{table*}

\noindent\textbf{Advancing Existing Methods with Synthesized Data.} 
We conduct experiments using SNR-Net \cite{xu2022snrnet}, Retinexformer \cite{cai2023retinexformer}, and RetinexMamba \cite{bai2024retinexmamba} to investigate the advantages of models initially trained on low-light data synthesized using our proposed pipeline and subsequently fine-tuned on paired datasets, compared to models trained exclusively on paired datasets. The results are presented in Table \ref{tab:other_method_comparision}, where improvements are highlighted in red and reductions in blue. Notably, LLIE methods that incorporated our synthetic pipeline in their training process demonstrate significant performance gains across most evaluation metrics for both LOL-v1 and LOL-v2 datasets.  Fig. \ref{fig:problem} provides visual comparisons, where it is evident that SNR-Net and Retinexformer, initially trained on synthetic data from our pipeline, produce more visually pleasing results by correcting inaccurate white balance and avoiding abnormal enhancements. Overall, both the quantitative and qualitative results underscore the effectiveness of our synthetic pipeline in enhancing the performance of LLIE models, yielding outputs that align more closely with human visual perception.

\subsection{High-level Perception}

Insufficient lighting also presents challenges for machine visual analytics. In this subsection, we further evaluate the effectiveness of LLIE methods as preprocessing techniques for machine vision tasks under low-light conditions, including object detection, semantic segmentation, and face detection, where the vanilla U-Net is not fine-tuned on the LOL-v1 or LOL-v2 datasets.

\subsubsection{Low-light Object Detection} 

We first compare the preprocessing effects of different enhancement algorithms for the object detection task under low-light conditions.  The experiments are conducted on the ExDark dataset \cite{loh2019getting}, which consists of 7,363 low-light images spanning 12 categories, each annotated with corresponding bounding boxes. Initially, we apply LLIE models to enhance the images in the ExDark dataset, where the supervised methods are trained on the LOL-v2 dataset. Subsequently, we fine-tune the object detection methods on the low-light enhanced dataset. The object detector employed in this study is YOLO-v5, which is pre-trained on the COCO dataset \cite{lin2014microsoft}. We calculate mAP scores as our primary evaluation metric for performance comparison. The results are presented in TABLE \ref{tab:exdark}, where we compare vanilla U-Net against 21 current SOTA methods. Notably, both AGLLNet and vanilla U-Net, trained on two large synthetic datasets, achieve the highest performance results. In contrast to AGLLNet, which leverages synthetic low-light levels and noise in the sRGB domain, vanilla U-Net is trained on a synthetic dataset simulated in the RAW domain, achieving much superior performance with an improvement of at least 0.11 in terms of mAP. These findings underscore the efficacy and practicability of our low-light synthetic pipeline.

\subsubsection{Nighttime Semantic Segmentation}  We then conduct experiments on the ACDC nighttime subset \cite{sakaridis2021acdc} to compare the effects of various enhancement algorithms as preprocessing steps for nighttime semantic segmentation, a critical and challenging task in autonomous driving applications. The ACDC nighttime subset consists of 400 training images, 106 validation images, and 500 test images. We use PSPNet \cite{zhao2017pyramid} as our baseline segmentation model. Since PSPNet was not originally designed for low-light conditions, we apply a "pretrain + fine-tune" strategy across all LLIE methods. Specifically, for each LLIE method, we fine-tune the pretrained PSPNet—initially trained on the COCO dataset \cite{lin2014microsoft}—using the 400 enhanced training images. The 106 enhanced validation images are used to select the optimal model for comparison, and we report performance results on the remaining 500 enhanced test images. TABLE \ref{tab:acdc} shows the results in terms of the mIoU metric on the 500 test images. We observe that no single LLIE method achieves overwhelming advantages across most classes. However, vanilla U-Net demonstrates the best and second-best results in four specific classes, surpassing all other algorithms except for IAT. Additionally, vanilla U-Net achieves the highest average performance overall, with an improvement of at least 0.17 in terms of mIoU.

\begin{figure}[t]
    \centering
    \includegraphics[width=1.0\linewidth]{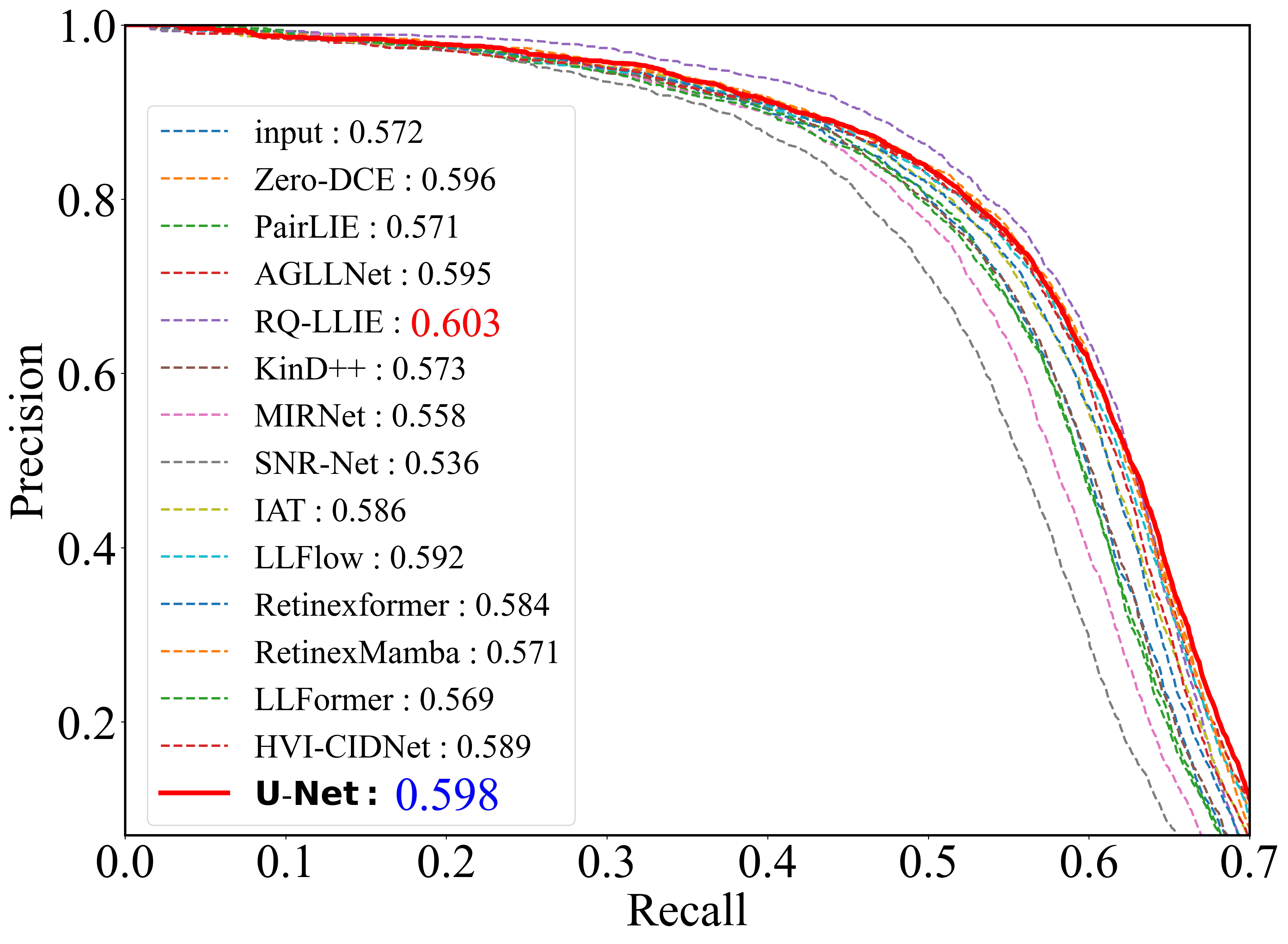}
    \caption{Precision-Recall (P-R) curves and Average Precision (AP) values at an IoU threshold of 0.6 on the DARK FACE dataset. the best AP result is in \textcolor{red}{red} text whereas the second best one is in \textcolor{blue}{blue} text. Best viewed on a high-resolution color display with zoom.}
    \label{fig:darkface}
\end{figure}

\subsubsection{Dark Face Detection}

\begin{table}[t]
\setlength\tabcolsep{3pt}
\caption{Quantitative comparisons of different low-light synthetic datasets, where vanilla U-Net is trained on datasets used by each respective LLIE method. }
\label{tab:comparision_synthetic_data}
\centering
\begin{tabular}{l|ccc|ccc}
\toprule
Metric &  PSNR$\uparrow$  & SSIM$\uparrow$  & LPIPS$\downarrow$  & MUSIQ$\uparrow$ & LIQE$\uparrow$ & Q-Align$\uparrow$\\  
\midrule
RetinexNet \cite{chen2018retinex}& 18.296 & 0.718 & 0.319  & 54.784 & 2.100 &  44.165 \\
MBLLEN \cite{lv2018mbllen}    & 19.837 & 0.782 & 0.224  & 55.170 & 2.249 & 44.662  \\
GladNet \cite{wang2018gladnet}   & 21.115 & 0.707 & 0.294 & 54.250 & 2.548 & 50.702  \\
AGLLNet \cite{lv2021attention}   & \textbf{21.758} & 0.798 & 0.227 & 55.685 & 2.401  & 46.158  \\
\rowcolor{gray!20} Ours  & 20.289 & \textbf{0.822} & \textbf{0.152} & \textbf{68.857} & \textbf{3.914} & \textbf{72.470}  \\ 
\bottomrule
\end{tabular}
\end{table}

Finally, we utilize the DARK FACE dataset \cite{darkface} to evaluate the high-level benefits of applying LLIE methods as a preprocessing step to improve face detection performance in low-light environments. The DARK FACE dataset contains 6,000 real-world images captured at night in various locations, such as teaching buildings, streets, bridges, overpasses, and parks, each annotated with bounding boxes for human faces, where 5,890 images are selected for training while the left 1,473 images are used for testing. We employ the YOLO-v5  \cite{zhang2017s3fd}, a prominent model in the field, as the detector and trained from scratch. Different low-light enhancement methods serve as the preprocessing modules with fixed parameters.
We depict the Precision-Recall (P-R) curves and mAP values at an Intersection over Union (IoU) threshold of 0.6 in Fig. \ref{fig:darkface}, using the evaluation tool\footnote{https://github.com/Ir1d/DARKFACE\_eval\_tools} provided with the DARK FACE dataset. The results show that vanilla U-Net achieves the second-highest AP under the given IoU threshold, following RQ-LLIE, which also performs well in terms of image enhancement (see TABLE \ref{tab:quantitative_comparision}). These findings indicate the practicality and feasibility of our synthetic pipeline for low-light vision tasks.

\subsection{Ablation Studies}
In this section, we begin by comparing our novel data synthesis pipeline with existing low-light image synthesis methods. Next, we conduct a quantitative assessment to evaluate the impact of key parameter configurations within our low-light image synthesis pipeline. For a fair comparison, the vanilla U-Net is trained exclusively on synthetic low-light data produced by our pipeline and is directly tested on the LOL-v2 dataset across all ablation studies.

\begin{table}[t]
\setlength\tabcolsep{3pt}
\caption{The ablation studies on noise synthesis, \eg, the impact of not adding noise or synthesizing noise in the sRGB domain or RAW domain. The default setting is highlighted with a \colorbox{gray!20}{gray background}.}
\centering
\begin{tabular}{l|ccc|ccc}
\toprule
Metric & PSNR$\uparrow$ & SSIM$\uparrow$  & LPIPS$\downarrow$  & MUSIQ$\uparrow$  & LIQE$\uparrow$  & Q-Align$\uparrow$\\ 
\midrule
Without Noise & 19.221 & 0.742 & 0.274 & 57.828 & 2.387  & 52.109  \\
sRGB domain    & 20.010 & 0.813 & 0.181 & 63.541 & 3.076 & 62.789    \\
\rowcolor{gray!20} RAW domain  & \textbf{20.289} & \textbf{0.822} & \textbf{0.152} & \textbf{68.857} & \textbf{3.914} & \textbf{72.470}    \\ 
\bottomrule
\end{tabular}
\label{tab:noise_synthesis}
\end{table}

\begin{figure}[t]
\begin{minipage}[]{0.5\textwidth}
\centering
     \subfigure[Without noise]{\includegraphics[width=0.3\linewidth]{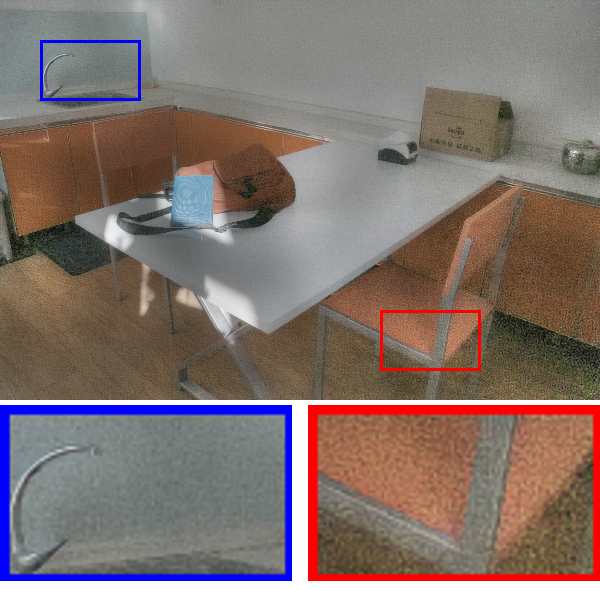}}\hskip.4em
    \subfigure[sRGB domain]{\includegraphics[width=0.3\linewidth]{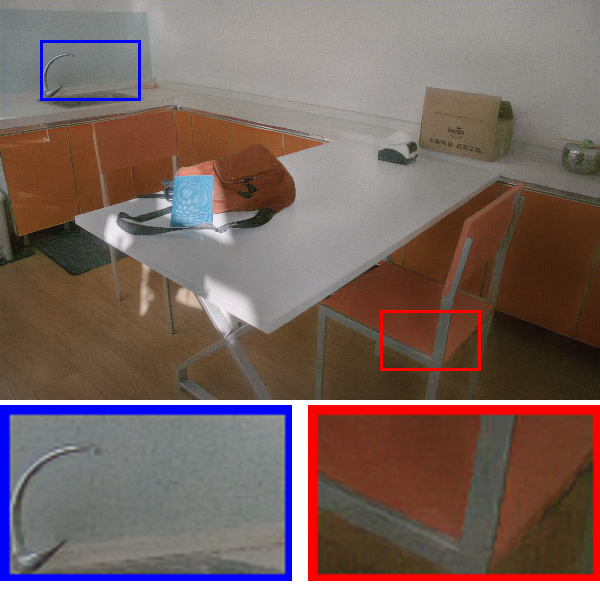}}\hskip.4em
    \subfigure[RAW domain]{\includegraphics[width=0.3\linewidth]{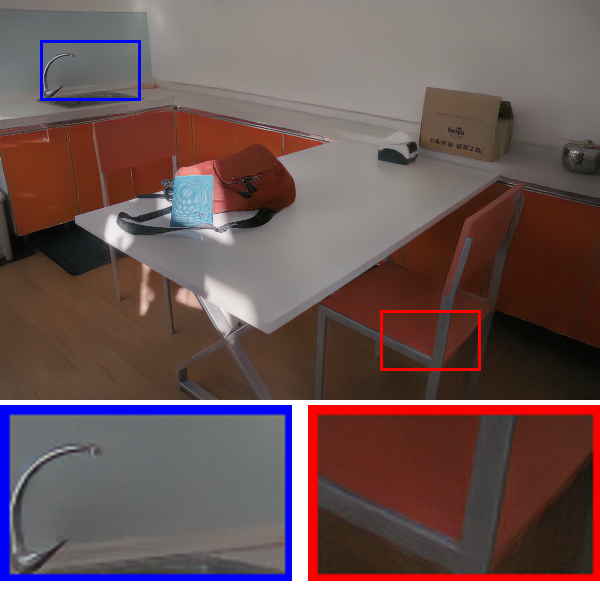}}
\end{minipage}

\caption{Ablation study on noise synthesis. Noise synthesis in the RAW domain demonstrates superior noise suppression performance. Please zoom in for details.}
\label{fig:qc_noise}
\end{figure}

\noindent\textbf{Comparison of Different Low-light Synthetic Datasets.} We begin by conducting a series of comprehensive experiments to compare the performance of various low-light synthetic datasets utilized for training the vanilla U-Net model, as presented inTABLE\ref{tab:comparision_synthetic_data}. The results indicate that the synthetic data generated by our pipeline outperforms other methods on both full-reference and no-reference IQA metrics. These findings provide strong validation for the effectiveness of our proposed low-light synthetic pipeline in producing high-quality, realistic low-light images, thereby effectively addressing the scarcity of real-captured paired datasets for LLIE.

\begin{table}[t]
\setlength\tabcolsep{5pt}
\caption{The ablation studies on the number of tone curves  used in our ISP pipeline. The default setting is highlighted with a \colorbox{gray!20}{gray background}.}
\centering
\begin{tabular}{l|ccc|ccc}
\toprule
Metric & PSNR$\uparrow$ & SSIM$\uparrow$  & LPIPS$\downarrow$  & MUSIQ$\uparrow$  & LIQE$\uparrow$ & Q-Align$\uparrow$\\ 
\midrule
10       & 18.573 & 0.794 & 0.169 & 68.499 & 3.887 & 71.837  \\
50       & 19.760 & 0.813 & 0.163 & 69.031 &  3.828&72.124    \\
100      & 19.428 & 0.816 & 0.161 & 68.664 & 3.846&70.702    \\
\rowcolor{gray!20} 200  &  \textbf{20.289} & \textbf{0.822} & \textbf{0.152} & \textbf{68.857} & \textbf{3.914}  & \textbf{72.470}  \\ 
\bottomrule
\end{tabular}
\label{tab:tone_curves}
\end{table}

\begin{figure}[t]
\centering
\begin{minipage}[]{0.5\textwidth}
\hskip -0.5em
\centering
     \subfigure[10]{\includegraphics[width=0.245\linewidth]{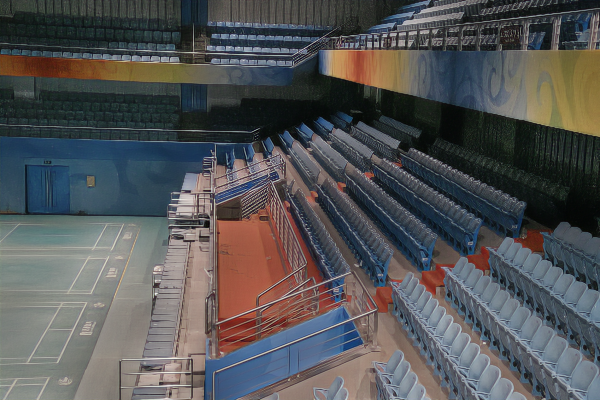}}\hskip.3em
    \subfigure[50]{\includegraphics[width=0.245\linewidth]{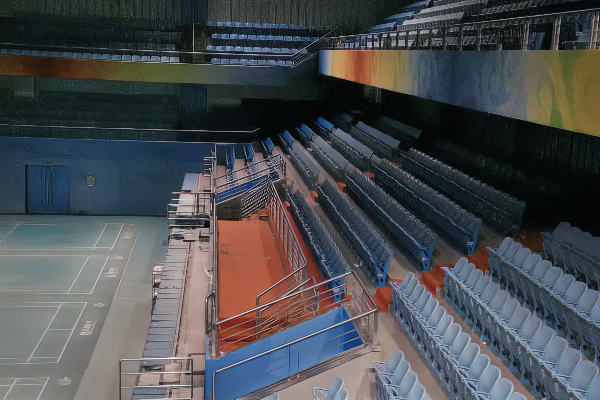}}\hskip.3em
    \subfigure[100]{\includegraphics[width=0.245\linewidth]{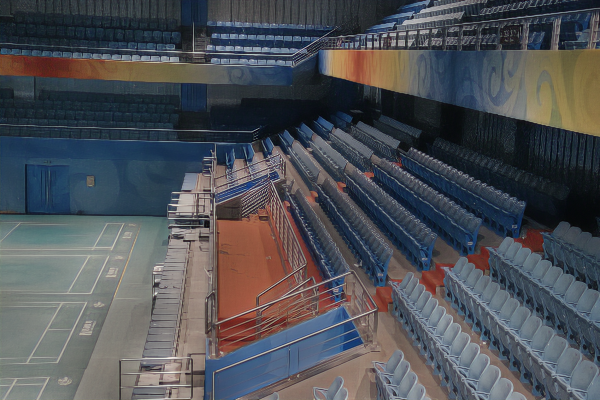}}\hskip.3em
    \subfigure[200]{\includegraphics[width=0.245\linewidth]{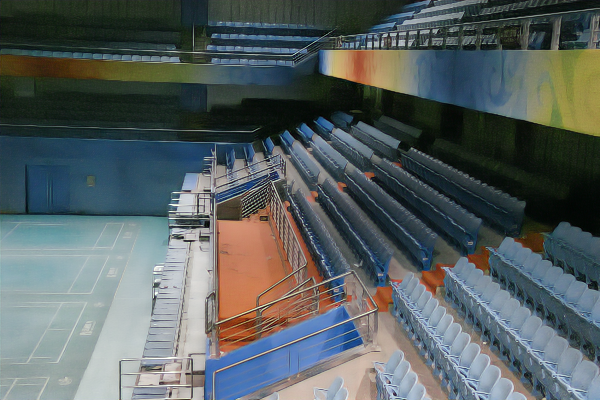}}
\end{minipage}
\caption{Ablation study on the number of tone curves. Increasing the number of tone curves results in more vivid color reproduction. Please zoom in for details.}
\label{fig:qc_Curve}
\end{figure}

\begin{table}[t]
\setlength\tabcolsep{5pt}
\caption{The ablation studies on the effect of exposure range. The default setting is highlighted with a \colorbox{gray!20}{gray background}.}
\centering
\begin{tabular}{l|ccc|ccc}
\toprule
Metric & PSNR$\uparrow$ & SSIM$\uparrow$  & LPIPS$\downarrow$  & MUSIQ$\uparrow$  & LIQE$\uparrow$ & Q-Align$\uparrow$\\ 
\midrule
(1,$2^5$)      & 19.502 & 0.811 & 0.160 & 68.724 &  3.833  &\textbf{75.525}  \\
(1,$2^{10}$)   & 19.888 & 0.813 & 0.152 & 68.077  & 3.621 & 70.045   \\
(1,$2^{15}$)   & 20.069 & 0.820 & 0.155 & 68.087  & 3.790 & 67.896   \\
\rowcolor{gray!20} (1,$2^{20}$)  & \textbf{20.289} & \textbf{0.822} & \textbf{0.152} & \textbf{68.857} & \textbf{3.914}  & 72.470\\
\bottomrule
\end{tabular}
\label{tab:reduction_range}
\end{table}

\noindent\textbf{Impact of Noise Synthesis.} Most low-light synthetic datasets add noise directly in the sRGB domain \cite{chen2018retinex}, which does not accurately represent sensor-specific noise, especially after processing through the ISP pipeline. TABLE \ref{tab:noise_synthesis} and Fig. \ref{fig:qc_noise} compare three noise synthesis approaches for low-light scenarios: without noise, noise synthesized in the sRGB domain, and noise synthesized in the RAW domain.  According to TABLE \ref{tab:noise_synthesis}, noise synthesized in the RAW domain yields superior performance across various quantitative metrics. Fig. \ref{fig:qc_noise} visually illustrates improved noise suppression with RAW domain noise synthesis. These results strongly validate the effectiveness of our low-light noise synthesis in the RAW domain, demonstrating its strong alignment with the noise characteristics in real-captured low-light images.

\noindent\textbf{Impact of Tone Curve Quantity}. Tone curves are employed to adjust the contrast, brightness, and colors of input images. Previous ISP synthesis pipelines typically utilized only a single tone curve for this purpose \cite{brooks2019unprocessing}. In contrast, our pipeline incorporates 200 distinct S-shaped tone curves to produce low-light images, introducing greater color variability in the low-light degradation space. To assess the impact of tone curve quantity in our pipeline, we conducted ablation experiments. TABLE \ref{tab:tone_curves} presents the quantitative results, revealing that increasing the number of tone curves significantly enhances performance. Fig. \ref{fig:qc_Curve} provides a visual comparison, showing that increasing the number of tone curves results in more vivid color reproduction. These findings highlight the efficacy of employing an expanded set of tone curves within our low-light synthetic pipeline, which serves to extend the color degradation space and enhance the vanilla U-Net model's capacity for color correction.

\noindent\textbf{Impact of of Exposure Reduction Ranges.} 
In our pipeline, the low-light effect is synthesized by varying the range of exposure reduction (0, $e$), where a larger $e$ results in a darker synthesized low-light image. Here, we investigate the impact of different exposure reduction ranges. Quantitative results are reported in TABLE \ref{tab:reduction_range}. The results show a notable improvement in most IQA metrics with an increased exposure reduction range. We argue this performance enhancement arises because the broader exposure reduction range generates more challenging samples. LLIE methods can further enhance model performance by facilitating learning from these difficult samples. These findings suggest that a wider spectrum of synthetic low-light conditions can increase the degradation space, thereby improving performance in darker environments.

\section{Conclusion}
\label{sec:conclusion}
In this paper, we tackle the critical challenge of limited and impractical training data in LLIE by introducing a novel data synthesis pipeline that serves as a robust foundation for training more effective LLIE models. By simulating low-light conditions directly in the RAW domain and processing them through multiple stages of the ISP pipeline—with variations in white balance, color space conversion, tone mapping, and gamma correction at each stage—we generate unlimited paired training data. This approach broadens the degradation space, enhancing the diversity and realism of the data, and effectively captures a wide range of degradations inherent in the ISP pipeline.
Extensive experiments using the vanilla U-Net model and existing SOTA LLIE methods across multiple datasets demonstrate that our data synthesis significantly improves performance, delivering high-fidelity enhancements that surpass current methods both quantitatively and visually. By focusing on data synthesis, we enhance the practicality and generalizability of LLIE models, effectively overcoming limitations posed by data scarcity and lack of diversity.
We believe our approach represents a significant step toward the practical deployment of LLIE technologies in real-world scenarios. Future research may refine the synthesis process and explore additional variations in the ISP stages to capture even more nuanced degradations.

\bibliographystyle{IEEEtran}
\bibliography{main}

\end{document}